\documentclass[twoside,11pt]{article}

\usepackage{bm}
\usepackage{url}
\usepackage{float}
\usepackage{subfig}
\usepackage{natbib}
\usepackage{amsmath}
\usepackage{afterpage}
\usepackage{enumerate,enumitem}
\usepackage{graphicx,psfrag,epsf}
\usepackage[ruled,linesnumbered]{algorithm2e}
\usepackage{jmlr2e}

\usepackage{xcolor}
\usepackage{afterpage}
\usepackage{multirow}
\usepackage{color, colortbl}
\definecolor{Gray}{gray}{0.9}

\newcommand{\minitab}[2][l]{\begin{tabular}{#1}#2\end{tabular}}
\usepackage{footnote}
\usepackage{rotating}
\makesavenoteenv{tabular}
\def\x{\mathbf{x}}
\def\te{\mathbf{\theta}}

\usepackage{lastpage}
\jmlrheading{22}{2021}{1-\pageref{LastPage}}{1/20; Revised 6/21}{6/21}{20-058}{Zebin Yang and Aijun Zhang}
\ShortHeadings{Hyperparameter Optimization via Sequential Uniform Designs}{Yang and Zhang}

\begin{document}
	
	\title{Hyperparameter Optimization via Sequential Uniform Designs}
	
	\author{\name Zebin Yang  \email yangzb2010@hku.hk\\
		\name Aijun Zhang \email ajzhang@umich.edu\\
		\addr Department of Statistics and Actuarial Science\\ 
		University of Hong Kong\\
		Pokfulam Road, Hong Kong}
	
	\editor{Isabelle Guyon}

	\maketitle
	
	\begin{abstract}%   <- trailing '%' for backward compatibility of .sty file
		Hyperparameter optimization (HPO) plays a central role in the automated machine learning (AutoML). It is a challenging task as the response surfaces of hyperparameters are generally unknown, hence essentially a global optimization problem. This paper reformulates HPO as a computer experiment and proposes a novel sequential uniform design (SeqUD) strategy with three-fold advantages: a) the hyperparameter space is adaptively explored with evenly spread design points, without the need of expensive meta-modeling and acquisition optimization; b) the batch-by-batch design points are sequentially generated with parallel processing support; c) a new augmented uniform design algorithm is developed for the efficient real-time generation of follow-up design points. Extensive experiments are conducted on both global optimization tasks and HPO applications. The numerical results show that the proposed SeqUD strategy outperforms benchmark HPO methods, and it can be therefore a promising and competitive alternative to existing AutoML tools.
	\end{abstract}
	
	\begin{keywords}
		Automated machine learning, Hyperparameter optimization, Global optimization, Sequential uniform design, Centered discrepancy. 
	\end{keywords}
	
	\section{Introduction} \label{introduction}
	Machine learning models are becoming increasingly popular due to their strong predictive performance. Meanwhile, the number of hyperparameters for these models also explodes, and we often have to spend considerable time and energy on hyperparameter tuning~\citep{probst2019tunability}, also known as hyperparameter optimization (HPO). Such HPO procedure is indeed essential but very tedious in machine learning. It is generally accepted that a manual tuning procedure often fails to achieve the best model performance, and faces the critical reproducibility issue. In recent years, the automated machine learning (AutoML) has attracted great attention, highlighting the automatic procedure of hyperparameter tuning.
	
	% Motivation 
	In this paper, we introduce the classical design of computer experiments to solve the HPO problem with the purpose of maximizing algorithmic prediction accuracy. A computer experiment is defined as a deterministic function or code that is very complicated and time-consuming to evaluate~\citep{fang2006design}. We may view HPO as a special computer experiment in the sense that each hyperparameter configuration is an input, and the corresponding predictive performance is the output. In Figure~\ref{fig:AutoML}, a sequential uniform design (SeqUD) approach is proposed for such HPO-type of computer experiment. It is a multi-stage coarse-to-fine optimization framework based on uniform exploration and sequential exploitation. At each stage, the search space is automatically adjusted, and a new batch of design points is augmented with uniformity consideration.
	\afterpage{
		\begin{figure}[!t] 
			\centering
			\includegraphics[width=0.8\textwidth]{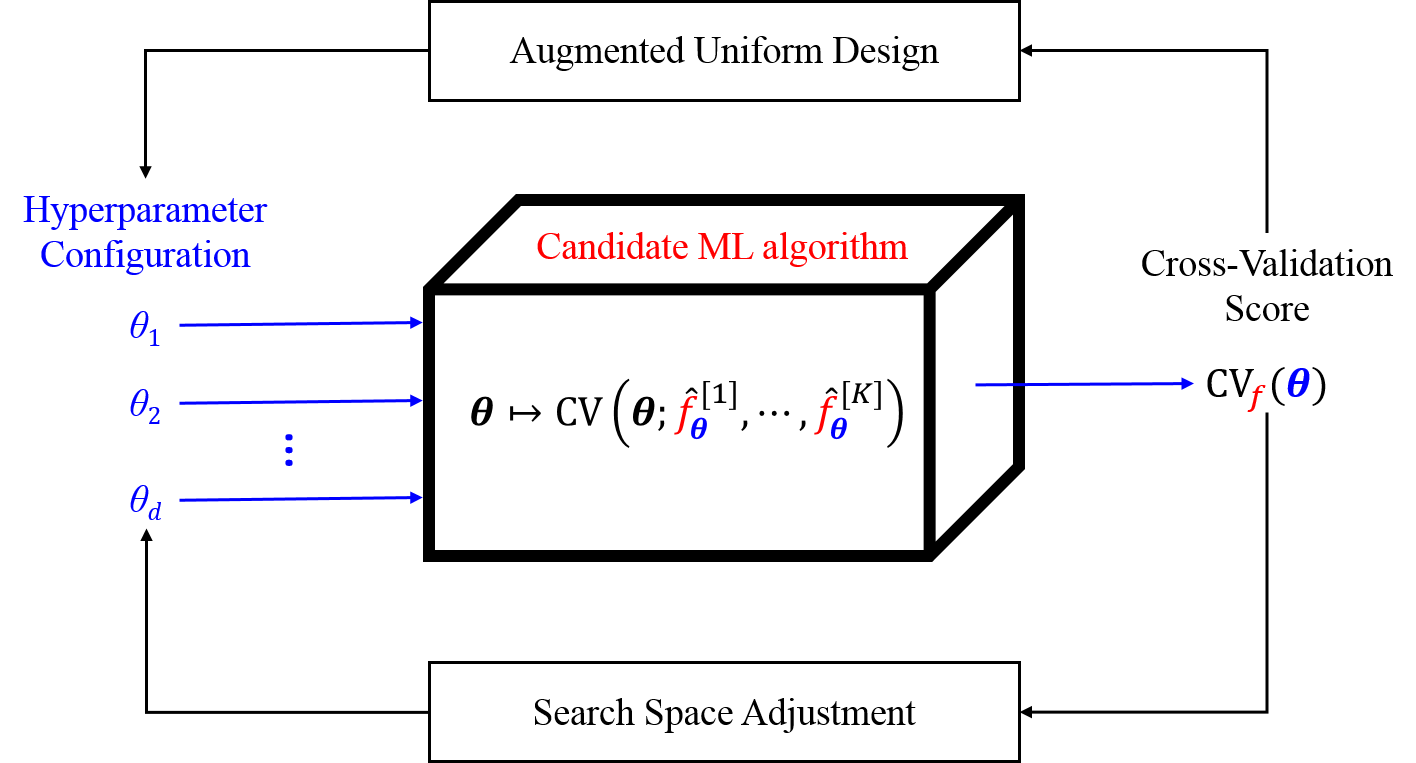}
			\caption{HPO problem reformulated as a kind of computer experiment based on sequential uniform designs.}\label{fig:AutoML}
		\end{figure}
	}
	
	% Advantages in concept and experiments
	Compared with existing HPO methods in the literature, 
	%to be reviewed in Section~\ref{Related_Work}, 
the proposed SeqUD approach has the following advantages: a) the hyperparameter trials are defined in a sequential manner, as the SeqUD points are generated based on the information of existing design points; b) each time the uniform design of points are considered, which targets the most representative exploration of the search space; c) the SeqUD points generated at the same stage could be evaluated in parallel, which brings additional computation efficiency especially when training large-scale machine learning algorithms. In addition, the proposed method differs from the commonly used Bayesian optimization methods, as SeqUD is free from the time-consuming surrogate modeling and acquisition optimization procedures.
	
	The proposed SeqUD method is tested through extensive synthetic global optimization tasks and real-world HPO experiments. The machine learning algorithms under our consideration include the support vector machine (SVM), extreme gradient boosting (XGBoost), and a machine learning pipeline that involves data preprocessing, feature engineering, model selection, and hyperparameter tuning. Based on the results for a large amount of regression and classification data sets, it is demonstrated that the SeqUD method outperforms Bayesian optimization and other benchmark methods. In summary, this paper contributes to the HPO and AutoML literature in the following three aspects:
	\begin{itemize}
		\item We develop a novel AugUD algorithm for the efficient augmentation of uniform design points. This is not only a new idea in AutoML, but also a new contribution to the field of experimental designs. 
		
		\item The SeqUD strategy generalizes Bayesian optimization from one-point-at-a-time to batch-by-batch. Meanwhile, it avoids expensive meta-modeling and acquisition optimization procedures. The improved effectiveness and efficiency are achieved. 
		
		\item Two open-source Python packages are developed, including the \textsl{PyUniDOE} package that employs efficient C++ code for generating uniform designs, and the \textsl{SeqUD} package that implements the proposed SeqUD method for HPO tasks.
	\end{itemize}
	
	The rest of the paper is organized as follows. Section~\ref{Related_Work} reviews the related HPO literature. In Section~\ref{AugUD}, we introduce the background of uniform designs and develop the AugUD algorithm. The new SeqUD method is proposed for HPO in Section~\ref{SeqUD}.  A large amount of global optimization experiments and HPO experiments are presented in Section~\ref{Synthetic_experiments} and Section~\ref{AutoML_experiments}, respectively. Finally, we conclude in Section~\ref{Conclusion} and outline future works.
	
	\section{Related Work} \label{Related_Work}
	This section reviews existing HPO methods from non-sequential and sequential perspectives.
	
	\subsection{Non-sequential HPO Methods}
	Most of the practical HPO methods are non-sequential, as they are easy to implement and can evaluate multiple hyperparameter trials in parallel. For machine learning models with one or two hyperparameters (e.g., SVM with regularization strength and kernel width), it is common to use the exhaustive grid search method~\citep{chang2011libsvm}. The random search is an alternative method that generates design points randomly in low or high dimensions~\citep{bergstra2012random}. It is commonly believed that the random search is more flexible and useful for machine learning tasks. % with low-dimensional hyperparameters.
	
\begin{figure}[!t]
		\centering
		\subfloat[Grid Search]{
			\label{Demo_Grid} %% label for first subfigure
			\includegraphics[width=0.45\textwidth]{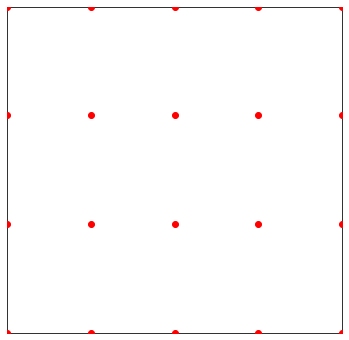}}
		\subfloat[Random Search]{
			\label{Demo_Random} %% label for first subfigure
			\includegraphics[width=0.45\textwidth]{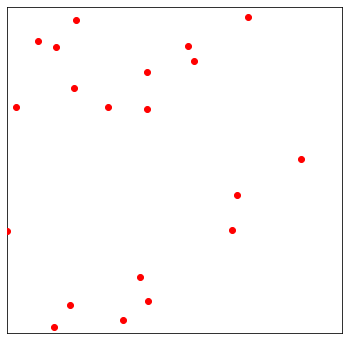}}\\
		\subfloat[Sobol Sequence]{
			\label{Demo_Sobol} %% label for second subfigure
			\includegraphics[width=0.45\textwidth]{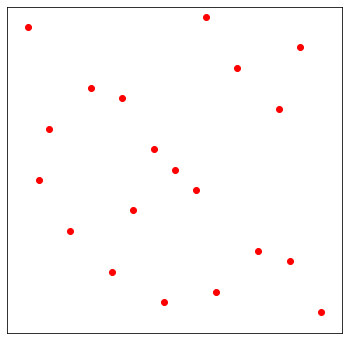}}
		\subfloat[Uniform Design]{
			\label{Demo_AugUD} %% label for second subfigure
			\includegraphics[width=0.45\textwidth]{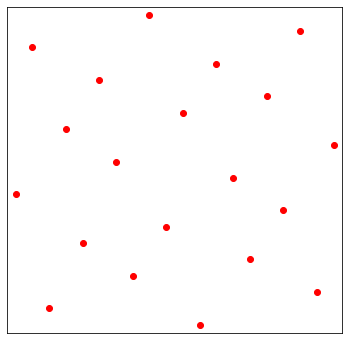}}
		\caption{An example that compares four different designs in a 2-D design space.}
		\label{UD_Demo} %% label for entire figure
	\end{figure}

	The space-filling design is an optimal strategy when there is no prior information about hyperparameter distribution~\citep{crombecq2011novel}, for which the uniform designs~\citep{fang2000uniform}, Sobol sequences~\citep{sobol1998quasi} and Latin hypercube sampling (LHS;~\citealp{kenny2000algorithmic}) can be used. These methods can generate design points with a low discrepancy from the assumed uniform distribution. Given the same maximal number of runs, the space-filling design has a lower risk of missing the optimal location than the random search. We demonstrate grid search, random search, Sobol sequences, and uniform designs in Figure~\ref{UD_Demo}, where each method generates 20 design points in the same two-dimensional space. For the random search, it is observed that design points can be clustered, while lots of other areas remain unexplored. The design points generated by Sobol sequences spread more evenly than the random search. The best coverage is by the uniform design, as it is constructed by optimizing a measure of uniformity to be introduced in Section~\ref{AugUD}. 
	
	\subsection{Sequential HPO Methods}
	Sequential methods are adaptive variants of non-sequential methods, where new design points are generated based on the information of existing design points. Exploration and exploitation are two synergistic  objectives of sequential methods. The former aims to better explore the search space, and the latter aims to find the global optimum.
	
	Bayesian optimization~\citep{jones1998efficient} is the most widely used sequential approach, which samples one-point-at-a-time in the search space.  At each iteration, it fits a surrogate model for approximating the relationship between the design points and the evaluated outcomes. Then, the next design point is generated by optimizing a predefined acquisition function. For HPO tasks in machine learning, the three most influential Bayesian optimization works are arguably the GP (Gaussian process)-EI (expected improvement) method~\citep{snoek2012practical}, the sequential model-based algorithm configuration (SMAC;~\citealp{hutter2011sequential}) and the tree-structured parzen estimator (TPE;~\citealp{bergstra2011algorithms}), which we review below.
	
	\medbreak
	\textbf{GP-EI}. It uses the GP as the surrogate model and selects the next design point by maximizing the following EI acquisition function:
	\begin{equation} \label{GPEI}
	\begin{split}
	\alpha_{\mbox{EI}}(\x) = \sigma(\x) \Big[z^{*} \Phi(z^{*}(\x))+\phi(z^{*}(\x)) \Big], 
	\end{split}
	\end{equation}
	where  $ z^{*}(\x) = (\mu(\x)-y^{*})/\sigma(\x)$, $y^{*}$ is the observed maximum, and $ (\mu(\x), \sigma^{2}(\x))$ are the GP-predicted posterior mean and variance, respectively. 
	
	\medbreak
	\textbf{SMAC}. It uses the random forest as the surrogate model. Compared to GP, the random forest can be easily scaled up to high-dimensional settings and are more flexible for handling discrete hyperparameters. However, as pointed out by \citep{shahriari2016taking},  SMAC has a potential drawback in that the estimated response surface is discontinuous, which makes the optimization of acquisition functions difficult.
	
	\medbreak
	\textbf{TPE}. It models the conditional distribution $p(\x|y)$ instead of $p(\bm y|\x)$, and then the EI acquisition function can be parameterized by  
	\begin{equation} \label{TPE}
	\begin{split}
	\alpha_{\mbox{EI}}(\x) \propto \left( \frac {  \ell( \x ) } { g ( \x ) } \gamma + 1 - \gamma \right) ^ { - 1 }.
	\end{split}
	\end{equation}
	Note that the equation is slightly different from its original form in~\citet{bergstra2011algorithms}, as we define it as a maximization problem. The functions $ g ( \x )$ and $\ell ( \x ) $ denote the density functions of $ x|y $ as $y \geq y^{*}$ and $y < y ^{*}$, respectively; they are estimated by hierarchical Parzen estimators. The value of $ y ^ { * } $ is chosen as a quantile $\gamma$ of the observed $y$ values, such that $p(y < y^{*}) = \gamma$~\citep{bergstra2011algorithms}.
	
	These Bayesian optimization methods have been implemented in various AutoML software packages. For example, the GP-EI method is implemented in the \textsl{Spearmint} package~\citep{snoek2012practical}; the SMAC method appears in \textsl{SMAC3}, \textsl{Auto-WEKA}~\citep{kotthoff2017auto} and \textsl{Auto-sklearn}~\citep{feurer2015efficient}; the TPE is wrapped into the \textsl{Hyperopt} package~\citep{komer2014hyperopt, bergstra2015hyperopt}. 
There exist variants of Bayesian optimization in other problem settings, for instance, high-dimensional tasks~\citep{wang2013bayesian, kandasamy2015high}, collaborative hyperparameter tuning~\citep{swersky2013multi,bardenet2013collaborative,feurer2015efficient}, no-regret Bayesian optimization~\citep{berkenkamp2019no}, and parallelized Bayesian optimization~\citep{snoek2012practical, hutter2012parallel}. One may refer to \cite{shahriari2016taking} for a comprehensive review. 
	
	There are other sequential HPO methods, e.g., evolutionary methods~\citep{escalante2009particle, di2018genetic} and reinforcement learning~\citep{lillicrap2015continuous, zoph2016neural}. These methods generally require extensive computing resources and are expensive in practice. One adaptive resource allocation idea is to allocate more computing resources to the hyperparameter configurations that tend to perform better. For instance, Successive Halving~\citep{domhan2015speeding} allocates the same budget to the best half hyperparameter configurations,  while the worst half would be dropped. Such procedure is repeated until only one hyperparameter configuration is left. 
Hyperband~\citep{li2017hyperband} can be viewed as a natural extension of Successive Halving, as it treats Successive Halving as a subroutine and calls it with different parameter settings.
	
	Another sequential method related to our work can be referred to~\cite{huang2007model}. It used a two-stage nested uniform design for training SVM with two hyperparameters, with 13 design points in the first stage (original search space) and 9 points in the second stage (adjusted search space). %As the trial points are specially designed, the total number of trial points is 21, with one duplicate point. 
This can be viewed as a special two-dimensional case of our SeqUD approach, but the follow-up 9 points were not optimized with respect to the existing 13 points in the first stage. 
%Moreover, it is not flexible to extend it to HPO tasks with different numbers of hyperparameters or design points. 
In this work, we propose a general framework of sequential uniform designs for HPO of various machine learning algorithms.
	
	\section{Augmented Uniform Design} \label{AugUD}
	The uniform design is a typical space-filling design~\citep{fang2000uniform}, which aims to find a discrete set of design points to represent the search space as well as possible. 
	%We briefly introduce the uniform design.
	\begin{definition}[Uniform Design]\label{def:UD}
		In the unit hypercube $C^{s}\equiv [0, 1]^s$, a uniform design with $n$ runs is the point set $D^*_{n} = \{\x_{1},\x_{2}, ..., \x_{n}\}$ that minimizes a certain discrepancy criterion:
		\begin{equation} \label{UD_objective}
		D_{n}^{*} \leftarrow \min_{D_{n} \subset C^{s}} \phi(D_{n}). 
		\end{equation}
	\end{definition}
	
	To limit the search space, the balanced U-type designs are often used for uniform design construction. Denote by $\mathbf{U}_{n, s}=\left(u_{i j}\right)$ a U-type design with $n$ runs, $s$ factors, and $q$ levels. Each factor with $q$ levels is a permutation of the balanced arrangement 
	\begin{equation}
	\{\underbrace{1,\cdots,1}_{n/q} , \cdots,\underbrace{q,\cdots,q}_{n/q}\},
	\end{equation}
	and it is required that  $n$ is divisible by $q$. One can convert it to the design matrix $\mathbf{X}_{n, s}=\left(x_{i j}\right)$  in $C^{s}$ by $x_{i j}=\left(2u_{i j}-1\right) /2q$. 
	Such a U-type design that minimizes a certain discrepancy criterion is called a U-type uniform design, denoted as $U_{n}(q^{s})$. More details about the uniform design theory and methods can be found in~\citet{fang1990sequential, fang1994number} and \citet{fang2000uniform, fang2006design, fang2018theory}.

	For example, a U-type uniform design $U_{20}(20^{2})$ with 20 runs, 2 factors, and 20 levels is shown in Table~\ref{UD_Table_Demo}, which corresponds to the last plot in Figure~\ref{UD_Demo}.  Below we provide the commonly used definitions for uniform designs, as well as their connections with HPO problems via Figure~\ref{fig:AutoML}. 
	\begin{itemize}
		\item Factors: the input variables of interest for a computer experiment, or the hyperparameters configured for a machine learning task.
		\item Levels: each factor or hyperparameter is assigned with discrete values within the experimental domain or the hyperparameter space. For a continuous factor, we typically divide it into $q$ levels. For categorical or integer-valued factors, their level assignment will be discussed in Section~\ref{SeqUD}.
		\item Design points: a design point is a possible level combination of different factors, and it corresponds to a unique configuration of hyperparameters to be conducted.
		\item Runs or trials: a run or trial corresponds to the implementation of a design point, and it generates the output or response for further analysis.
	\end{itemize}

	\begin{table}[htbp!]
		\renewcommand\tabcolsep{4pt}
		\renewcommand\arraystretch{1}
		\begin{center}
			\begin{tabular}{c|cccccccccccccccccccc}
				\hline
				No. &  1 & 2 & 3 & 4 & 5 & 6 & 7 & 8 & 9 & 10 & 11 &  12 &  13 & 14 & 15 & 16 & 17 & 18 & 19 & 20 \\ \hline
				$x_{1}$ &  16  &  18  &  12  &  19  &  1  &  10 & 9 & 4 & 2 & 14 & 6 & 15 & 5 & 20 & 11 & 13 &  8 & 7 & 3 & 17  \\
				$x_{2}$ &  15  &  19  &  1  &  3  & 9  &  7 & 20 & 13 & 18 & 10 & 16 & 5 & 6 & 12 & 14 & 17 &  4 & 11 & 2 & 8  \\ \hline
			\end{tabular}
		\end{center}
		\caption{U-type uniform design $U_{20}(20^{2})$.  Each integer of factors $x_{1}, x_{2}$ represents a level $k \in \{1,2,...,q\}$, which corresponds to $ (2k-1) / 2q$ in the design space.}	\label{UD_Table_Demo}
	\end{table}

	%Based upon the notion of uniform design, we define the augmented uniform design. 
	
	\begin{definition}[Augmented Uniform Designs]\label{def:AugUD}
		Given an initial uniform design $D_{n_{1}} \subset C^{s}$ and the follow-up run size $n_2$, 
		%(with $n_{1}$ runs) has already been evaluated. 
		an augmented uniform design is to find a follow-up design $D^*_{n_{2}}$ %(with $n_{2}$ runs) 
		that minimizes the discrepancy of the combined design, i.e., %  $D_{n_{1}} \cup D_{n_{2}}$
		%Mathematically, the optimal $D_{n_{2}}^{*}$ is obtained by 
		\begin{equation} \label{augud_obj}
		D_{n_{2}}^{*} \leftarrow \min_{D_{n_{2}}  \subset C^{s}} \phi \left(\left[\begin{array}{l}D_{n_{1}}\\ D_{n_{2}}\end{array}\right]\right).
		\end{equation}
	\end{definition}
	
	Clearly, when the initial design is empty, the augmented uniform design reduces to the ordinary uniform design as in Definition~\ref{def:UD}. Next, we introduce the discrepancy for measuring the uniformity of the design points. Traditionally, the star discrepancy~\citep{niederreiter1992random} is  the most popular choice in quasi-Monte Carlo methods. It is defined as the maximum deviation between the empirical distribution and uniform distribution,  
	\begin{equation}
	\phi(D_{n}^{*}) = \sup _{x \in C^{s}}\left|\frac{|D_{n} \cap[0, x)|}{N}-\operatorname{Vol}([0, x))\right|,
	\end{equation}
	where the symbol $|\cdot|$ denotes the number of points in a set and $\operatorname{Vol}([0, x))$ is the uniform distribution function on the unit cube $[0, x)$. Later, the generalized $\ell_{p}$-discrepancy~\citep{hickernell1998generalized} extends the star discrepancy to be more computationally tractable. Among various generalized $\ell_{p}$-discrepancies, the centered $\ell_2$-discrepancy ($\mbox{CD}_{2}$) is given by
	\begin{equation} \label{CD2}
	\begin{split}
	\mbox{CD}_2 (D_{n})^{2}  = & \left( {\frac{13}{12}} \right)^{s} -  \frac{2}{n}\sum\limits_{k=1}^{n} 
	\prod\limits_{j=1}^{s} \left( 1+\frac{1}{2}\left| {x_{kj} - \frac{1}{2}} \right|  
	-\frac{1}{2}\left| {x_{kj} - \frac{1}{2}} \right|^{2} \right) +                         \\
	& \frac{1}{n^2}\sum\limits_{k=1}^{n} \sum\limits_{j=1}^{n} \prod\limits_{i=1}^{s} \left[ 1+\frac{1}{2}\left| {x_{ki} - \frac{1}{2}} \right| + 
	\frac{1}{2}\left| {x_{ji} - \frac{1}{2}} \right|- \frac{1}{2}\left| {x_{ki} -x_{ji} } \right| \right].
	\end{split}
	\end{equation}
	The $\mbox{CD}_{2}$ discrepancy can be intuitively interpreted as the relative proportion of design points belonging to subregions of the search space~\citep{hickernell1998generalized}. It has several appealing properties: a) easiness to compute; b) projection uniformity over all sub-dimensions; c) reflection invariance around the plane $x_{j} = \frac{1}{2}$ (for any $j = 1, \cdots, s$).
	
	There also exist other uniformity criteria, including the wrap-around $\ell_2$-discrepancy ($\mbox{WD}_{2}$) and the mixture $\ell_2$-discrepancy ($\mbox{MD}_{2}$). They share similar properties as $\mbox{CD}_{2}$. For simplicity,  we use $\mbox{CD}_{2}$ as the default criterion for generating and evaluating uniform designs and augmented uniform designs throughout the paper.
	
	\subsection{Construction Algorithm} \label{AugUDConst}
	Due to the balance requirement for U-type designs, we assume the total number of runs $n = n_1 + n_2$ is divisible by $q$. The construction of U-type augmented uniform designs is a combinatorial optimization problem, which is extremely difficult for large design tables. The enhanced stochastic evolutionary (ESE) algorithm proposed by~\cite{jin2005efficient} is the most influential work for constructing space-filling designs. It is built upon threshold accepting (TA) and stochastic evolutionary (SE), for which a rather sophisticated procedure is designed to automatically control the acceptance threshold. This algorithm has been implemented in the R package~\textsl{DiceDesign}~\citep{dupuy2015dicedesign}.

	Inspired by ESE, we provide a simple yet effective AugUD algorithm for constructing augmented uniform designs. The proposed AugUD algorithm is composed of two nested loops. The inner loop rolls over columns for element-wise exchange while the outer loop adaptively changes the acceptance threshold. It involves the following critical steps.
	
	\medbreak
	\textbf{Initialization}. Given a fixed $D_{n_{1}}$, initialize the augmented design  $D_{n_{2}}^{*}$ such that $\left(\left[\begin{array}{l}D_{n_{1}}\\ D_{n_{2}}^{*}\end{array}\right]\right)$ is a balanced U-type design. Note that $D_{n_{2}}^{*}$ can be randomly generated or user-specified as long as it satisfies the balance requirement.
	
	\medbreak
	\textbf{Element-wise Exchange}. This is a basic procedure for searching optimal designs~\citep{fang2000uniform, jin2005efficient}. Given the current $D_{n_{2}}^{*}$, we randomly exchange two elements of a factor in $D_{n_{2}}^{*}$, in order to obtain a new combined design with improved uniformity. Repeat this operation for $M_{\rm E} $ times. According to the CD2 criterion  (\ref{CD2}), only a small part of terms needs to be updated for each element-wise exchange, which is an appealing property in practice as it can help save a lot of computing time.
	
	\medbreak
	\textbf{Threshold Accepting}. The TA strategy is employed for jumping out of local optima. The best candidate design $D_{n_{2}}$ obtained by element-wise exchange can be accepted with a probability:
	\begin{equation}\label{Acc_prob}
	p =  1 - \min \left(1, \max \left(0, \frac{\Delta}{T_{h}} \right)\right),
	\end{equation}
	where $T_{h} $ is the acceptance threshold for accepting suboptimal solutions and $\Delta$ is the change of the uniformity criterion:
	\begin{equation} \label{EXDiff}
	\Delta = \phi \left(
	\left[\begin{array}{l}D_{n_{1}}\\ D_{n_{2}}\end{array}\right]
	\right) - 
	\phi \left(
	\left[\begin{array}{l}	D_{n_{1}}\\D_{n_{2}}^{*}\end{array}\right]
	\right).
	\end{equation}
	When $\Delta<0$, $D_{n_{2}}$  will be accepted with the 100\% probability; otherwise, $D_{n_{2}}$ with worse performance may still be accepted with a probability $p$.
	
	\medbreak
	\textbf{Adaptive Threshold}. It is critical to select the threshold $T_{h}$ in the TA algorithm. We propose an adaptive updating rule for $T_{h}$. It is firstly initialized as 
	\begin{equation} \label{iniThres}
	T_{h} = \gamma \phi \left(\left[\begin{array}{l}D_{n_{1}}\\ D_{n_{2}}^{*}\end{array}\right]\right),
	\end{equation} 
	where $\gamma$ is a factor controlling the initial threshold. During optimization,  $ T_h $ can be adaptively updated by
	\begin{equation} \label{adaThres}
	T_{h}=\left\{
	\begin{aligned}
	T_{h} / \alpha  &\quad \mbox{if} \quad h_{i} < \eta,\\
	\alpha T_{h} &\quad \mbox{otherwise},
	\end{aligned}
	\right.
	\end{equation}
	where $\alpha$ is the scaling factor for adjusting the threshold. The symbol  $ h_{i} $ represents the hit ratio in the $i$th iteration (outer loop). When $ h_{i} $ is smaller than $\eta$, $ T_h $ will be increased for more exploration; when $ h_{i} $ remains a large value, the threshold should be decreased for better exploitation.
	
	A formal description of the AugUD construction is given in Algorithm~\ref{Algo_AugUD}, which requires input of multiple pre-specified parameters. In practice, we mimic the settings used in~\citet{jin2005efficient} and specify the parameters as follows.
	\begin{itemize}
		\item  $\gamma=0.005$. It denotes the multiplier of the initial acceptance threshold.
		\item $\eta=0.1$. The hit ratio threshold controls the acceptance threshold adjustment.
		\item $\alpha=0.8$. It is the scaling factor for adjusting the acceptance threshold.
		\item $ M_{\rm E} = \min\{50, 0.2 \times n_{2}^{2}(q-1)/(2q)\} $. It controls the number of element-wise exchange.
	\end{itemize}
	
	Finally, the numbers of loops are empirically determined as $ M_{\rm outer} = 50$ and $ M_{\rm inner} = 100$, which appear to work well in our tested cases. The AugUD algorithm can be further enhanced by restarting multiple times with different random seeds, and the one with the best uniformity criterion can be selected.
	\afterpage{
		\begin{algorithm}[!t]
			\caption{The proposed AugUD algorithm} \label{Algo_AugUD}
			\KwIn{
				$D_{n_{1}}$ (Existing Design), $n_1$ (\# Existing Runs), $n_{2}$ (\# Augmented Runs), $s$ (\# Factors), $q$ (\# Levels), $M_{\rm outer}, M_{\rm inner}$ (\# Outer and Inner Loops).
			}
			\KwOut{The optimal augmented design $D_{n_{2}}^{*}$.}
			Initialize $D_{n_{2}}^{*}$ and calculate $T_{h}$ by (\ref{iniThres}).\\
			\For{i = 1, 2, ..., $M_{\rm outer}$}
			{
				Set $k = 0$.\\
				\For{j = 1, 2, ..., $M_{\rm inner}$}
				{
					Select the column $j \pmod{s}$ of $D_{n_{2}}^{*}$.\\
					Randomly pick $ M_{\rm E} $ element pairs on the selected column.\\
					Perform element-wise exchange for selected pairs and evaluate discrepancy.\\
					Choose the best candidate design and calculate $p$ by (\ref{EXDiff}).\\
					Update $D_{n_{2}}^{*}$ and set $k=k+1$ (with the probability $p$).
				}
				Calculate the hit ratio $h_{i} = k / M_{\rm inner}$.\\
				Adaptively update $T_{h}$ by (\ref{adaThres}).
			}
		\end{algorithm}
	} % end of afterpage
	
	\medbreak
	\textbf{Software Implementation}.
	The proposed AugUD algorithm and related functionalities have been wrapped and implemented in our open-source Python package \textsl{PyUniDOE}~\footnote{\url{https://github.com/SelfExplainML/PyUniDOE}}. The core algorithm is written by C++ programming language, and we provide user-friendly application programming interfaces (APIs) in Python. It supports the real-time generation of uniform designs and augmented uniform designs under various uniformity criteria, e.g., $\mbox{CD}_{2}$, $\mbox{WD}_{2}$, and $\mbox{MD}_{2}$. In \textsl{PyUniDOE}, we also include a database of many state-of-the-art U-type uniform designs, which can be directly queried.
		
	\subsection{Experimental Evaluation} \label{cont_augud}
	The proposed AugUD algorithm serves as an efficient tool for the design community regarding the generation of both uniform designs and augmented uniform designs.
	
	\subsubsection{AugUD for generating uniform designs}
	
	We download the $U_{n}(n^{s})$ designs from the uniform design website~\footnote{\url{http://www.math.hkbu.edu.hk/UniformDesign}} maintained by Hong Kong Baptist University, which collects the commonly used uniform designs
	% (w.r.t. the $\mbox{CD}_{2}$ criterion).  It contains various U-type uniform designs 
with factors $s = 2, 3, \cdots, 29$. Each factor has different number of runs $n = s + 1, s + 2, \cdots, 30$. In total, 406 U-type uniform designs are downloaded. Although these designs are widely used, we experimentally show that most of them are not optimal.
	
\begin{figure}[!t]
		\centering
		\subfloat[Average Improvement Ratios]{
			\label{imp_num_avg} %% label for second subfigure
			\includegraphics[width=0.45\textwidth]{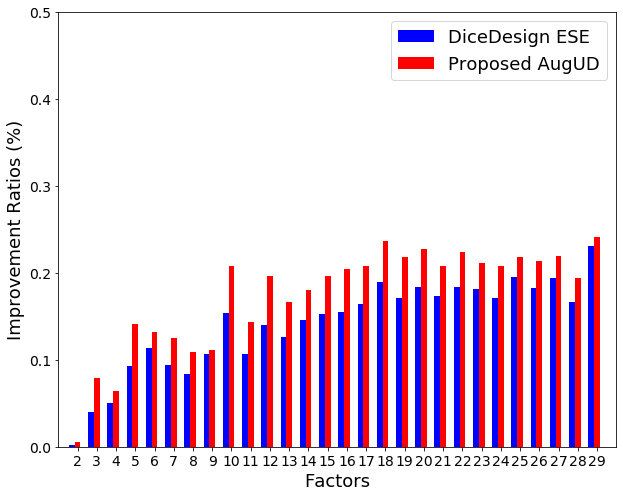}}
		\subfloat[Time Cost]{
			\label{ud_time} %% label for second subfigure
			\includegraphics[width=0.45\textwidth]{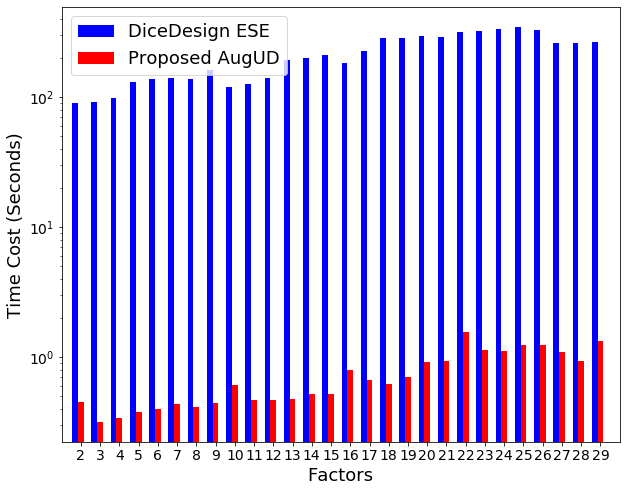}}
		\caption{Experimental results for generating uniform designs averaged against different number of factors. (a) Average improvement ratios compared with uniform designs obtained from the uniform design website; (b) Time cost.}
		\label{UD_OPT} %% label for entire figure
	\end{figure}

	We treat the downloaded U-type uniform designs as initializations and then compare the proposed AugUD algorithm (implemented in our \textsl{PyUniDOE} package) with the ESE algorithm (implemented in the \textsl{DiceDesign} package by~\citealp{dupuy2015dicedesign}). For a fair comparison, these two methods are configured with the same number of iteration loops and element-wise exchange. Each method is repeated ten times with different random seeds. For evaluation, we report the average and best performance of each method. The average $\mbox{CD}_{2}$ improvement ratios (over ten repetitions) of different initial uniform designs are reported in Figure~\ref{UD_OPT_ALL} in the appendix. For simplification, we aggregate the improvement ratios by the number of factors, and the averaged results against different numbers of factors are reported in Figure~\ref{UD_OPT}. For example, the bars at $x=2$ are averaged over all the 2-factor designs with runs ranging from 3 to 30, while the bars at $x=29$ only represent the results of $U_{30}(30^{29})$.
	
	It is surprising to observe that a large proportion of classic uniform design tables can be improved. According to the best designs found in the ten repetitions, 317 out of the tested 406 uniform designs have been improved by AugUD compared to 313 by ESE. This means that almost 80\% of existing uniform designs maintained by the uniform design website are not optimal, which further reflects the difficulty of generating uniform designs. Meanwhile, the best results of AugUD (over the ten repetitions) improve $\mbox{CD}_{2}$ of existing designs by 0.2433\% on average, which is much better than that of ESE (0.1875\%). It is also observed that the average improvement for 2-factor designs is relatively small compared to other high-dimensional designs. The reason is that these small-scale designs are easy to construct, and most of them are already optimal.
	
	In general, the proposed AugUD algorithm tends to outperform the ESE algorithm. Due to the efficient computational enhancement (e.g., C++ acceleration) in \textsl{PyUniDOE}, AugUD only uses around 0.34\% computing time of ESE (which is based on pure R language). The results demonstrate the superiority of the proposed AugUD algorithm and its efficient implementation in \textsl{PyUniDOE}. Moreover, it is worth mentioning that all the existing designs are already widely used, and such improvement marks a non-trivial contribution to the uniform design community.  
	
	\subsubsection{AugUD for generating augmented uniform designs} 
	
	In the literature, a commonly used strategy for generating sequential designs is the nested uniform design~\citep{fang1994number}, in which a new uniform design is embedded into existing designs. However, nested uniform designs do not consider the relationship between new and existing design points. The AugUD algorithm provides a practical and fast solution for augmenting design points subject to overall uniformity. Thus, AugUD can avoid clustered or duplicated design points and better explore the search space.
	
	Experimentally, we consider a test scenario in which 5 design points are randomly obtained from $U_{n}(n)^{s}$ for each $s = 2, \cdots, 24$, and $n = 8, \cdots , 30, (n > s + 5)$. These points are treated as existing designs, and we then augment $(n-5)$ design points to the design space. Five strategies are involved for augmentation, i.e., random augmentation, nested LHS (i.e., a new LHS is embedded into the design space), nested Sobol (i.e., a new Sobol sequence is embedded into the design space), nested uniform design (i.e., a new uniform design $U_{n-5}(n-5)^{s}$ is embedded into the design space; nested UD), and AugUD. Each method is repeated ten times, and we calculate the average $\mbox{CD}_2$ over ten repetitions for comparison.
	
\begin{figure}[!t]
		\centering
		\subfloat[Average Improvement Ratios]{
			\label{augud_imp_ratio} %% label for first subfigure
			\includegraphics[width=0.45\textwidth]{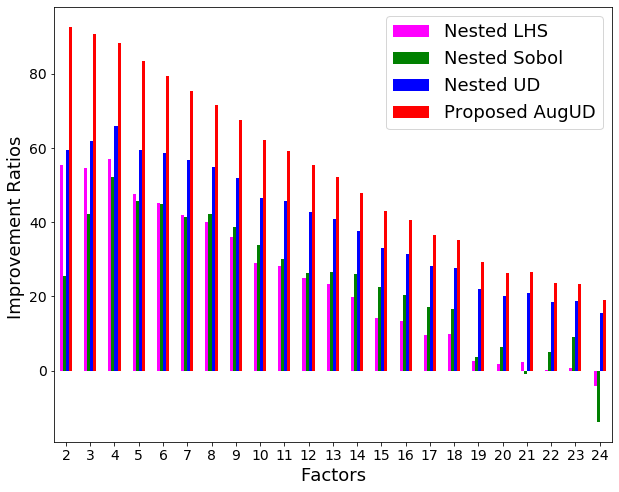}}
		\subfloat[Time Cost]{
			\label{augud_time} %% label for second subfigure
			\includegraphics[width=0.45\textwidth]{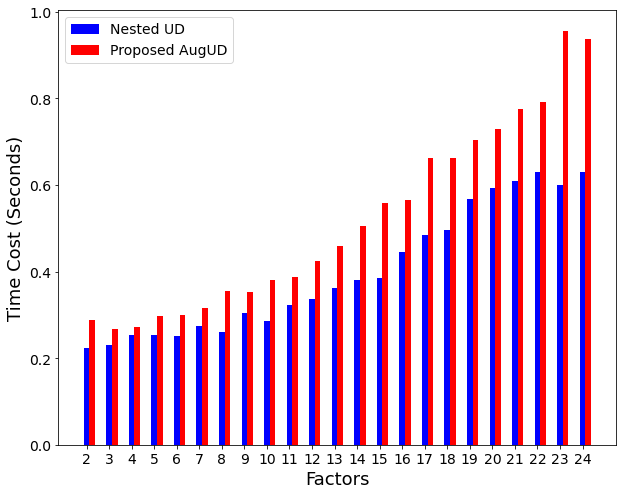}}
		\caption{Average results for generating design augmentation. (a) Improvement ratios of nested UD and AugUD against random augmentation; (b) Time cost.}
		\label{AUD_OPT} %% label for entire figure
	\end{figure}

	The detailed improvement ratios of AugUD and nested UD against random augmentation for each (factor, run) pair are reported in Figure~\ref{AUD_OPT_ALL} in the appendix. A summary of the experimental results is provided in Figure~\ref{AUD_OPT}. The bars at $x=2$ denote the improvement ratios averaged over all the 2-factor designs with runs ranging from 8 to 30. In contrast, the bars at $x=24$ only represent the average improvement ratio of the 30-run, 24-factor design. Both AugUD and nested UD show superior performance to random augmentation, nested LHS, and nested Sobol regarding the overall uniformity.
	
	It is also observed that AugUD significantly outperforms nested UD in all the compared cases. For AugUD, the improvement ratios show a decreasing trend as the number of factors increases. That is, large-sized designs are generally hard to optimize, and the augmented designs found by AugUD may be still not optimal. In Figure~\ref{augud_time}, the computing time of AugUD is slightly larger than that of nested UD (generated using our \textsl{PyUniDOE} package), as the evaluation of (\ref{CD2}) is a little bit expensive for large-sized designs, i.e., $n$ versus $(n-5)$. Note that nested LHS and nested Sobol can be easily generated without any complicated optimization, and hence their time costs are not reported (almost zero).
	
\begin{figure}[!t]
		\centering
		\subfloat[Random Augmentation \protect\\ \protect\centering $\mbox{CD}_2 = 0.02599$]{
			\label{Aug_Demo_Random} %% label for first subfigure
			\includegraphics[width=0.32\textwidth]{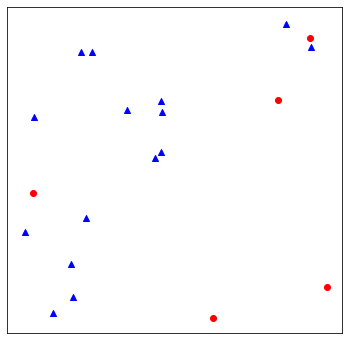}}
		\subfloat[Nested UD Augmentation \protect\\ \protect\centering $\mbox{CD}_2 = 0.00503$]{
			\label{Aug_Demo_Nested} %% label for first subfigure
			\includegraphics[width=0.32\textwidth]{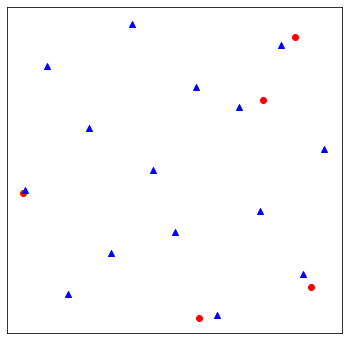}}
		\subfloat[AugUD \protect\\ \protect\centering $\mbox{CD}_2 = 0.00077$]{
			\label{Aug_Demo_AugUD} %% label for second subfigure
			\includegraphics[width=0.32\textwidth]{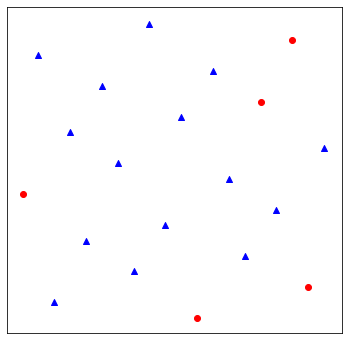}}
		\caption{A 2D example illustrating different augmentation methods. (a) Random augmentation; (b) Nested UD; (c) AugUD. In each plot, the 5 red dots represent the existing design $D_{n_{1}}$ and the 15 blue dots represent the augmented design $D_{n_{2}}$.}
		\label{AugUD_Demo} %% label for entire figure
	\end{figure}

	For illustration, Figure~\ref{AugUD_Demo} shows the results of adding 15 design points to 5 existing design points in a 2-D design space. It can be observed that the added points by random augmentation and nested UD can be quite close to existing points. In contrast, the proposed AugUD performs significantly better in terms of space-filling performance.
	
	%	\newpage
	\section{Sequential Uniform Design} \label{SeqUD}
	Sequential uniform design (SeqUD) is a general multi-stage optimization framework that incorporates uniform designs and a simple yet effective search strategy, i.e., sequential space halving. We first present the general SeqUD framework, then discuss its application for HPO, and finally remark its benefits and limitations.
	
	\subsection{SeqUD Framework} 
	The SeqUD framework comprises the following components: A) initial uniform design; B) subspace zooming \& level doubling for sequentially adjusting the search space; and C) uniform design augmentation.
	
	\medbreak
	\textbf{A) Initial Uniform Design}. At the first stage, a U-type uniform design $U_{n}(q^{s})$ is generated. These design points are evaluated through the corresponding experiments. Note that the number of initial design points $n$ and the level number $q$ should be predetermined for a specific task. For problems with possibly complex response surface, it is recommended to use more runs and more levels so that the initial design can better cover the search space. After initialization, steps B) and C) are repeated until the maximal number of runs $T_{\rm max}$ is reached. 
	
	\medbreak
	\textbf{B) Subspace Zooming \& Level Doubling}. At the $j$th stage ($j \geq 2$), the search space is halved into a subspace while the  doubled granularity. The optimal point $\x_{j}^{*}$ among all the evaluated design points is treated as the center of the new subspace, and the new search space is defined with levels (for each factor $i = 1,2,..., s$)
	\begin{equation}
	U_{j,i} = \left\{  x^{*}_{j,i} - \frac{{q-1}}{2^{j}q}, ..., x^{*}_{j,i}, ..., x^{*}_{j,i} + \frac{q-1}{2^{j}q} \right\}, 
	\label{subspace_bound1}
	\end{equation}
	when $q$ is odd, or
	\begin{equation} 
	U_{j,i} = \left\{ x^{*}_{j,i} - \frac{q - 2}{2^{j}q}, ..., x^{*}_{j,i}, ..., x^{*}_{j,i} + \frac{1}{2^{j}} \right\},
	\label{subspace_bound2}
	\end{equation}
	when $q$ is even. It is possible that the selected optimal center point is close to the search boundary, and some parts of the reduced subspace can be outside of $C^s$. Accordingly, we introduce a subspace shifting procedure to prevent this from happening. The reduced subspace is moved perpendicularly towards the inner side of the search space until all the levels are within the boundary. % $C^{s} = [0, 1]^{s}$.
	
	\medbreak
	\textbf{C) Uniform Design Augmentation}. This step is to augment design points in the reduced subspace. After the existing design points are converted to the new level space (\ref{subspace_bound1}) or (\ref{subspace_bound2}),  the new design points can be augmented via the AugUD algorithm. % to compose the new $U_{n}(q^{s})$.
	
	\medbreak
	A summary of the above procedures is provided in Algorithm~\ref{SeqUD_Algo}. As the termination of SeqUD is controlled by $T_{\rm max}$, the number of design points per stage should be accordingly specified. In the beginning, the granularity is generally not sufficient for finding the optimal point, e.g., for stages $j \leq 3$; as the search space is halved sequentially, the search space will be sufficiently small as, e.g., for stages $j \geq 10$. Further exploration over such a small region is somehow meaningless. Hence, the number of design points per stage should be roughly selected within the range $[T_{\rm max} / 10, T_{\rm max} / 3]$, considering the complexity of the task. For illustration, a two-stage example of SeqUD is provided in Figure~\ref{DemoSeqUD}. It can be observed that the search space is well covered by the initial uniform design. The reduced subspace is centered on the best-evaluated point and more design points can be added accordingly.
	%%% Algorithm table	
	\begin{algorithm} [!t]
		\caption{The proposed SeqUD framework} \label{SeqUD_Algo}
		\KwIn{$T_{\rm max}$ (\# Total Runs), $n$ (\# Runs per Stage),  $s$ (\# Factors), $q$ (\# Levels).}
		\KwOut{The optimal design point $ \x^{*} $ from all evaluations.}
		Generate an initial uniform design $U_{n}(q^{s})$.\\
		Evaluate each initial design point.\\
		Collect the design-response pairs $H = \{(\x_{1},y_{1}),...,(\x_{n},y_{n})\}$.\\
		Set stage $j=2$ and $T = n$.\\
		\While {True}	
		{
			Reduce the space of interest by centering on $ x^{*}_{j} = \mathop{\arg\max}_{\x \in H}{y}$.\\
			Count the number of existing design points in the subspace as $n_{e}$.\\
			Calculate the number of new design points to be augmented as $n_{j} =  n - n_{e}$.\\
			\lIf {$T+n_{j}>T_{max}$}	{break}
			Augment $ n_{j} $ design points via the AugUD algorithm.\\
			Evaluate each augmented design point, and update $ H $.\\
			Set $j = j+1$ and $T=T+n_{j}$.\\
		}
	\end{algorithm}
	
	\afterpage{
		\begin{figure}[!t] 
			\centering
			\includegraphics[width=0.4\textwidth]{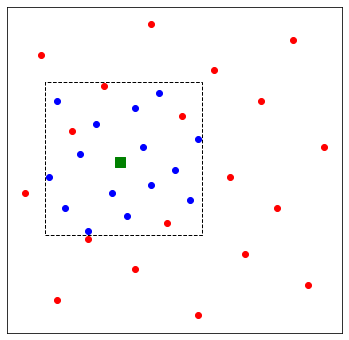}
			\caption{A two-stage example of SeqUD in a 2-D space. The circle points represent the initial uniform design via $U_{20}(20^{2})$. The surrounding box serves as the subspace of interest centered on the optimal design point $\x^{*}_{1}$ at the first stage, which is denoted by a square point in green. At the second stage, a new design is augmented (blue points) considering the overall uniformity within the subspace.} \label{DemoSeqUD}
		\end{figure}
	}
	
	\bigbreak
	\begin{remark}
		A sequential random search (SeqRand) approach is proposed as a na\"ive version of SeqUD. It is based on the space halving strategy while design points at each stage are randomly generated. Since uniformity is not a concern of SeqRand, the number of randomly generated design points can be directly set to $n$ for each stage. Moreover, SeqRand can also be viewed as a sequential version of the random search.
	\end{remark}

	\subsection{SeqUD for Hyperparameter Optimization}
	In this part, we present the details of applying SeqUD for HPO. Let $f$ be a function that measures the performance of a machine learning algorithm with hyperparameters $\te = (\theta_{1}, \theta_{2}, ..., \theta_{d})$, where $d$ denotes the number of hyperparameters to be optimized. Our objective is to find the best hyperparameter configuration $\te^{*}$ that maximizes $f$, which can be the cross-validation or hold-out validation score.
	
	Hyperparameters are usually of different types and scales. In HPO, the first step is to identify the tunable hyperparameters and corresponding search domains. Hyperparameters of machine learning algorithms may have various ranges and formats, and it is necessary to do some preprocessing. In general, they can be classified into three types, i.e., continuous (or numerical), integer-valued, and categorical. Continuous and integer-valued hyperparameters can be linearly transformed within the range $[0, 1]$. For categorical hyperparameters, one-hot encoding should be employed for transformation.
	
	SeqUD proceeds HPO as follows. First, as the initial uniform design is constructed, we inversely transform these design points to their original forms. For continuous hyperparameters, the inverse mapping can be directly implemented, while for integer-valued hyperparameters, they should be rounded to the nearby integers. Each categorical hyperparameter is represented by multiple dummy variables so that the corresponding design space dimension is greater than the number of hyperparameters. The encoded dummy variables are inversely transformed to corresponding hyperparameters by the arg max operation. More detailed discussion for handling categorical and integer-valued hyperparameters can be referred to~\cite{garrido2020dealing}.
	
	All the generated hyperparameter configurations are then evaluated by training the machine learning algorithm and calculating the predefined evaluation metric. The best performing configuration is selected for further investigation at the next stage. Through subspace zooming \& level doubling, new design space will be generated, and the AugUD algorithm can be used to augment sequential design points. After the optimization terminates, the machine learning model will be configured with the optimal hyperparameters and refitted to the whole training data.
	
	\medbreak
	\textbf{Software Implementation}. The procedures mentioned above are wrapped in our Python package \textsl{SeqUD}~\footnote{\url{https://github.com/SelfExplainML/SeqUD}}. It includes the proposed SeqUD method and some related benchmark methods with an interface to the well-known machine learning platform \textsl{scikit-learn}. In addition to  SeqUD and SeqRand, the APIs for some non-sequential methods are also provided by \textsl{SeqUD}, including grid search, random search, uniform designs, Latin hypercube sampling (by Python package \textsl{pyDOE}), and Sobol sequences (by Python package \textsl{sobol\_seq}). Moreover, the three classic Bayesian optimization methods are included by using the interfaces of \textsl{Hyperopt}, \textsl{Spearmint}, and \textsl{SMAC3}.
	%  see the  \textsl{SeqUD} manual~\footnote{\url{https://zebinyang.github.io/SeqUD}} for details.
	
	\subsection{Discussion}
	The idea behind the proposed SeqUD framework is intuitive and straightforward. It uses the space halving strategy to adjust the search space, and the main difference between SeqUD and other space halving-based approaches lies in its uniformity consideration. Compared to coarse-to-fine grid search methods, a) SeqUD is not limited to low-dimensional problems; b) the uniformity of new design points with existing design points is considered. Thus, it should have better optimization performance. We summarize its beneficial aspects as follows.
	
	\begin{itemize}
		\item SeqUD shares the benefits of all the other sequential methods. Except for the initial design, design points in SeqUD are sequentially constructed based on the preliminary information of existing design points. This procedure is more flexible and efficient than non-sequential methods, e.g., grid search and random search. 
		\item SeqUD makes a good balance between exploration (by uniform designs) and exploitation (by sequential space halving). SeqUD is less likely to be trapped into local areas for complicated hyperparameter response surfaces as the design points are uniformly located in the area of interest.
		\item SeqUD is free from the surrogate modeling and acquisition optimization used in Bayesian optimization. These procedures are all difficult tasks. For example, the GP model may fail when design points are close to each other; building a random forest on the hyperparameter space may be more expensive than conducting the experiments; in high-dimensional settings, it may find the best design point using the fitted surrogate model is also time-consuming. In contrast, new design points in SeqUD can be quickly generated without too much computation.
		\item Design points generated at the same stage can be evaluated in parallel. Given sufficient computing resources, this property will bring significant computation efficiency, especially for training large-scale machine learning algorithms. Methods like GP-EI, SMAC, and TPE, are initially designed to select new design points one by one, leading to a waste of computing resources. There also exist some strategies for speeding up computations for Bayesian optimization methods~\citep{snoek2012practical, hutter2012parallel}, while the optimization performance may be sacrificed. Also, these methods are not natural for performing parallelization~\citep{shahriari2016taking}.
	\end{itemize}
	
	The possible limitation of SeqUD, SeqRand, and all the other space halving strategies lies in the local optima problem. Although sampling uniformity is considered in SeqUD, it can still be trapped into local optima. This problem can be mitigated by performing more exploration over the search space, and in practice, the following two ways can be used to enhance the exploration.
	\begin{itemize}
		\item Employ more design points per stage for complex tasks, so that the algorithm is less likely to be trapped into locally optimal areas.
		\item Multiple shooting, i.e., except for zooming into the best-evaluated point per stage, we may simultaneously search the nearby subspace of the second and the third-best points (when these points are distant from each other).
	\end{itemize}
	
	Given a sufficient number of runs, these two strategies may help increase the success rate of optimization. However, as the total budget is usually limited, the trade-off between exploration and exploitation still exists.
	
	\section{Experiments for Global Optimization}\label{Synthetic_experiments}
	Extensive synthetic functions are involved in testing the performance of SeqUD on global optimization tasks. The benchmark models include grid search (Grid), random search (Rand), Latin hypercube sampling (LHS), Sobol Sequences (Sobol), uniform designs (UD), sequential random search (SeqRand), GP-EI, SMAC, and TPE. A total budget of 100 runs is allowed for each method. That is, all the compared methods can evaluate at most 100 design points. Finally, grid search is only tested on 2-D tasks. 
	
	All the benchmark methods are kept to their default settings. In SeqUD, we set the number of runs and levels per stage as $n = q = 15$ when $s \leq 5$; otherwise, we use $n = q = 25$ for higher-dimensional tasks. This setting compromises exploration and exploitation and works well in our experiments. For a fair comparison, the SeqRand approach is also configured with 15 or 25 runs per stage (depending on $s$). All the global optimization experiments are repeated 100 times. Throughout this paper, the statistical significance is reported based on paired t-test, with a significance level of 0.05. Due to the page limit, results reported in tables are all rounded to a certain precision, while the rank and statistical significance comparisons among different methods are based on the original precision.
	
	\subsection{Example Functions} \label{synthetic_examples}
	The working mechanism of each compared method is first investigated through the following 2-D synthetic functions.
	
\begin{figure}[!t]
		\centering
		\subfloat[3-D Surface]{
			\label{cliff_3D} %% label for first subfigure
			\includegraphics[width=0.38\textwidth]{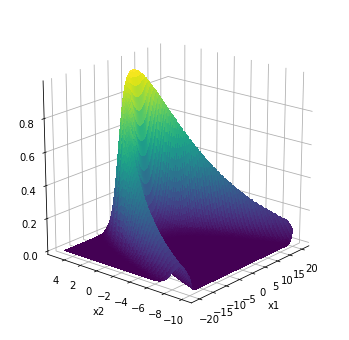}}
		\subfloat[SeqUD Evaluated Points]{
			\label{cliff_Sequd} %% label for second subfigure
			\includegraphics[width=0.33\textwidth]{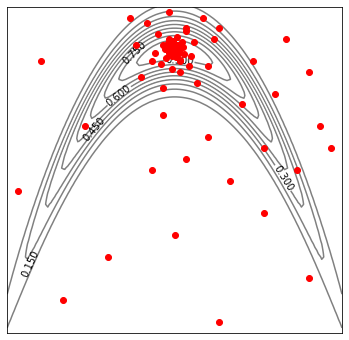}}
		\caption{The 3-D surface of the cliff function and SeqUD evaluated points against the ground truth contour plot.}
		\label{cliff} %% label for entire figure
		
		\centering 
		\subfloat[3-D Surface]{\label{octopus_3D}	
			\includegraphics[width=0.38\linewidth]{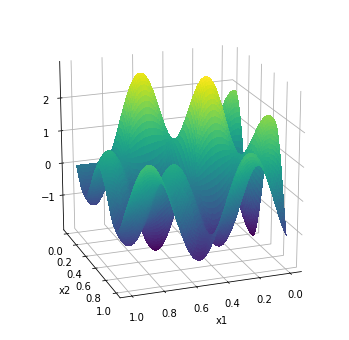}} 
		\subfloat[SeqUD Evaluated Points]{\label{octopus_Sequd}	
			\includegraphics[width=0.33\linewidth]{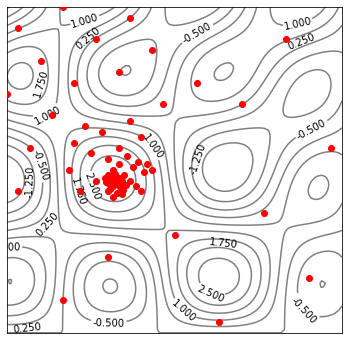}} 
		\caption{The 3-D surface of the octopus function and SeqUD evaluated points against the ground truth contour plot.}
		\label{octopus}
	\end{figure}

	\medbreak
	\textbf{Cliff Function}. The first example is obtained from~\citet{haario1999adaptive, haario2001adaptive}. As shown in Figure~\ref{cliff_3D}, its mesh plot looks like a ``cliff'', where the non-zero-value area is narrow and long.
	\begin{equation}
	\begin{split}
	\begin{aligned}
	f_{1} \left(x_{1}, x_{2} \right) = & \exp \left\{-\frac{1}{2} \frac {x_{1}^{2}}{100}  - \frac{1}{2} \left(x_{2} + 0.03 x_{1}^{2} - 3 \right)^{2}\right\}, \\
	& x_{1} \in [-20,20 ] , x_{2} \in [-10,5].
	\end{aligned}
	\end{split}
	\label{SimuA}
	\end{equation}
	
	\medbreak
	\textbf{Octopus Function}. The second scenario is much more complicated, with multiple local extrema within the response surface~\citep{renka1999algorithm}. Accordingly, we name it ``octopus'' due to its shape as shown in Figure~\ref{octopus_3D}.
	\begin{equation}
	\begin{split}
	\begin{aligned}
	f_{2}(x_{1}, x_{2}) = & 2\cos(10x_{1})\sin(10x_{2})+\sin(10x_{1}x_{2}), \\
	& x_{1}, x_{2} \in [0, 1].
	\end{aligned}
	\end{split}
	\label{SimuB}
	\end{equation}
	
	The goal is to find the maximal values using the compared methods. For demonstration, the evaluated design points of SeqUD are visualized in Figure~\ref{cliff_Sequd} and Figure~\ref{octopus_Sequd}; the corresponding plots for benchmark methods are placed in Figure~\ref{cliff_Demo} and Figure~\ref{octopus_Demo} in the appendix. Some impressive results are observed. All the non-sequential methods together with SMAC and TPE use many design points in less promising areas in the cliff function. The SeqRand shares the benefits of the space halving strategy and can finally find the global optimal region. However, as randomly generated samples are not representative, SeqRand is less efficient as compared to SeqUD. Similar results could be found in the octopus function. For example, SeqRand, TPE, and GP-EI are trapped in a sub-optimal area; the best location found by SMAC is just close to the global optimum. In contrast, the proposed SeqUD approach tends to be more promising as it successfully finds the correct area and achieves the best performance.  
	
	Table~\ref{simu_example_res} reports the final optimization results of all compared methods, where the best performing methods are highlighted in bold and underlined results are significantly different from the best ones. The performance of sequential methods over the number of runs is visualized in Figure~\ref{simu_example_hist}. In the cliff function, SeqUD uses fewer trials to reach the best point. Given 100 runs, SeqUD performs slightly better than GP-EI, but no statistical significance is observed. In particular, SMAC fails in this task, and its performance is close to random search. For the octopus function, SeqUD achieves significantly better performance as compared to all the benchmarks. SeqRand and GP-EI show similar performance at 100 runs, and both of them outperform SMAC and TPE in this task.
	\afterpage{
		\begin{table}[!t]
			\scriptsize
			\renewcommand\tabcolsep{1.5pt}
			\renewcommand\arraystretch{1}
			\begin{center}
				\begin{tabular}{c|cccccccccc}
					\hline
					Data Set &             Grid              &             Sobol             &              UD               &             Rand              &              LHS              &            SeqRand            &              TPE              &             SMAC              &             GP-EI             &           SeqUD            \\ \hline
					cliff& $\underline{0.869}$ & $\underline{0.877}$ & $\underline{0.983}$ & $\underline{0.907}$$\pm$0.082 & $\underline{0.931}$$\pm$0.063 & $\underline{0.961}$$\pm$0.098 & $\underline{0.973}$$\pm$0.026 & $\underline{0.913}$$\pm$0.102 & 0.994$\pm$0.036 & $\mathbf{1.000}$ \\
					octopus& $\underline{2.889}$ & $\underline{2.778}$ & $\underline{2.849}$ & $\underline{2.784}$$\pm$0.136 & $\underline{2.805}$$\pm$0.132 & $\underline{2.904}$$\pm$0.157 & $\underline{2.858}$$\pm$0.113 & $\underline{2.857}$$\pm$0.163 & $\underline{2.898}$$\pm$0.198 & $\mathbf{2.996}$\\ \hline
				\end{tabular}
			\end{center}
			\caption{The optimization results of cliff and octopus functions. The best performing methods are highlighted in bold and underlined results are significantly different from the best ones. Note that the standard deviations of Grid, Sobol, UD and SeqUD are omitted as they are all zero.} \label{simu_example_res}
		\end{table}
		\begin{figure}[!t]
			\centering
			\subfloat[Cliff]{
				\label{cliff_hist} %% label for second subfigure
				\includegraphics[width=0.5\textwidth]{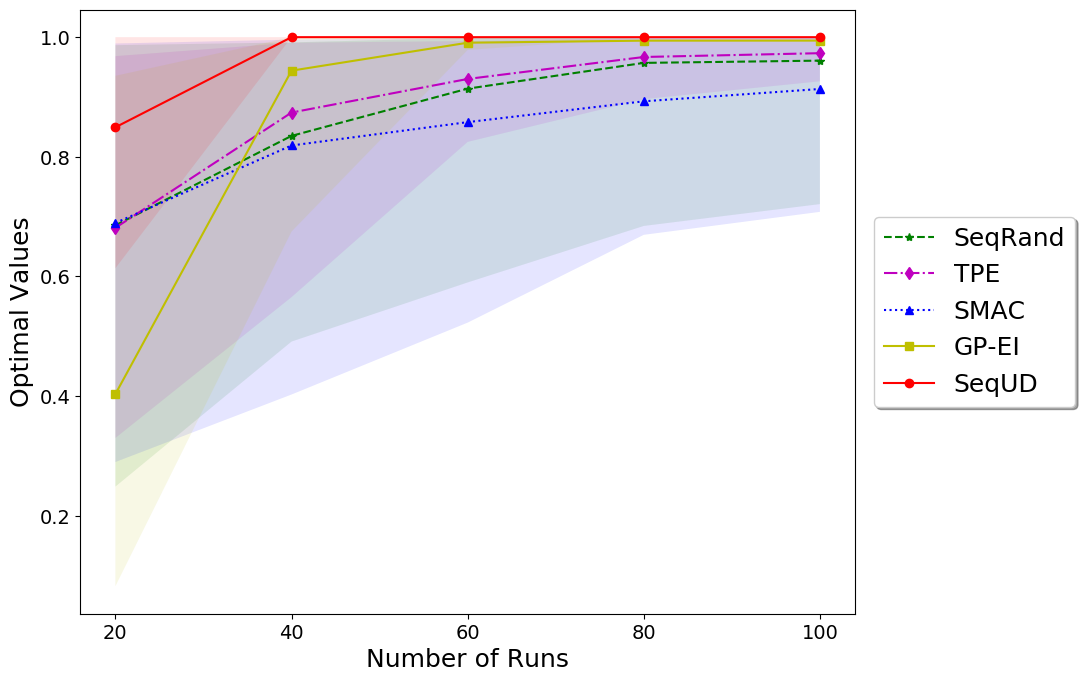}}
			\subfloat[Octopus]{
				\label{octopus_hist} %% label for first subfigure
				\includegraphics[width=0.5\textwidth]{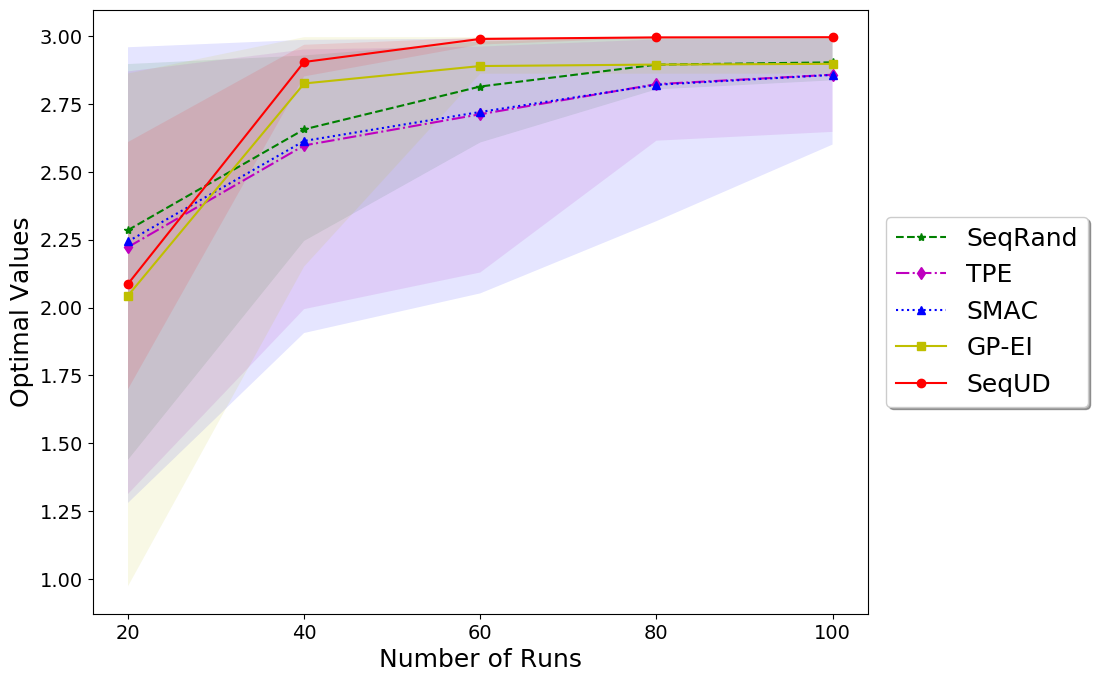}}
			\caption{The optimization results against the number of runs of the two example functions. Each point is averaged over 100 repetitions, and the areas between the 5th and 95th percentiles are shadowed.}
			\label{simu_example_hist} %% label for entire figure
		\end{figure}
	} % afterpage end
	
	\subsection{Systematic Investigation}
	A systematic investigation is conducted on extensive standard unconstrained global optimization tasks summarized by~\citet{simulationlib}. Out of the 47 synthetic functions, 32 are selected using the following rules: a) one-dimensional optimization problems are removed (too simple to use for comparison); b) tasks with global optimal points exactly located at the center of the search spaces are removed due to fairness consideration, as some of the compared methods routinely search the center. Table~\ref{simu_list} provides more information for these selected synthetic functions, which are grouped into six categories with 2 to 8 dimensions. Note that some of the functions are allowed to have various dimensions. We follow their recommended settings for powersum, zakharov, dixonpr, rosen, powell, and michal. For the remaining synthetic functions, a somehow arbitrary setting is used, that is, langer, levy, perm0db, trid, prmdb are set to have two dimensions; schwef and stybtang are set to have six dimensions.
	
	\afterpage{
		\begin{table}[!t]
			\small
			\renewcommand\tabcolsep{1.2pt}
			\renewcommand\arraystretch{1}
			\begin{center}
				\begin{tabular}{c|ccc||c|ccc}
					\hline
					Category                       &      Function Name      &   Abbr   & Dim &                       Category                       &        Function Name         &   Abbr   & Dim \\ \hline
					\multirow{9}*{\minitab[c]{Many \\ Local Minima}}  &   Bukin N. 6   &  bukin6  &  2  &    	\multirow{2}*{\minitab[c]{ Bowl-Shaped}}     & Perm 0, d, $\beta $ & perm0db  &  2  \\
					& Cross-in-Tray  & crossit  &  2  &                                                  &        Trid         &   trid   &  2  \\ \cline{5-8}
					&   Eggholder    &   egg    &  2  & \multirow{3}*{\minitab[c]{Steep Ridges \\ or Drops}} &    De Jong  N. 5     & dejong5  &  2  \\
					&  Holder Table   &  holder  &  2  &                                                  &        Easom         &  easom   &  2  \\
					&   Langermann    &  langer  &  2  &                                                  &     Michalewicz      &  michal  &  5  \\ \cline{5-8}
					&      Levy       &   levy   &  2  &   \multirow{11}*{Other}         &        Beale         &  beale   &  2  \\ 
					&   Levy  N. 13   &  levy13  &  2  &                                                    &       Branin         &  branin  &  2  \\
					&    Schwefel    &  schwef  &  6  &                                                  &      Colville       & colville &  4  \\ 
					&    Shubert     & shubert  &  2  &                                                  &   Goldstein-Price   &  goldpr  &  2  \\ \cline{1-4}
					\multirow{4}*{Plate-Shaped}    &     Booth      &  booth   &  2  &                                                  &    Hartmann 3-D     &  hart3   &  3  \\ 
					&   McCormick    &  mccorm  &  2  &                                                  &    Hartmann 4-D     &  hart4   &  4  \\ 
					&   Power Sum    & powersum &  4  &                                                  &    Hartmann 6-D     &  hart6   &  6  \\
					&    Zakharov    & zakharov &  4  &                                                  &  Perm d, $\beta $   &  permdb  &  2  \\ \cline{1-4}
					\multirow{3}*{Valley-Shaped}    & Six-Hump Camel &  camel6  &  2  &                                                  &       Powell        &  powell  &  4  \\
					&  Dixon-Price   & dixonpr  &  4  &                                                  &       Shekel        &  shekel  &  4  \\ 
					&   Rosenbrock   &  rosen   &  8  &                                                  &   Styblinski-Tang   & stybtang &  6  \\ \hline
				\end{tabular}
			\end{center}
			\caption{The 32 synthetic functions for global minimization.} \label{simu_list}
		\end{table}
		
		\begin{figure}[!t]
			\centering
			\subfloat[Ranks at 100 Runs]{
				\label{simu_rank} %% label for first subfigure
				\includegraphics[width=0.44\textwidth]{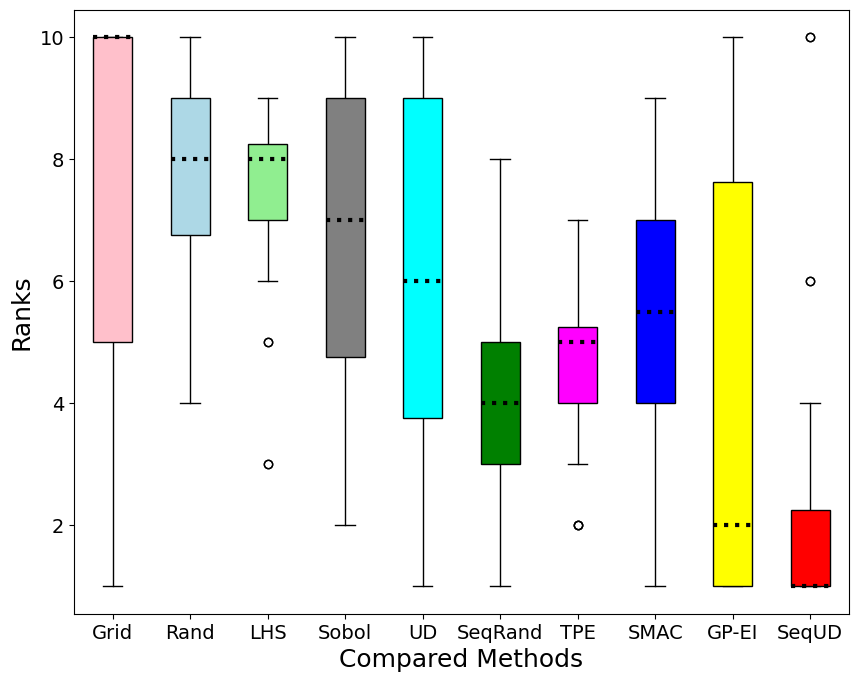}}
			\subfloat[Average Ranks During Optimization]{
				\label{simu_rank_hist} %% label for second subfigure
				\includegraphics[width=0.55\textwidth]{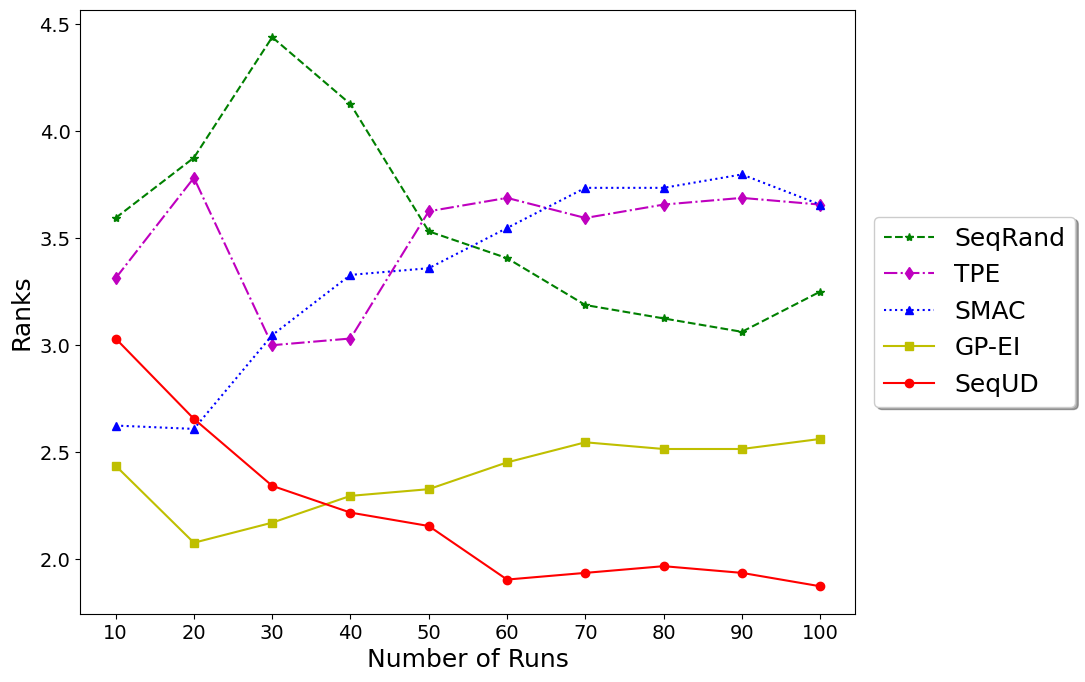}}
			\caption{The ranks of compared methods over the 32 synthetic functions.}
			\label{simu_ranks} %% label for entire figure
		\end{figure}
	} % afterpage end
	
	Similar to Table~\ref{simu_example_res}, the averaged optimization results over 100 repetitions are reported in Table~\ref{SimuRes} in the appendix, in which all the listed results should be multiplied by the corresponding scaling factors in the last column. To have an exact comparison of different methods, we evaluate their relative performance using rank statistics. For each synthetic function, the averaged results of all compared methods are ranked from 1 (the best) to 10 (the worst). A box plot for the rank statistics across the 32 synthetic functions is reported in Figure~\ref{simu_rank}. The five sequential methods are also ranked (from 1 to 5) for every ten runs, and the ranks averaged over the 32 synthetic functions are reported in Figure~\ref{simu_rank_hist}.
	
	From the results, we can observe that sequential methods are superior to non-sequential methods, and among the compared sequential methods, the proposed SeqUD achieves the best overall performance. It ranks the first in more than half of the 32 synthetic functions and the second in 7 functions. The overall performance of GP-EI is inferior to SeqUD but better than other compared methods. Although GP-EI performs the best on 9 synthetic functions, it also ranks the last in 5 functions. The rest sequential methods, i.e., SeqRand, SMAC, and TPE, are slightly poorer than SeqUD and GP-EI.
	
	Note that two failure cases are observed in SeqUD, i.e., for langer and holder. Both of these two functions have a lot of locally optimal regions. It is worth mentioning that SeqRand does not fail in these two functions. That is because design points in SeqUD are constructed with uniformity consideration, such that its design points are relatively stable across different repetitions. In contrast, SeqRand may have very different design points for different random seeds. Therefore, it is possible that SeqRand outperforms SeqUD.
	
	\subsection{Ablation Study for 1000 Runs}
	An ablation study is further conducted with 1000 runs for each compared method. For SeqUD, we increase the number of runs and levels (per stage) proportionally, i.e., $n = q = 150$ for $s \leq 5$ and $n = q = 250$ for $s > 5$. The SeqRand approach is accordingly configured. Given more runs, the design points of GP-EI tend to cluster. Such clustered design points may make GP-EI extremely slow and even crash due to the singular matrix problem. Therefore, we remove GP-EI from the benchmark list, and all the other models are still employed here for comparison.
	
	\afterpage{
		\begin{figure}[!t]
			\centering
			\subfloat[Ranks at 1000 Runs]{
				\label{simu_ablation_rank_1000} %% label for first subfigure
				\includegraphics[width=0.44\textwidth]{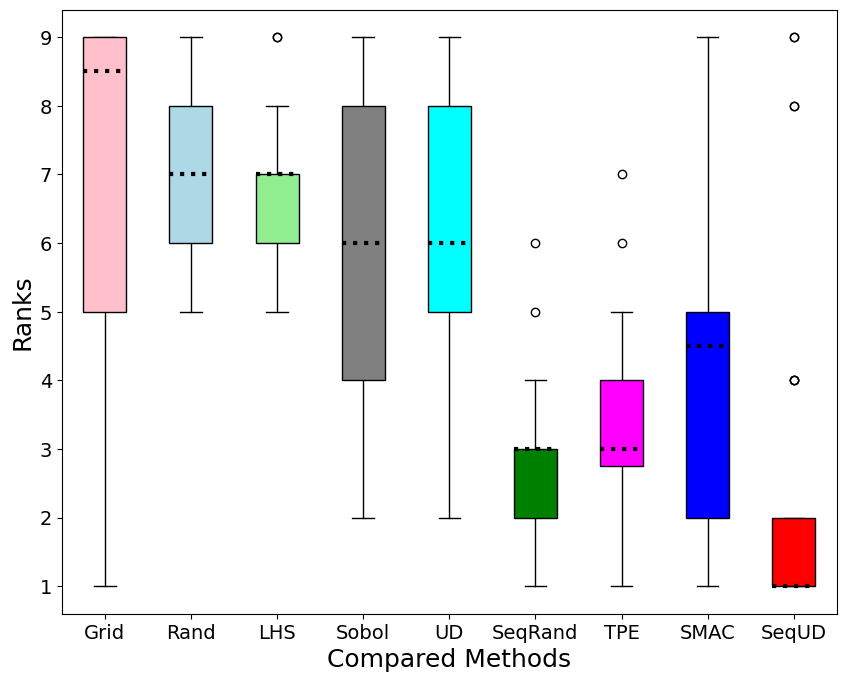}}
			\subfloat[Average Ranks During Optimization]{
				\label{simu_ablation_rank_hist} %% label for second subfigure
				\includegraphics[width=0.55\textwidth]{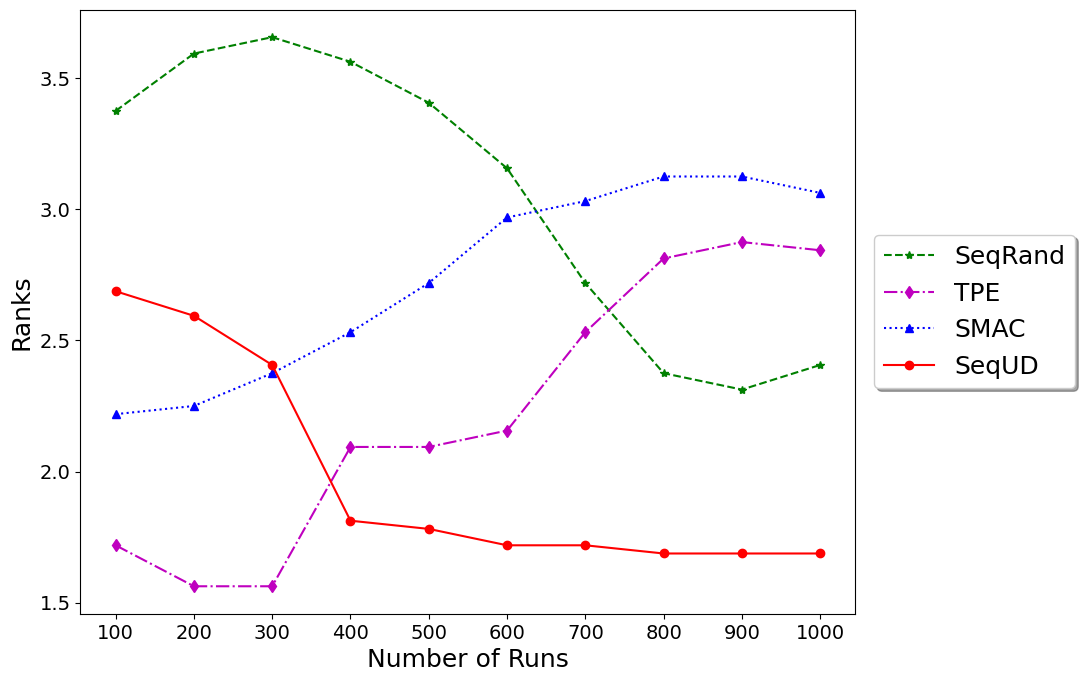}}
			\caption{The ranks of compared methods over the 32 synthetic functions (with 1000 runs).}
			\label{simu_ablation_ranks} %% label for entire figure
		\end{figure}
	}
	
	Table~\ref{SimuAblationRes} in the appendix presents the experimental results of this ablation study, with the same formatting style as in Table~\ref{SimuRes}. We calculate the rank statistics for each synthetic function, and the corresponding box plots are shown in Figure~\ref{simu_ablation_ranks}. The ranks based on 1000 runs are consistent with that of 100 runs. SeqUD still performs the best in most synthetic functions, followed by SeqRand, TPE, SMAC, and all the non-sequential methods. 
	
	In the early stage of the optimization, SeqUD and SeqRand have poorer performance than TPE and SMAC. For instance, both TPE and SMAC are better than SeqUD at 100, 200, and 300 runs. That is because we use larger $n$ in this ablation study. The first 300 runs correspond to the early stages of SeqUD, which means that the search points are of low granularity. In contrast, TPE and SMAC keep their original settings and perform better for the first 300 runs. But as more design points are evaluated, SeqUD gradually exceeds these two Bayesian optimization methods. Consistent patterns are observed for SeqRand, while it needs more design points to exceed TPE and SMAC.
	
	Given a larger number of design points per stage, SeqUD tends to achieve better performance in 29 out of 32 synthetic functions. The results of SeqUD on langer and holder are largely improved, and SeqUD even outperforms all the compared benchmarks at 1000 runs. However, SeqUD also gets poorer results on three synthetic functions. Although using a larger number of design points per stage can have a better exploration of the search space, it is still possible that the initial design points are scattered within locally optimal regions.
	
	\section{Experiments for Hyperparameter Optimization} \label{AutoML_experiments}
	We continue to test the proposed SeqUD method for HPO tasks in the AutoML context.
%in addition to the global optimization tasks. 
	
	\subsection{Experimental Setup} \label{AutoML_experiments_Setup}
	We consider 20 regression and 20 binary classification data sets obtained from the UCI machine learning repository and OpenML platform, in which we select the data with moderate features and sample sizes. Each data is preprocessed by imputing missing values by the median (for continuous variables) or most frequent (for categorical variables) values, as summarized in Table~\ref{data_list}. Before training, categorical features are preprocessed using one-hot encoding, and numerical features are scaled within $[0, 1]$. For each data, we split 50\% of the data samples for training, and the remaining 50\% is used for testing. Five-fold cross-validation (CV) performance in the training set is employed as the optimization target. The root-mean-square error (RMSE) and accuracy score are used as the evaluation criteria for regression and classification tasks, respectively.
	\afterpage{
		\begin{table}[!t]
			\small
			\renewcommand\tabcolsep{2pt}
			\renewcommand\arraystretch{1}
			\begin{center}
				\begin{tabular}{cccc||cccc}
					\hline
					\multicolumn{4}{c||}{Regression}     &      \multicolumn{4}{c}{Classification}       \\ \hline
					Abbr. &      Data Set       & Feature & Size & Abbr. &         Data Set         & Feature & Size \\ \hline
					R1  &        no2         &    7     &   500   &  C1  & breast cancer wisc diag &    30    &   569   \\
					R2  &      sensory       &    11    &   576   &  C2  &    ilpd indian liver    &    10    &   583   \\
					R3  &    disclosure z    &    3     &   662   &  C3  &     credit approval     &    15    &   690   \\
					R4  &   bike share day   &    11    &   731   &  C4  &   breast cancer wisc    &    9     &   699   \\
					R5  &        era         &    4     &  1000   &  C5  &          pima           &    8     &   768   \\
					R6  &      treasury      &    15    &  1049   &  C6  &       tic tac toe       &    9     &   958   \\
					R7  &      airfoil       &    5     &  1503   &  C7  &  statlog german credit  &    24    &  1000   \\
					R8  &      wine red      &    11    &  1599   &  C8  &           pc1           &    21    &  1109   \\
					R9  &    skill craft     &    18    &  3395   &  C9  &      seismic bumps      &    15    &  2584   \\
					R10  &      abalone       &    8     &  4177   & C10  &          churn          &    20    &  5000   \\
					R11  &  parkinsons tele   &    19    &  5875   & C11  &         banana          &    2     &  5300   \\
					R12  &        wind        &    14    &  6574   & C12  &         twonorm         &    20    &  7400   \\
					R13  &     cpu small      &    12    &  8192   & C13  &        ringnorm         &    20    &  7400   \\
					R14  &      topo 2 1      &   266    &  8885   & C14  &           jm1           &    21    &  10885  \\
					R15  &        combined cycle power plant        &    4     &  9568   & C15  &      eeg eye state      &    14    &  14980  \\
					R16  &  electrical grid   &    11    &  10000  & C16  &          magic telescope          &    10    &  19020  \\
					R17  &      ailerons      &    40    &  13750  & C17  &          adult          &    14    &  32561  \\
					R18  &     elevators      &    18    &  16599  & C18  &          nomao          &   118    &  34465  \\
					R19  &  bike share hour   &    12    &  17379  & C19  &          bank marketing         &    16    &  45211  \\
					R20  & california housing &    8     &  20640  & C20  &       electricity       &    8     &  45312  \\ \hline
				\end{tabular}
			\end{center}
			\caption{The data sets for testing different HPO methods.} \label{data_list}
		\end{table}
	} % end of afterpage
	
	Two representative machine learning algorithms are first involved, i.e., support vector machine (SVM) and extreme gradient boosting (XGBoost). A pipeline optimization task is also considered, which involves data preprocessing, feature engineering, model selection, and HPO. Some of the compared methods are implemented to utilize dependence information among hyperparameters, while the rest are not. To eliminate this influence, the dependence information of hyperparameters is not utilized throughout. All the experiments are conducted based on the \textsl{scikit-learn} platform and related packages (e.g., \textsl{xgboost}).
	
	\medbreak
	\textbf{SVM}. We consider a classical 2-D hyperparameters optimization problem in SVM. The popular Gaussian kernel is fixed, and we tune the two continuous hyperparameters, i.e., the kernel parameter and penalty parameter. They are optimized in the base-2 log scale within $[2^{-16}, 2^{6}]$ and $[2^{-6}, 2^{16}]$, respectively. The training algorithm of SVM is not scalable even for data with moderate sample sizes. To save computing time, we instead use \textsl{scikit-learn}'s ``SGDRegressor'' or ``SGDClassifier'' (with hinge loss for regression and epsilon insensitive loss for classification) with the ``Nystroem'' transformer. Here we use this approach to handle data with more than 3000 samples. The number of components in the ``Nystroem'' transformer is fixed to 500. The initial learning rate is set to 0.01, and we use the ``adaptive'' optimizer to adjust the learning rate during optimization.
	
	\medbreak
	\textbf{XGBoost}. Hyperparameter optimization of XGBoost is much more complicated than that of SVM. Eight important hyperparameters in XGBoost are introduced, including booster (categorical; ``gbtree'' or ``gblinear''), maximum tree depth (integer-valued; within the range $[1, 8]$), number of estimators (integer-valued; within the range $ [100, 500] $), ratio of features in each tree (continuous; within the range $ [0.5, 1]$), learning rate (continuous; the base-10 log scale within the range $ [10^{-5},10^{0}] $), minimum loss reduction (continuous; the base-10 log scale within the range $ [10^{-5},10^{0}] $), $\ell_{1}$-regularization (continuous; the base-10 log scale within the range $ [10^{-5},10^{0}] $) and $\ell_{2}$-regularization (continuous; the base-10 log scale within the range $ [10^{-5},10^{0}] $). 
	
	\medbreak
	\textbf{Pipeline Optimization}. In addition to optimizing a single machine model's hyperparameters, we move one step further to the challenging pipeline optimization task. In particular, we consider data preprocessing (``MinmaxScaler'' and ``Standardizer''), feature engineering (All Features, ``SelectKBest'', and ``PCA''), model selection (SVM and XGBoost) and HPO for the selected model. Each of the first three steps, i.e., data preprocessing, feature engineering, and model selection, can be treated as a categorical hyperparameter. In data preprocessing, ``MinmaxScaler'' linearly maps each feature within $[0, 1]$; ``Standardizer'' instead standardizes each feature with zero mean and unit variance.  While for feature engineering, ``SelectKBest'' selects the top-$K$ features with the highest F-values. We tune $K$ within $[1, \min\{m, 20\}]$ ($m$ denotes the number of features after one-hot encoding); similarly, ``PCA'' denotes the principal component analysis, and the number of principal components is selected within $[1, \min\{m, 20\}]$. For the selected machine learning model (either SVM or XGBoost), we use their corresponding HPO configurations as mentioned above.
	
	\bigskip
	In total, six groups of tasks are involved, i.e., SVM-Regression (SVM-Reg), SVM-Classification (SVM-Cls), XGBoost-Regression (XGB-Reg), XGBoost-Classification (XGB-Cls), Pipeline-Regression (Pipe-Reg), and Pipeline-Classification (Pipe-Cls). The same settings are used for the compared HPO methods as in Section~\ref{Synthetic_experiments}. That is, all the compared methods are allowed to evaluate at most 100 hyperparameter configurations. For SeqUD and SeqRand, we use $n = q = 15$ when $s \leq 5$ and $n = q = 25$ when  $s > 5$. The optimization results regarding the 5-fold CV performance, computing time, and test set performance are all recorded. Each experiment is repeated ten times. The final results are reported with average, standard deviation, and statistical significance across the ten repetitions.

	\subsection{Results Analysis}
	The 5-fold CV and test set results of HPO experiments on the 40 data sets are reported in the appendix. Bold numbers indicate the best-performing methods, and results that are significantly different from the best are underlined. Note that the RMSE results should be multiplied by the corresponding scaling factors in the last column. We analyze the results from the following perspectives.
	
\begin{figure}[!t]
		\centering
		\subfloat[SVM-Reg]{
			\label{reg_svm_rank_cv} %% label for first subfigure
			\includegraphics[width=0.33\textwidth]{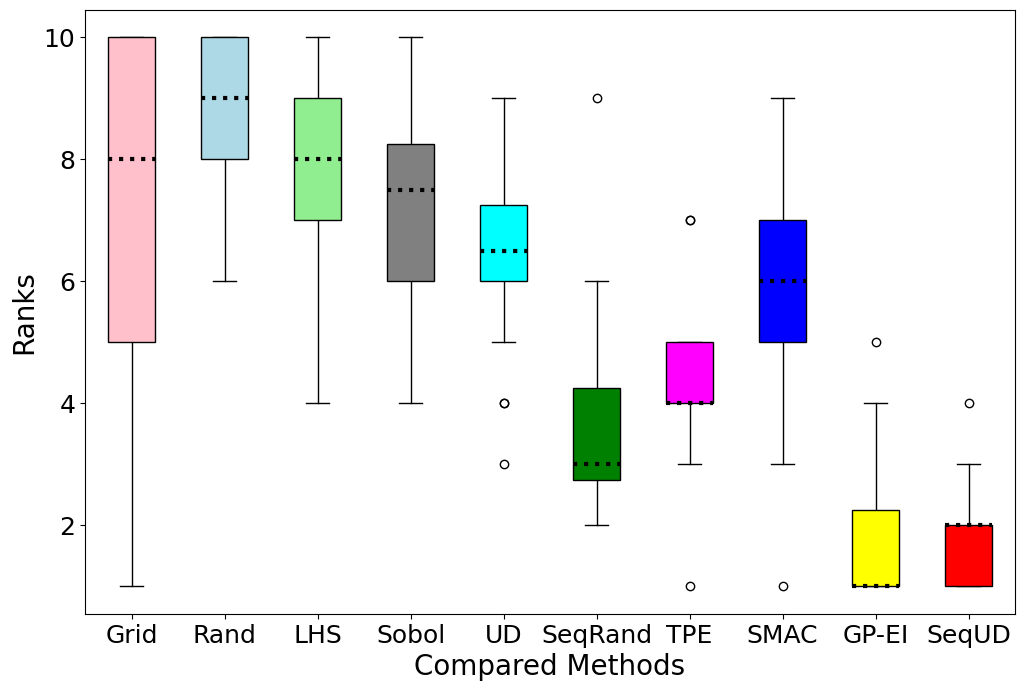}}
		\subfloat[XGB-Reg]{
			\label{reg_xgb_rank_cv} %% label for first subfigure
			\includegraphics[width=0.33\textwidth]{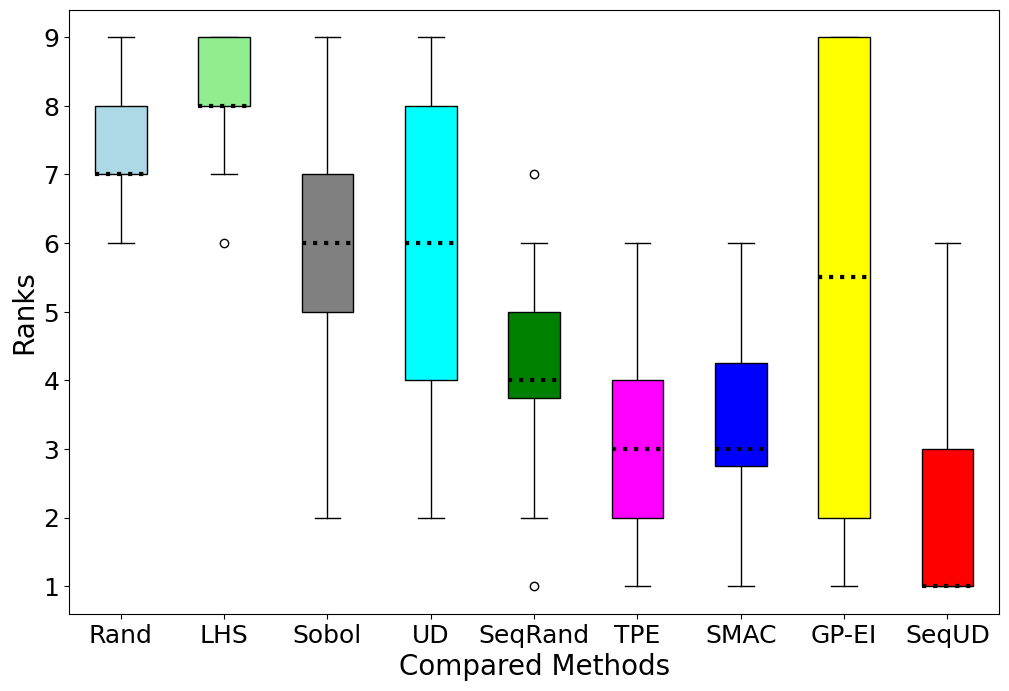}}
		\subfloat[Pipe-Reg]{
			\label{reg_pipe_rank_cv} %% label for first subfigure
			\includegraphics[width=0.33\textwidth]{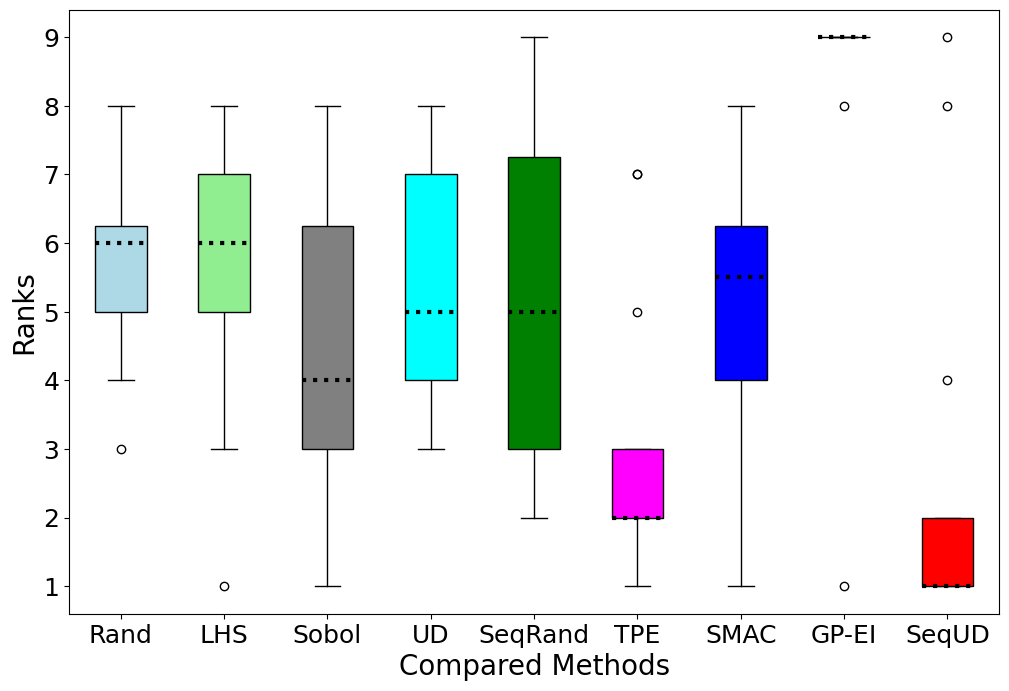}}\\
		\subfloat[SVM-Cls]{
			\label{cls_svm_rank_cv} %% label for first subfigure
			\includegraphics[width=0.33\textwidth]{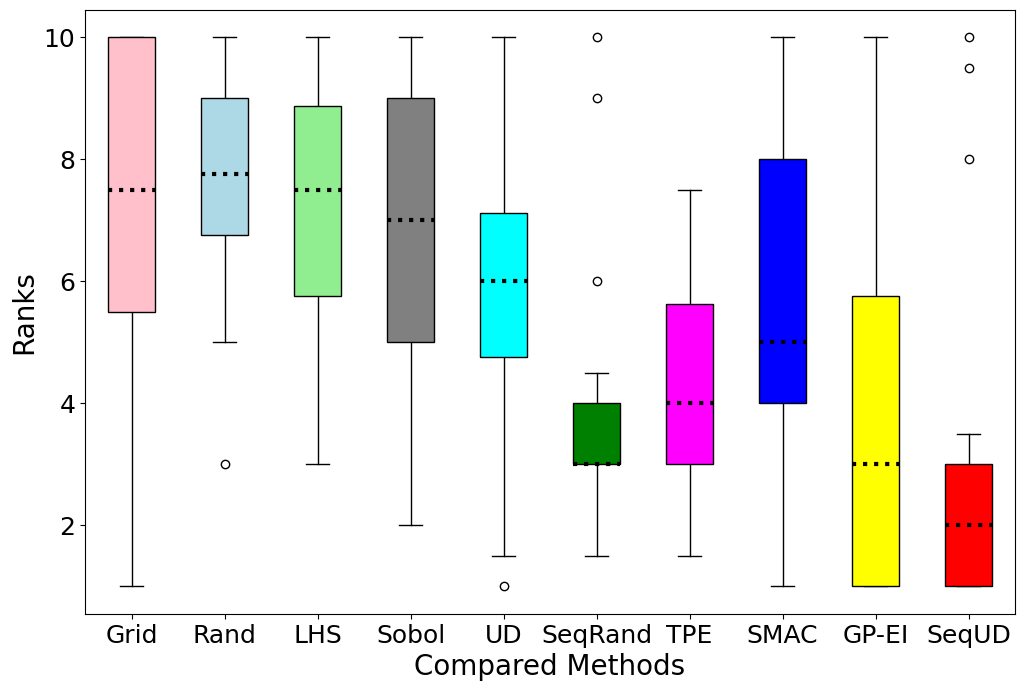}}
		\subfloat[XGB-Cls]{
			\label{cls_xgb_rank_cv} %% label for first subfigure
			\includegraphics[width=0.33\textwidth]{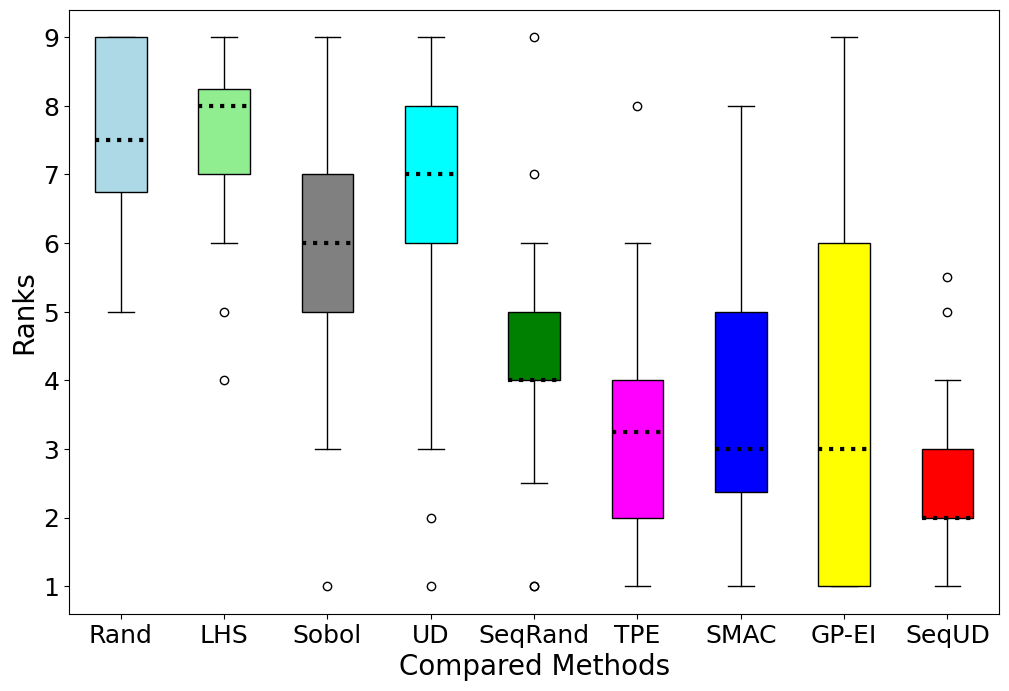}}
		\subfloat[Pipe-Cls]{
			\label{cls_pipe_rank_cv} %% label for first subfigure
			\includegraphics[width=0.33\textwidth]{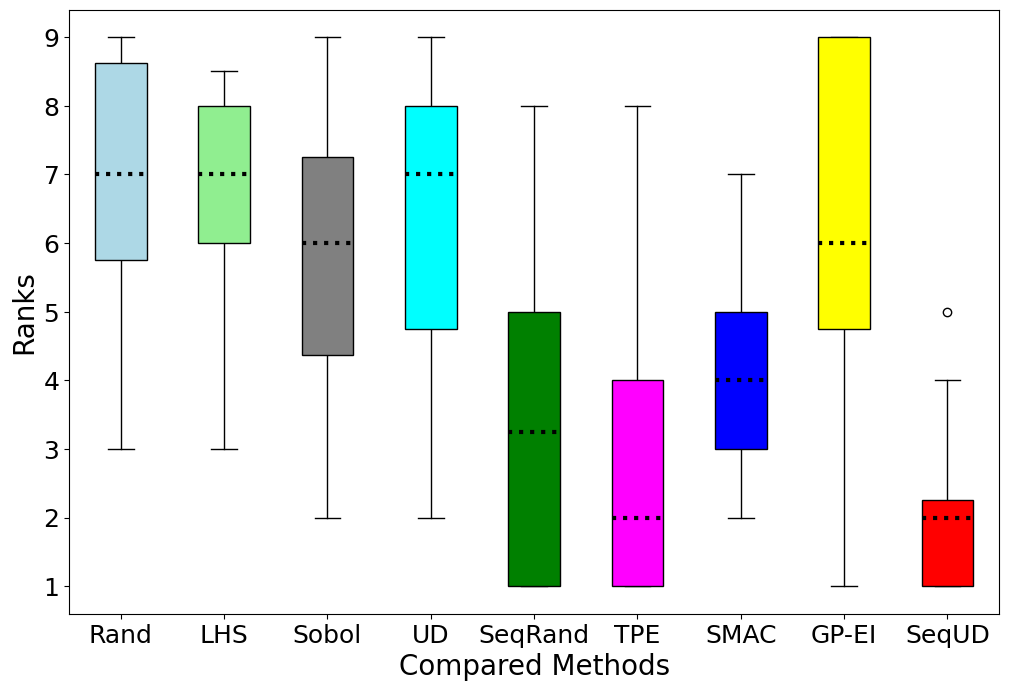}} 
		\caption{The ranks of 5-fold CV performance across different tasks. Each sub-figure represents one of the six tasks, and the boxes are drawn based on the averaged results of corresponding 20 data sets.}
		\label{cv_rank} %% label for entire figure
	\end{figure}

	\medskip
	\textbf{Five-fold CV Performance}. As the optimization target, 5-fold CV performance directly reflects the optimization capability of each compared method. To make a clear comparison, we rank the methods according to their averaged results for each data, as shown in Figure~\ref{cv_rank}. A one-verse-one win/loss comparison for all the methods is further presented in Table~\ref{cv_comp}, which is summarized over the 120 tasks (40 data sets by 3 machine learning models). For instance, in the cell for ``Rand''-``Grid'', ``19 (7)'' denotes that random search is better than grid search on 19 tasks in which 7 tasks are tested to be significantly better. Similarly, random search also shows inferior performance to grid search on 21 tasks, with 9 tasks being significantly worse. Note that grid search is applied only for 2-D synthetic functions, hence the total number of compared tasks is 40. 
	
	Several interesting findings are observed. First, the uniformity of design points is positively related to the performance of non-sequential methods. With better uniformity, both UD and Sobol show slightly better performance than random search and LHS, but no big difference is observed between UD and Sobol. Given 100 runs, both UD and Sobol have better uniformity performance than random search and LHS. For instance, in the case of SVM with $s=2$, the best uniformity among non-sequential methods is achieved by UD ($\mbox{CD}_{2}=0.000035$), followed by Sobol ($\mbox{CD}_{2}=0.000142$), LHS ($\mbox{CD}_{2}=0.000340$), and random search ($\mbox{CD}_{2}=0.003440$).
	
	Second, sequential methods are in general superior to non-sequential methods. 	SeqRand outperforms random search on 106 out of the 120 tasks. The proposed SeqUD also significantly beats the non-sequential UD on 109 out of the 120 tasks. The other sequential methods, including TPE, SMAC, and GP-EI, also exhibit competitive performances.
	
	Third, by incorporating uniform design and sequential optimization, the proposed SeqUD achieves the best overall performance among all the compared methods. Given the proposed sequential strategy, SeqUD significantly improves over its na\"ive baseline SeqRand. The SeqUD results are better than or on par with its counterpart Bayesian optimization methods in most cases. GP-EI is the second-best among all the tested methods. The GP-EI method works well on low-dimensional tasks (i.e., SVM), but it is less efficient for high-dimensional tasks (i.e., XGBoost and pipeline optimization). On the contrary, TPE and SMAC are more robust for high-dimensional tasks.
	
	\begin{table}[!t]
		\small	
		\renewcommand\tabcolsep{1.5pt}
		\renewcommand\arraystretch{1}
		\begin{center}
			\begin{tabular}{c|cccccccccc}
				\hline
				&   Grid   &   Rand    &    LHS    &   Sobol   &    UD     & SeqRand  &   TPE    &   SMAC   &  GP-EI   &  SeqUD   \\ \hline
				Grid   &    -     &  21  (9)  &  19  (6)  & 19  (12)  &  16  (9)  &  7  (2)  &  9  (1)  & 12  (2)  &  6  (2)  &  5  (2)  \\
				Rand   & 19  (7)  &     -     &  57  (0)  &  32  (4)  &  39  (4)  & 14  (1)  &  9  (1)  & 17  (1)  & 39  (15) &  7  (2)  \\
				LHS   & 20  (10) &  61  (8)  &     -     &  34  (4)  &  40  (7)  & 20  (0)  &  7  (0)  & 23  (1)  & 43  (14) &  6  (3)  \\
				Sobol  & 21  (11) & 86  (30)  & 84  (21)  &     -     & 61  (24)  & 29  (4)  & 17  (2)  & 36  (4)  & 52  (19) & 10  (4)  \\
				UD    & 23  (13) & 80  (26)  & 79  (18)  & 59  (22)  &     -     & 30  (6)  & 21  (3)  & 41  (3)  & 46  (20) & 10  (3)  \\
				SeqRand & 33  (21) & 106  (54) & 98  (59)  & 90  (49)  & 89  (50)  &    -     & 49  (5)  & 65  (16) & 66  (19) & 26  (4)  \\
				TPE   & 31  (17) & 109  (71) & 112  (68) & 102  (60) & 95  (63)  & 69  (10) &    -     & 82  (18) & 70  (25) & 32  (7)  \\
				SMAC   & 27  (12) & 103  (45) & 95  (51)  & 82  (46)  & 79  (41)  & 54  (8)  & 38  (5)  &    -     & 61  (24) & 20  (5)  \\
				GP-EI  & 34  (27) & 81  (49)  & 75  (46)  & 66  (49)  & 73  (43)  & 54  (30) & 50  (31) & 57  (29) &    -     & 34  (18) \\
				SeqUD  & 35  (28) & 113  (83) & 114  (81) & 109  (80) & 109  (81) & 91  (36) & 87  (39) & 97  (47) & 85  (31) &    -     \\ \hline
			\end{tabular}
		\end{center}
		\caption{Pairwise win/loss over the 120 HPO tasks (5-fold CV): the numbers in each cell indicate how often the method in row (significantly) outperforms the method in column. The statistical significance is calculated by paired t-test with a significance level of 0.05.} \label{cv_comp}
	\end{table}
	
\begin{figure}[!t]
		\centering
		\subfloat[SVM-Reg]{
			\label{reg_svm_time} %% label for first subfigure
			\includegraphics[width=0.33\textwidth]{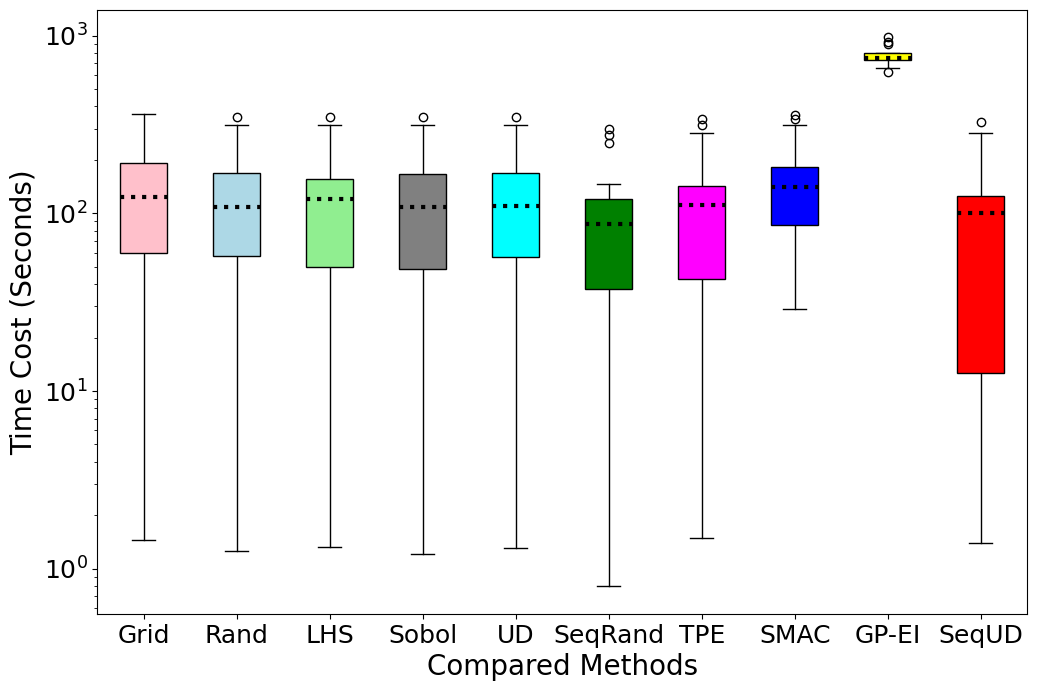}}
		\subfloat[XGB-Reg]{
			\label{reg_xgb_time} %% label for first subfigure
			\includegraphics[width=0.33\textwidth]{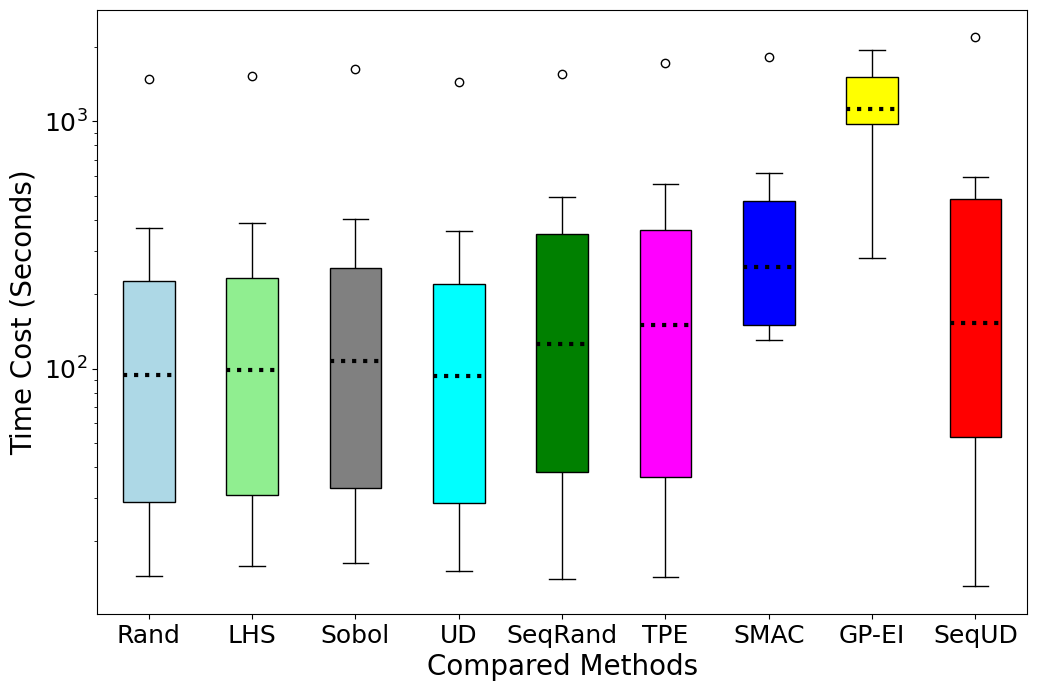}}
		\subfloat[Pipe-Reg]{
			\label{reg_pipe_time} %% label for first subfigure
			\includegraphics[width=0.33\textwidth]{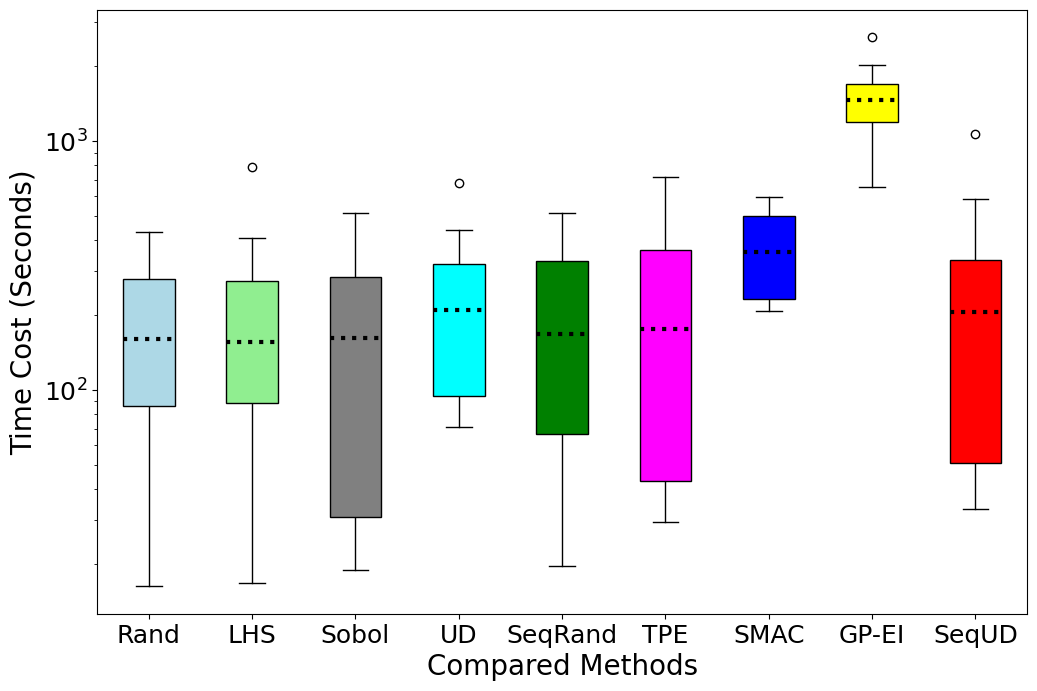}}\\
		\subfloat[SVM-Cls]{
			\label{cls_svm_time} %% label for second subfigure
			\includegraphics[width=0.33\textwidth]{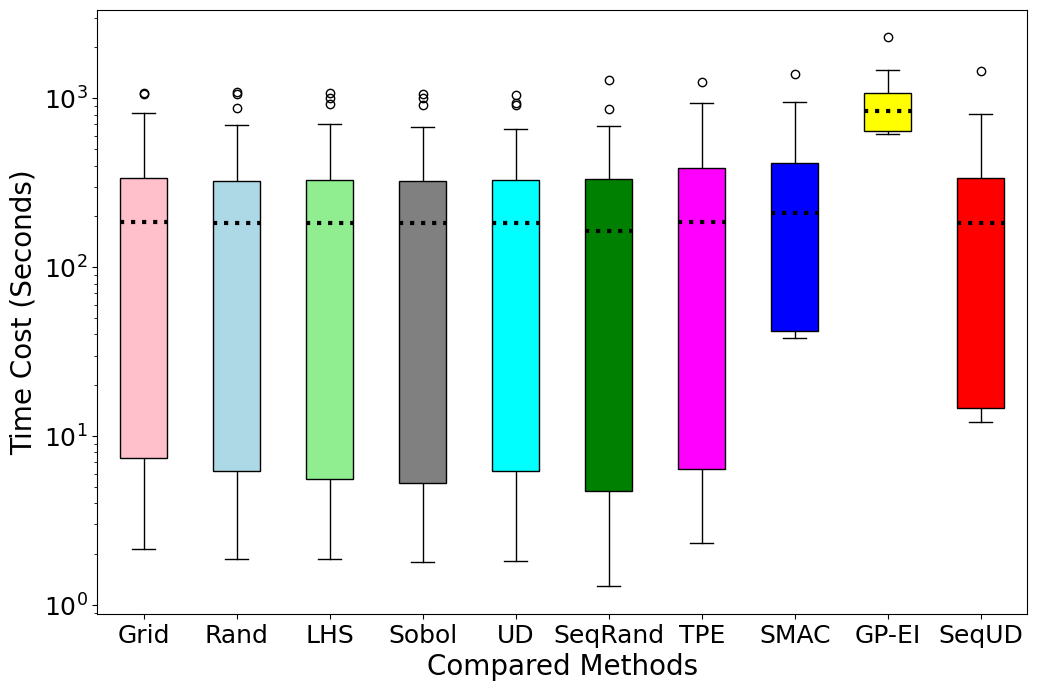}}
		\subfloat[XGB-Cls]{
			\label{cls_xgb_time} %% label for second subfigure
			\includegraphics[width=0.33\textwidth]{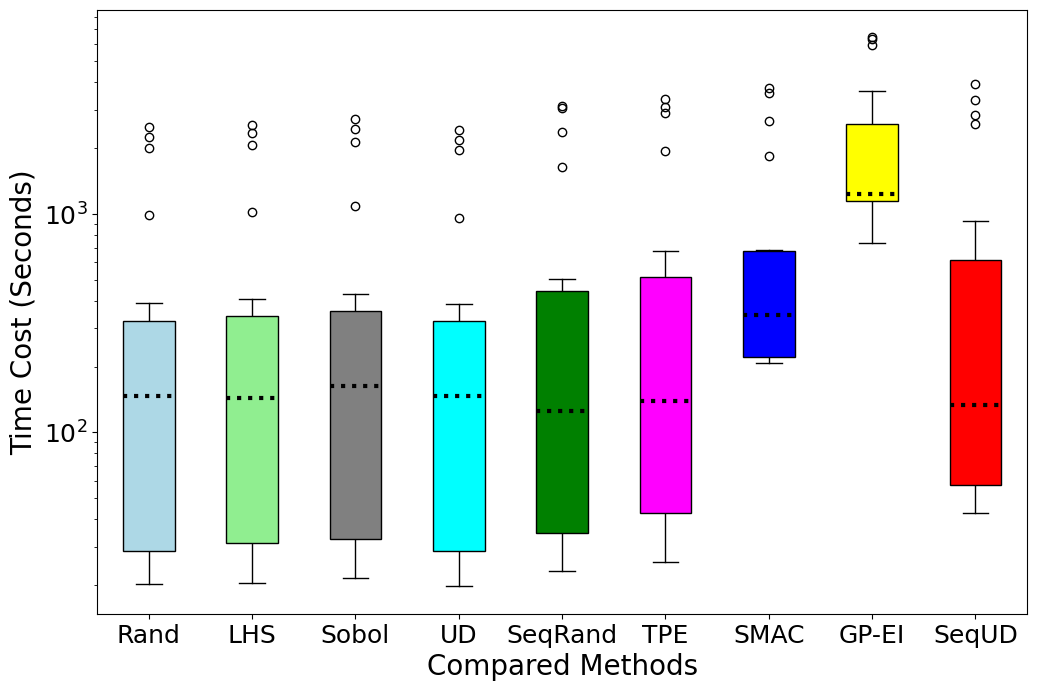}}
		\subfloat[Pipe-Cls]{
			\label{cls_pipe_time} %% label for second subfigure
			\includegraphics[width=0.33\textwidth]{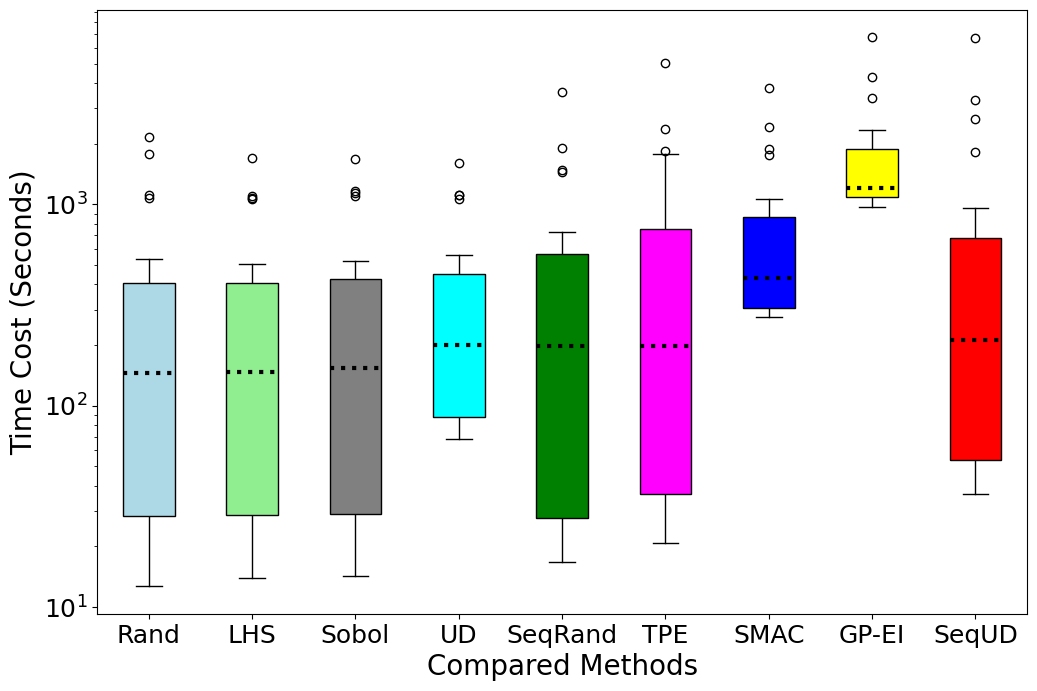}}
		\caption{Computing time comparison across different tasks. Each sub-figure represents one of the six tasks, and the boxes are drawn based on the averaged results of corresponding 20 data sets.}
		\label{time_cost} %% label for entire figure
	\end{figure}

	\medskip
	\textbf{Computational Cost}. The computing time required by each method is reported in Figure~\ref{time_cost}. In SeqUD, augmented design points can be efficiently generated within a few seconds. This is negligible as compared to the computational complexity of training a machine learning model. Therefore, the time complexity of SeqUD is close to that of simple non-sequential methods like grid search and random search.  Moreover, the proposed SeqUD, SeqRand, and all the non-sequential methods can be further accelerated by performing design point-level parallelization. This type of parallelization will not sacrifice the optimization performance. However, Bayesian optimization methods are less efficient regarding computing time. Both TPE and SMAC are fast for optimizing 2-D SVM tasks; while for high-dimensional tasks, SMAC and GP-EI would need more time for surrogate modeling and acquisition optimization.
	
	It is observed that SeqUD often runs faster than grid search and random search. 	First, the time cost of generating (augmented) uniform designs using AugUD is rather small. In most cases, it can be finished within a few seconds, which is negligible as compared to the training cost of machine learning models. Second, the training time of a machine learning model depends on specific hyperparameter settings. In grid search and random search, they may generate many design points on time-consuming hyperparameter configurations; while for SeqUD, it focuses more on the best performing regions, which could be less computationally intensive.

\begin{figure}[!t]
		\centering
		\subfloat[SVM-Reg]{
			\label{reg_svm_rank_test} %% label for second subfigure
			\includegraphics[width=0.33\textwidth]{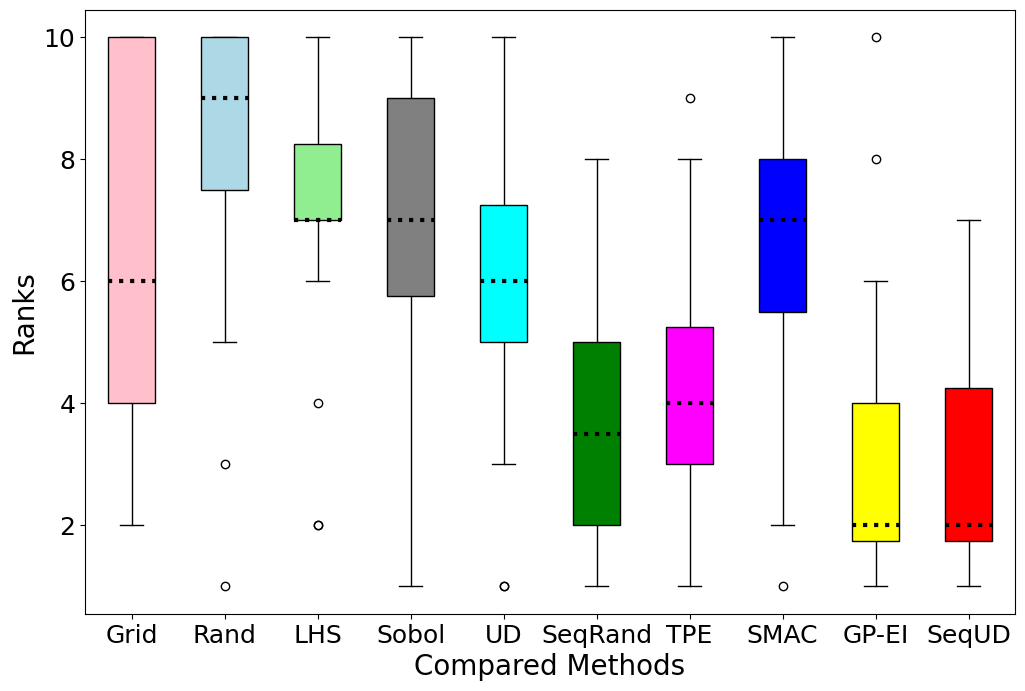}}
		\subfloat[XGB-Reg]{
			\label{reg_xgb_rank_test} %% label for second subfigure
			\includegraphics[width=0.33\textwidth]{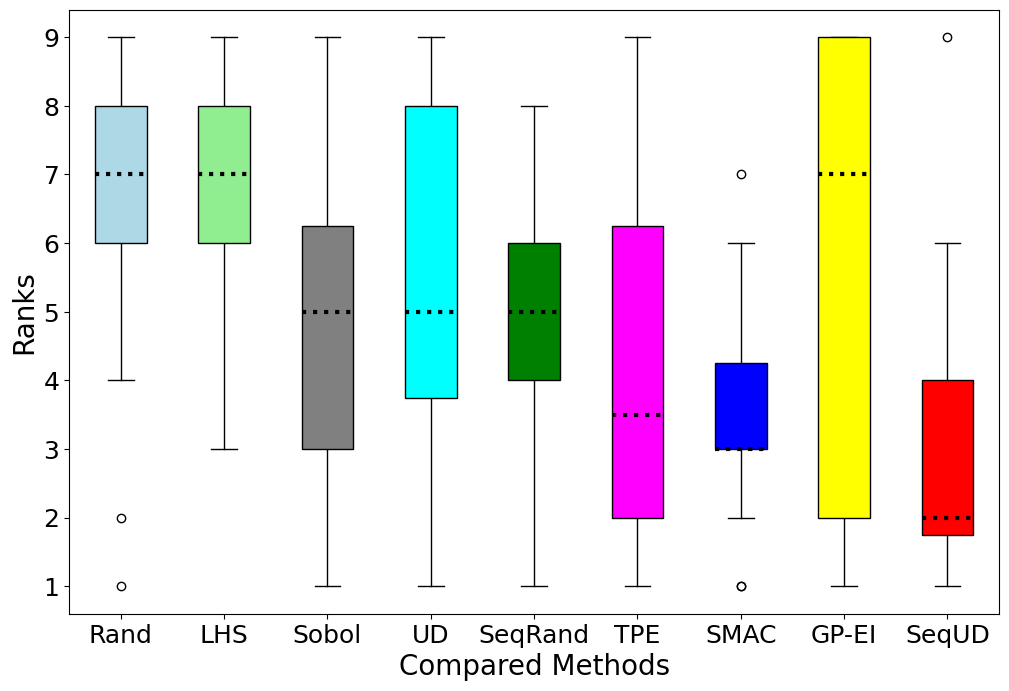}}
		\subfloat[Pipe-Reg]{
			\label{reg_pipe_rank_test} %% label for second subfigure
			\includegraphics[width=0.33\textwidth]{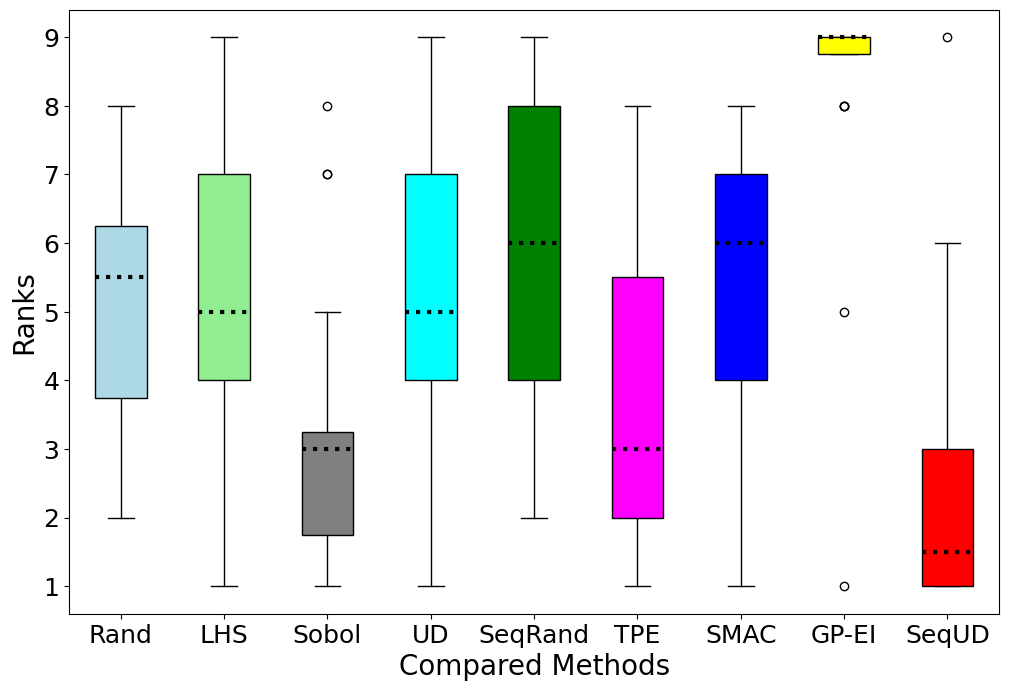}}\\
		\subfloat[SVM-Cls]{
			\label{cls_svm_rank_test} %% label for second subfigure
			\includegraphics[width=0.33\textwidth]{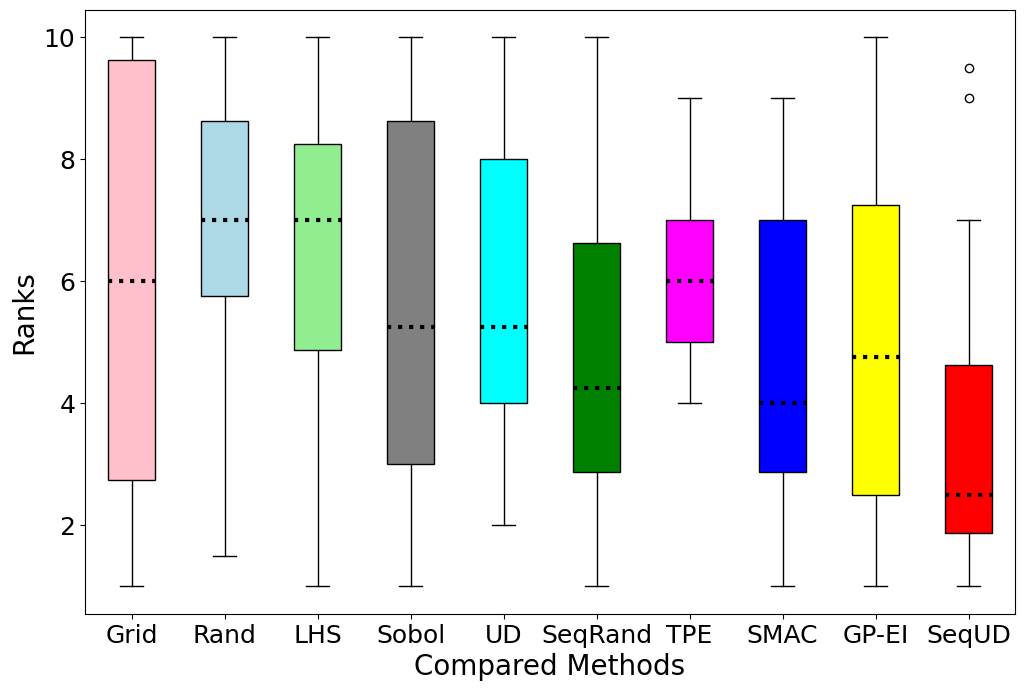}}
		\subfloat[XGB-Cls]{
			\label{cls_xgb_rank_test} %% label for second subfigure
			\includegraphics[width=0.33\textwidth]{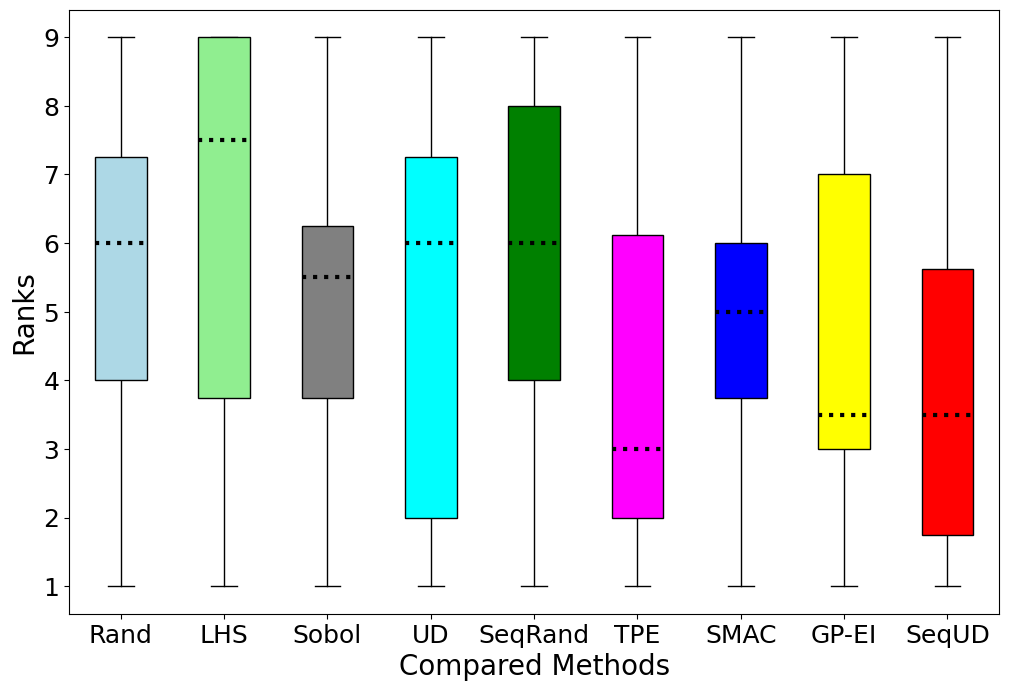}}
		\subfloat[Pipe-Cls]{
			\label{cls_pipe_rank_test} %% label for second subfigure
			\includegraphics[width=0.33\textwidth]{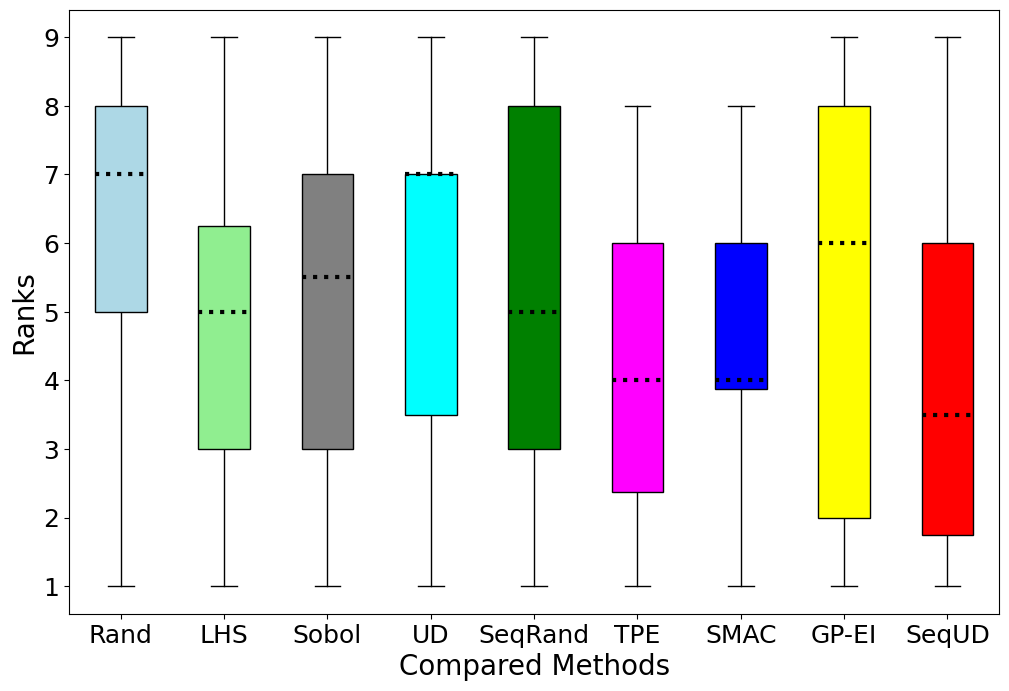}}
		\caption{The ranks of test set performance across different tasks. Each sub-figure represents one of the six tasks, and the boxes are drawn based on the averaged results of corresponding 20 data sets.}
		\label{test_rank} %% label for entire figure
	\end{figure}

	\medskip
	\textbf{Test Set Performance}. Figure~\ref{test_rank} reports the ranks of test set performance achieved by different methods. Table~\ref{test_comp} provides the one-verse-one win/loss comparison results.
	
	Hyperparameter configurations that achieve better 5-fold CV performance can usually perform well on the test set. The Spearman rank-order correlation coefficient is calculated between SeqUD's 5-fold CV and test set performance (over the 120 tasks), which are shown positively correlated with a correlation coefficient greater than 0.6. The proposed SeqUD achieves not only superior 5-fold CV performance but also competitive test set performance. For instance, SeqUD outperforms each of GP-EI, SMAC, and TPE 79 times. The superiority of SeqUD over other non-sequential methods is much more significant. Therefore, it is evident that the proposed SeqUD approach is a competitive HPO method.
	
	Despite the positive relationship between 5-fold CV and test set performance, there exist cases that are not consistent. For example, sequential methods generally perform much better than non-sequential methods regarding 5-fold CV performance, while for test set performance, the gap is reduced. Such an observation can be due to overfitting. It is known that when there are a limited number of validation samples, the hyperparameters can overfit the validation set. However, the best hyperparameter configuration on the validation set does not necessarily generalize well for the test set~\citep{hutter2019automated}. Some possible solutions have been proposed in the literature to alleviate this problem, e.g., data reshuffling for each function evaluation~\citep{levesque2018bayesian}, finding stable optima instead of sharp optima~\citep{dai2017stable}, and the ensemble of multiple good hyperparameter configurations~\citep{momma2002pattern}. However, overfitting is still an open question in HPO, which is worthy of further investigation.
	
	\begin{table}[!t]
		\small	
		\renewcommand\tabcolsep{1.5pt}
		\renewcommand\arraystretch{1}
		\begin{center}
			\begin{tabular}{c|cccccccccc}
				\hline
				&Grid&Rand&LHS&Sobol&UD&SeqRand&TPE&SMAC&GP-EI&SeqUD\\ \hline
				Grid&-& 20  (7) & 18  (10) & 19  (9) & 19  (4) & 16  (4) & 13  (2) & 19  (2) & 12  (2) & 10  (2) \\
				Rand& 18  (4) &-& 55  (4) & 40  (2) & 49  (9) & 40  (1) & 31  (3) & 34  (3) & 56  (14) & 24  (3) \\
				LHS& 21  (4) & 65  (5) &-& 37  (5) & 49  (10) & 38  (4) & 40  (0) & 42  (2) & 58  (19) & 24  (4) \\
				Sobol& 18  (7) & 77  (19) & 80  (21) &-& 64  (25) & 57  (9) & 44  (7) & 62  (7) & 61  (19) & 30  (9) \\
				UD& 19  (6) & 69  (18) & 69  (14) & 53  (18) &-& 54  (9) & 42  (5) & 53  (7) & 63  (19) & 35  (5) \\
				SeqRand& 24  (7) & 79  (23) & 80  (18) & 62  (16) & 64  (19) &-& 54  (8) & 56  (7) & 58  (17) & 32  (5) \\
				TPE& 24  (7) & 86  (30) & 79  (29) & 74  (29) & 75  (28) & 64  (9) &-& 67  (6) & 64  (15) & 39  (7) \\
				SMAC& 21  (7) & 86  (20) & 76  (20) & 56  (24) & 65  (17) & 62  (6) & 47  (4) &-& 67  (21) & 35  (7) \\
				GP-EI& 28  (13) & 64  (26) & 60  (27) & 58  (25) & 56  (23) & 59  (12) & 54  (11) & 53  (14) &-& 41  (8) \\
				SeqUD& 28  (12) & 94  (43) & 93  (43) & 87  (36) & 83  (36) & 86  (20) & 79  (13) & 79  (22) & 79  (24) &-\\
				\hline
			\end{tabular}
		\end{center}
		\caption{Pairwise win/loss over the 120 HPO tasks (test set): the numbers in each cell indicate how often the method in row (significantly) outperforms the method in column. The statistical significance is calculated by paired t-test with a significance level of 0.05.}         \label{test_comp}
	\end{table}
	
	\medbreak
	\textbf{Possible Limitations}. % of our experimental evaluation
	Although the proposed SeqUD achieves promising results on multiple HPO tasks, the experimental evaluation can be possibly limited and may lead to a biased conclusion.
	First, as the computing resources are limited, this paper only considers two machine learning tasks, three machine learning models, and small and medium-sized data sets. Such empirical results may not generalize universally on other machine learning tasks. Besides, the hyperparameter search space is according to our experience (as well as common practice), but it does not represent all possible hyperparameter settings.

	Second, the choice of evaluation metric for regression or classification tasks may affect the final result. In our experiments, it is observed that SeqUD performs slightly better on regression tasks than classification tasks. As SVM can only output binary prediction, we used the accuracy score metric (a discrete measurement) in classification tasks. It is possible that the difference between tested methods cannot be fully revealed. However, for regression tasks, the MSE metric is a continuous measure, under which the superiority of SeqUD has been shown more obvious.	
	
	\subsection{Comparison with Hyperband}
	Additional experiments are conducted to compare SeqUD with Hyperband.  Hyperband is an adaptive computing resource allocation algorithm based on random search. It is mainly used to tune the models that can be iteratively trained, where the performance of different trials can be evaluated during training, such that the worst half trials are early stopped. For instance, hyperparameters of deep neural networks can be tuned by Hyperband, as they are typically fitted by the iterative backpropagation algorithm. In this paper, we compare Hyperband with SeqUD using the XGBoost model, as the amount of computing resources in XGBoost can be determined by the number of estimators. In Hyperband, the search space of hyperparameters is kept the same as that of SeqUD, except for the number of estimators, which is instead used as the indicator of amount of resources. 
	
	In SeqUD, the optimization is terminated as the maximal number of runs is reached. But for Hyperband, the termination is not controlled by the maximal number of runs, but by ``min\_iter'' (the minimum amount of resource), ``max\_iter'' (the maximum amount of resource), and ``eta'' (inverse of the proportion of configurations that are discarded). For a fair comparison, we set ``min\_iter''=10, ``max\_iter''=500, and ``eta''=2. Under this setting, the total number of runs of Hyperband is 138, which is larger than that of SeqUD (100). 
	
	The test set performance is reported in Table~\ref{hb}, and the proposed SeqUD shows significantly better performance than Hyperband on the tested tasks. The reason is that Hyperband still uses random search as the core algorithm for optimization, and its main advantage lies in the adaptive resource allocation. However, it is still hard to observe the superior performance of Hyperband, even given more trials. 
	
	\begin{table}[!htp]
		\small
		\renewcommand\tabcolsep{3pt}
		\renewcommand\arraystretch{1}
		\begin{center}
			\begin{tabular}{cccc||ccc}
				\hline
				\multicolumn{4}{c||}{Regression (RMSE)}                &                          \multicolumn{3}{c}{Classification (Accuracy \%)}                          \\ \hline
				Data Set &           Hyperband           &           SeqUD            &       scale        & Data Set &          Hyperband           &            SeqUD             \\ \hline
				R1    & $\underline{5.064}$$\pm$0.261 & $\mathbf{4.829}$$\pm$0.184 &  $\times  0.  1$   &   C1    &  $\mathbf{96.63}$$\pm$0.82   &        96.28$\pm$0.90        \\
				R2    &        7.304$\pm$0.252        & $\mathbf{7.283}$$\pm$0.253 &  $\times  0.  1$   &   C2    &  $\mathbf{69.97}$$\pm$0.98   &        69.28$\pm$2.14        \\
				R3    &        2.426$\pm$0.104        & $\mathbf{2.424}$$\pm$0.105 &  $\times  10000$   &   C3    &  $\mathbf{86.52}$$\pm$1.46   &  $\mathbf{86.52}$$\pm$1.49   \\
				R4    &        7.293$\pm$0.341        & $\mathbf{7.146}$$\pm$0.274 &   $\times  100$    &   C4    & $\underline{96.09}$$\pm$0.89 &  $\mathbf{96.63}$$\pm$0.54   \\
				R5    &        1.576$\pm$0.018        & $\mathbf{1.576}$$\pm$0.018 &    $\times  1$     &   C5    &        75.42$\pm$1.61        &  $\mathbf{75.62}$$\pm$1.29   \\
				R6    &  $\mathbf{2.424}$$\pm$0.204   &      2.436$\pm$0.193       &  $\times  0.  1$   &   C6    &        98.71$\pm$0.61        &  $\mathbf{98.81}$$\pm$0.41   \\
				R7    & $\underline{2.089}$$\pm$0.159 & $\mathbf{1.980}$$\pm$0.179 &    $\times  1$     &   C7    &  $\mathbf{75.12}$$\pm$1.44   & $\underline{74.48}$$\pm$1.14 \\
				R8    &        6.201$\pm$0.115        & $\mathbf{6.136}$$\pm$0.115 &  $\times  0.  1$   &   C8    &  $\mathbf{93.05}$$\pm$0.45   &        93.01$\pm$0.65        \\
				R9    & $\underline{9.280}$$\pm$0.130 & $\mathbf{9.209}$$\pm$0.132 &  $\times  0.  1$   &   C9    &        93.27$\pm$0.20        &  $\mathbf{93.36}$$\pm$0.11   \\
				R10   & $\underline{2.219}$$\pm$0.040 & $\mathbf{2.185}$$\pm$0.032 &    $\times  1$     &   C10   & $\underline{95.43}$$\pm$0.22 &  $\mathbf{95.50}$$\pm$0.23   \\
				R11   & $\underline{2.477}$$\pm$0.325 & $\mathbf{2.015}$$\pm$0.111 &    $\times  1$     &   C11   &        74.13$\pm$0.78        &  $\mathbf{75.29}$$\pm$4.90   \\
				R12   & $\underline{3.144}$$\pm$0.034 & $\mathbf{3.114}$$\pm$0.032 &    $\times  1$     &   C12   &  $\mathbf{97.64}$$\pm$0.13   &        97.62$\pm$0.10        \\
				R13   &        2.792$\pm$0.068        & $\mathbf{2.753}$$\pm$0.051 &    $\times  1$     &   C13   &        97.61$\pm$0.27        &  $\mathbf{97.75}$$\pm$0.28   \\
				R14   & $\underline{2.899}$$\pm$0.153 & $\mathbf{2.886}$$\pm$0.154 &  $\times  0.0  1$  &   C14   &        81.23$\pm$0.19        &  $\mathbf{81.28}$$\pm$0.22   \\
				R15   & $\underline{3.355}$$\pm$0.091 & $\mathbf{3.276}$$\pm$0.053 &    $\times  1$     &   C15   & $\underline{92.84}$$\pm$0.50 &  $\mathbf{93.53}$$\pm$0.27   \\
				R16   & $\underline{9.757}$$\pm$0.476 & $\mathbf{9.042}$$\pm$0.195 & $\times  0.00  1$  &   C16   & $\underline{87.75}$$\pm$0.29 &  $\mathbf{87.96}$$\pm$0.16   \\
				R17   & $\underline{1.630}$$\pm$0.020 & $\mathbf{1.603}$$\pm$0.014 & $\times  0.000  1$ &   C17   &        87.08$\pm$0.21        &  $\mathbf{87.10}$$\pm$0.16   \\
				R18   & $\underline{2.292}$$\pm$0.077 & $\mathbf{2.214}$$\pm$0.052 & $\times  0.00  1$  &   C18   &        97.03$\pm$0.16        &  $\mathbf{97.09}$$\pm$0.17   \\
				R19   &        4.368$\pm$0.225        & $\mathbf{4.228}$$\pm$0.071 &    $\times  10$    &   C19   & $\underline{90.66}$$\pm$0.12 &  $\mathbf{90.76}$$\pm$0.10   \\
				R20   & $\underline{4.705}$$\pm$0.091 & $\mathbf{4.654}$$\pm$0.064 &  $\times  0.  1$   &   C20   & $\underline{92.08}$$\pm$0.60 &  $\mathbf{92.99}$$\pm$0.23   \\ \hline
			\end{tabular}
		\end{center}
		\caption{Test set performance comparison between HyperBand and SeqUD over XGBoost optimization tasks. The best performing methods are highlighted in bold and underlined results are significantly different from the best ones.}    \label{hb}
	\end{table}
	
	\subsection{Comparison with AutoSklearn}
%We also take AutoSklearn for comparison. 
Since AutoSklearn is a fully automated pipeline that integrates many machine learning models, it is employed to make comparison with SeqUD for the pipeline optimization task.  AutoSklearn is configured with the default settings, while SeqUD follows the pipeline optimization setup in Section~\ref{AutoML_experiments_Setup}.
In AutoSklearn, the termination of the optimization is mainly controlled by the time limit. For each data set, we first calculate the average time cost of SeqUD ($T_{sequd}$), then set the time limit of AutoSklearn to be 1.05$T_{sequd}$.
	
	The experimental results in Table~\ref{autosk} show that SeqUD beats AutoSklearn in 17 out of the 40 compared tasks but fails in the other 23 tasks. The results indicate that AutoSklearn is slightly better than SeqUD. The AutoSklearn package is mainly based on SMAC. The reasons why AutoSklearn can achieve such a good performance are summarized below. First, AutoSklearn has a much larger search space. For instance, it is optimized over tens of machine learning models and has a higher probability of achieving better predictive performance. Second, AutoSklearn is enhanced with meta-learning of data set information and ensemble learning. Therefore, the direct comparison between SeqUD and AutoSklearn is somehow unfair. 
	
Nevertheless, it is possible to enhance SeqUD with such practical pipeline optimization techniques in our future work.
	
	\begin{table}[!htp]
		\small
		\renewcommand\tabcolsep{3pt}
		\renewcommand\arraystretch{1}
		\begin{center}
			\begin{tabular}{cccc||ccc}
				\hline
				\multicolumn{4}{c||}{Regression (RMSE)}               &                      \multicolumn{3}{c}{Classification (Accuracy \%)}                      \\ \hline
				Data Set   &          AutoSklearn          &             SeqUD             &       scale        & Data Set &         AutoSklearn          &            SeqUD             \\ \hline
				R1  &        5.197$\pm$0.711        &  $\mathbf{4.829}$$\pm$0.184   &  $\times  0.  1$   &   C1    & $\underline{62.81}$$\pm$0.00 &  $\mathbf{96.95}$$\pm$0.63   \\
				R2  & $\underline{8.303}$$\pm$0.182 &  $\mathbf{7.283}$$\pm$0.253   &  $\times  0.  1$   &   C2    &  $\mathbf{70.58}$$\pm$1.30   & $\underline{68.66}$$\pm$2.70 \\
				R3  &  $\mathbf{2.421}$$\pm$0.107   &        2.424$\pm$0.105        &  $\times  10000$   &   C3    &  $\mathbf{85.88}$$\pm$1.22   &        85.83$\pm$1.50        \\
				R4  & $\underline{1.957}$$\pm$0.024 &  $\mathbf{0.715}$$\pm$0.027   &   $\times  1000$   &   C4    &  $\mathbf{96.57}$$\pm$0.72   &        96.54$\pm$0.84        \\
				R5  &  $\mathbf{1.568}$$\pm$0.017   & $\underline{1.576}$$\pm$0.018 &    $\times  1$     &   C5    & $\underline{72.14}$$\pm$4.72 &  $\mathbf{76.04}$$\pm$1.24   \\
				R6  & $\underline{2.127}$$\pm$1.515 &  $\mathbf{0.244}$$\pm$0.019   &    $\times  1$     &   C6    & $\underline{34.66}$$\pm$0.00 &  $\mathbf{99.35}$$\pm$0.30   \\
				R7  & $\underline{2.108}$$\pm$0.101 &  $\mathbf{1.980}$$\pm$0.179   &    $\times  1$     &   C7    &  $\mathbf{75.02}$$\pm$0.93   &        74.72$\pm$1.60        \\
				R8  &  $\mathbf{5.939}$$\pm$0.093   & $\underline{6.136}$$\pm$0.115 &  $\times  0.  1$   &   C8    &        92.77$\pm$0.40        &  $\mathbf{93.08}$$\pm$0.60   \\
				R9  & $\underline{9.459}$$\pm$0.185 &  $\mathbf{9.209}$$\pm$0.132   &  $\times  0.  1$   &   C9    &  $\mathbf{93.41}$$\pm$0.05   &        93.40$\pm$0.08        \\
				R10 &  $\mathbf{2.181}$$\pm$0.048   &        2.185$\pm$0.032        &    $\times  1$     &   C10   & $\underline{95.10}$$\pm$0.40 &  $\mathbf{95.46}$$\pm$0.55   \\
				R11 & $\underline{8.374}$$\pm$0.110 &  $\mathbf{2.015}$$\pm$0.111   &    $\times  1$     &   C11   & $\underline{89.83}$$\pm$0.47 &  $\mathbf{90.52}$$\pm$0.29   \\
				R12 & $\underline{3.338}$$\pm$0.245 &  $\mathbf{3.114}$$\pm$0.032   &    $\times  1$     &   C12   &  $\mathbf{97.73}$$\pm$0.16   &        97.72$\pm$0.18        \\
				R13 &  $\mathbf{2.723}$$\pm$0.085   & $\underline{2.753}$$\pm$0.051 &    $\times  1$     &   C13   &        98.58$\pm$0.14        &  $\mathbf{98.62}$$\pm$0.16   \\
				R14 &  $\mathbf{2.868}$$\pm$0.158   & $\underline{2.886}$$\pm$0.154 &  $\times  0.0  1$  &   C14   &  $\mathbf{81.57}$$\pm$0.23   & $\underline{81.23}$$\pm$0.22 \\
				R15 &  $\mathbf{3.234}$$\pm$0.064   & $\underline{3.276}$$\pm$0.053 &    $\times  1$     &   C15   &  $\mathbf{96.91}$$\pm$0.95   & $\underline{92.87}$$\pm$1.05 \\
				R16 &  $\mathbf{7.980}$$\pm$0.135   & $\underline{9.042}$$\pm$0.195 & $\times  0.00  1$  &   C16   &  $\mathbf{88.47}$$\pm$0.24   & $\underline{87.86}$$\pm$0.27 \\
				R17 &  $\mathbf{1.584}$$\pm$0.020   & $\underline{1.603}$$\pm$0.014 & $\times  0.000  1$ &   C17   &  $\mathbf{87.14}$$\pm$0.14   &        87.09$\pm$0.15        \\
				R18 &  $\mathbf{2.184}$$\pm$0.042   & $\underline{2.214}$$\pm$0.052 & $\times  0.00  1$  &   C18   & $\underline{96.96}$$\pm$0.18 &  $\mathbf{97.07}$$\pm$0.16   \\
				R19 &  $\mathbf{4.187}$$\pm$0.172   &        4.228$\pm$0.071        &    $\times  10$    &   C19   &  $\mathbf{90.79}$$\pm$0.11   & $\underline{90.70}$$\pm$0.10 \\
				R20 &  $\mathbf{4.507}$$\pm$0.042   & $\underline{4.654}$$\pm$0.064 &  $\times  0.  1$   &   C20   & $\underline{91.96}$$\pm$1.08 &  $\mathbf{92.66}$$\pm$0.11   \\ \hline
			\end{tabular}
		\end{center}
		\caption{Test set performance comparison between AutoSklearn and SeqUD over Pipeline optimization tasks. The best performing methods are highlighted in bold and underlined results are significantly different from the best ones. }    \label{autosk}
	\end{table}

	\section{Conclusion} \label{Conclusion}
	In this paper, we propose a SeqUD framework for global optimization and potential application in HPO of machine learning models. A real-time AugUD algorithm is introduced for fast construction of augmented uniform designs. The proposed SeqUD combines the benefits of uniform designs and sequential exploitation. It balances exploration and exploitation, and could be easily parallelized to save computation time. Both synthetic function optimization and HPO experiments are conducted. The results show that the proposed approach is a competitive alternative to existing HPO methods. 
	
	The SeqUD method can be improved in several directions.  First, to avoid SeqUD trials being trapped into local optima, the multiple shooting strategy mentioned in Section~\ref{SeqUD} is worthy of our future investigation. Second, the generalization ability of different hyperparameters can be incorporated into the optimization framework. Third, the current SeqUD implementation requires a suitable pre-specified range of each hyperparameter and needs to determine the number of runs/levels per stage. In the future, it is a promising direction to extend SeqUD to a fully automated HPO framework.
	
	%%% Acknowledgements should go at the end, before appendices and references

	\acks{We thank the action editor and the four independent reviewers for their valuable comments and suggestions, which helped us improve the quality of the paper.}

\clearpage
	
	%%% Manual newpage inserted to improve layout of sample file - not
	%%% needed in general before appendices/bibliography.
	% \newpage
	
	%%% Appendix
	\appendix
	\section{Additional Results}
	\begin{figure}[htp!]
		\centering
		\subfloat[ESE (Mean)]{
			\label{ddese_ir_avgall} %% label for first subfigure
			\includegraphics[width=0.48\textwidth]{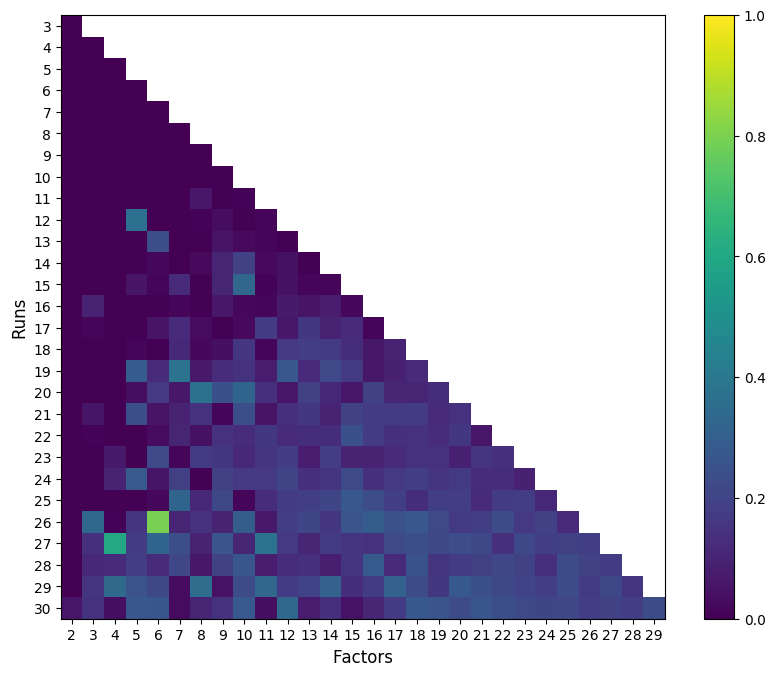}}
		\subfloat[AugUD (Mean)]{
			\label{pydoe_ir_avgall} %% label for second subfigure
			\includegraphics[width=0.48\textwidth]{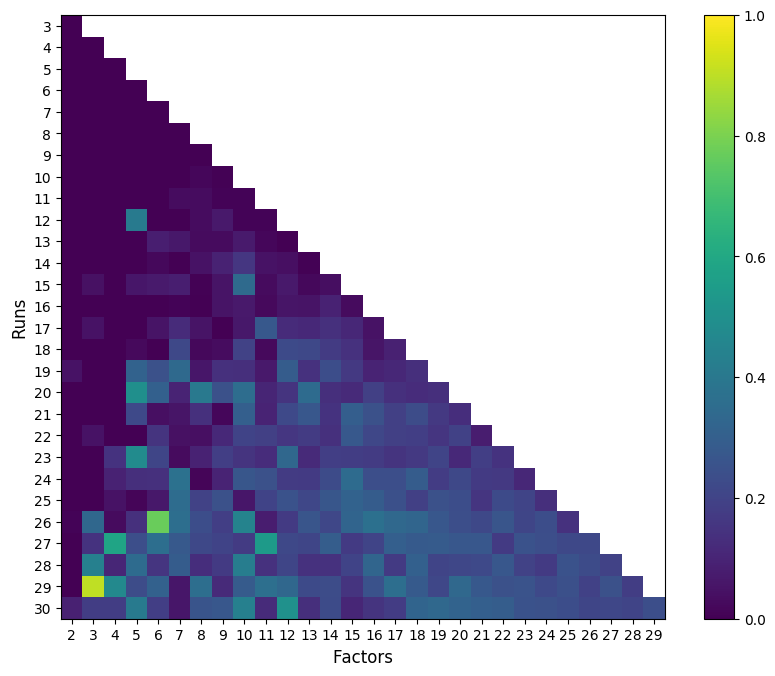}} \\
		\subfloat[ESE (Standard Deviation)]{
			\label{ddese_ir_stdall} %% label for first subfigure
			\includegraphics[width=0.48\textwidth]{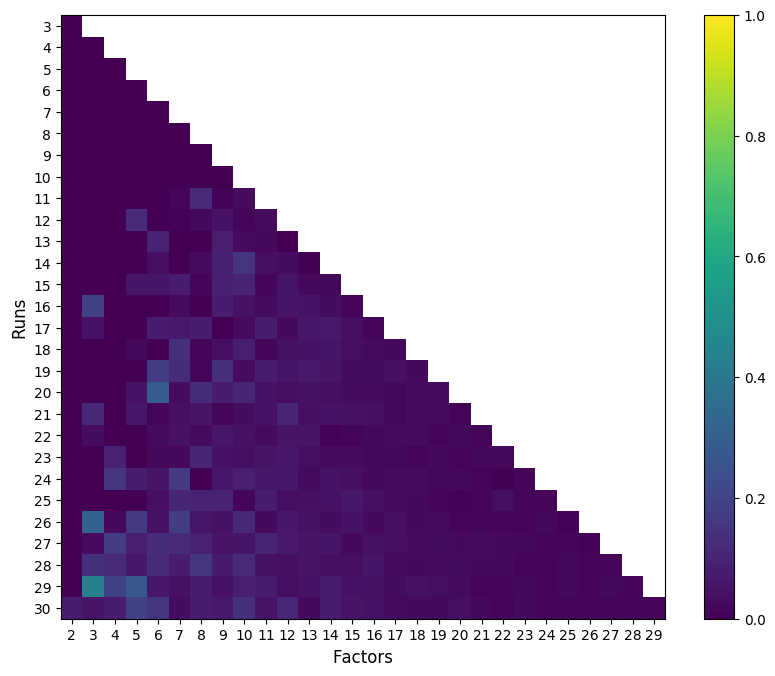}}
		\subfloat[AugUD  (Standard Deviation)]{
			\label{pydoe_ir_stdall} %% label for second subfigure
			\includegraphics[width=0.48\textwidth]{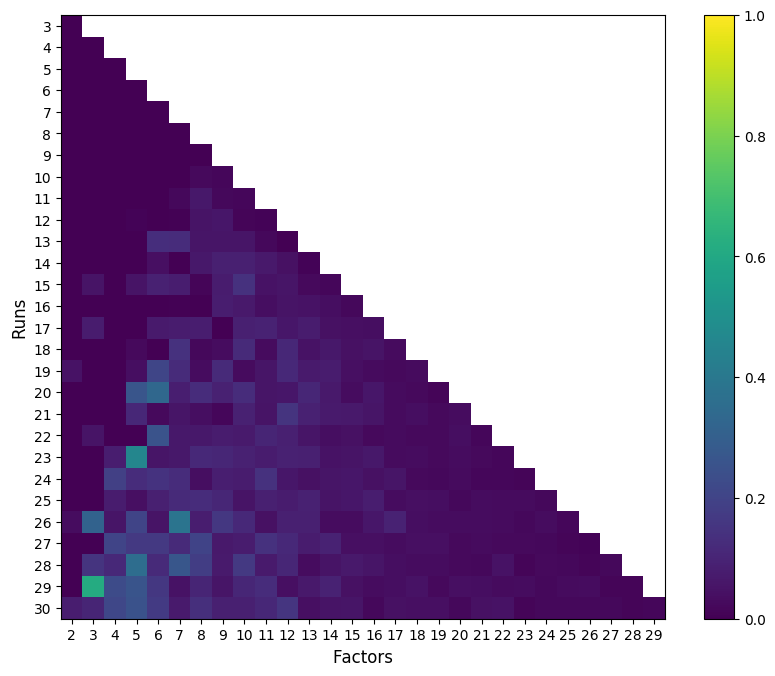}}
		\caption{Average Improvement ratios (\%) of ESE and AugUD against uniform designs obtained from the uniform design website. For each sub-figure, the x-axis represents the number of factors and the y-axis represents the number of runs.}
		\label{UD_OPT_ALL} %% label for entire figure
	\end{figure}

	\begin{figure}[htp!]
		\centering
		\subfloat[Nested UD (Mean)]{
			\label{nest_ir_avgall} %% label for first subfigure
			\includegraphics[width=0.48\textwidth]{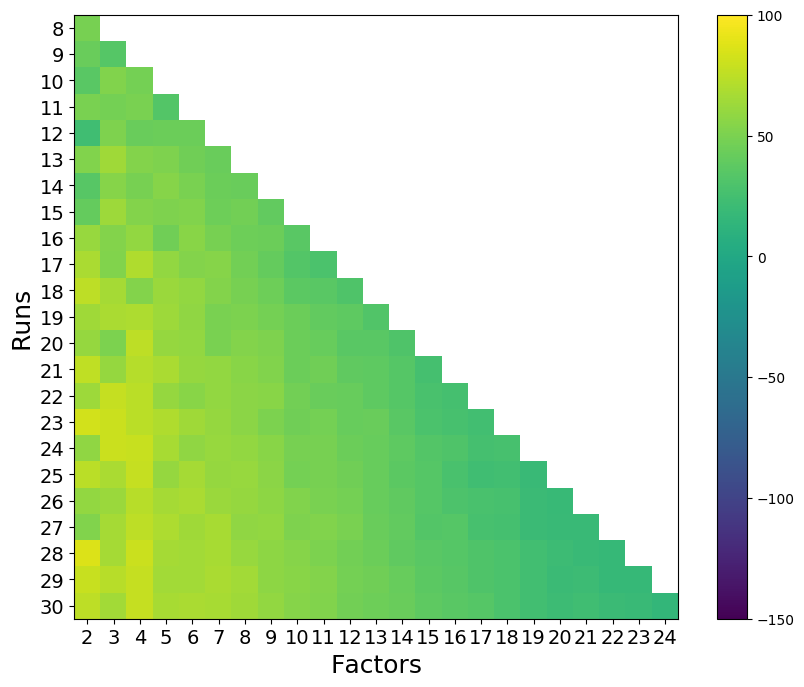}}
		\subfloat[AugUD  (Mean)]{
			\label{augud_ir_avgall} %% label for second subfigure
			\includegraphics[width=0.48\textwidth]{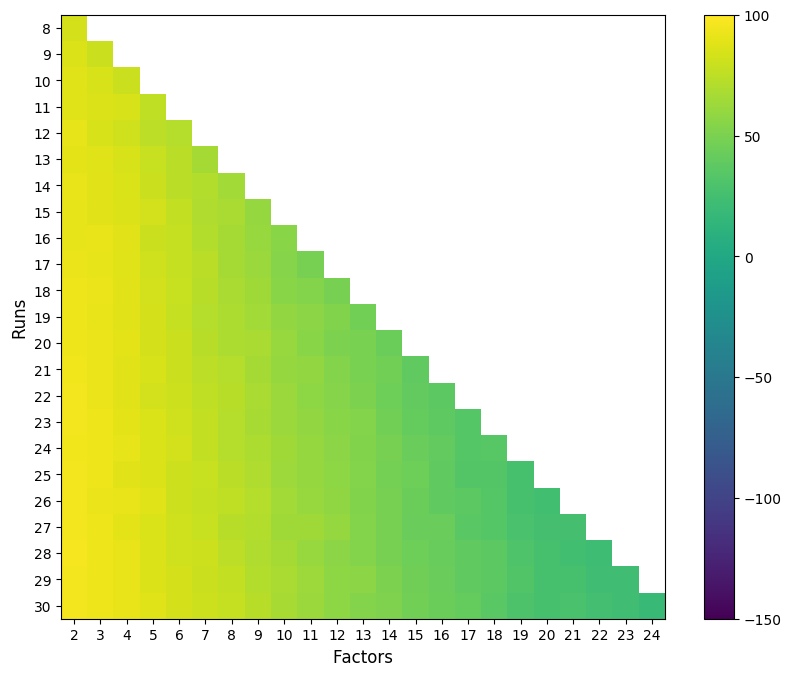}} \\
		\subfloat[Nested UD (Standard Deviation)]{
			\label{nest_ir_stdall} %% label for first subfigure
			\includegraphics[width=0.48\textwidth]{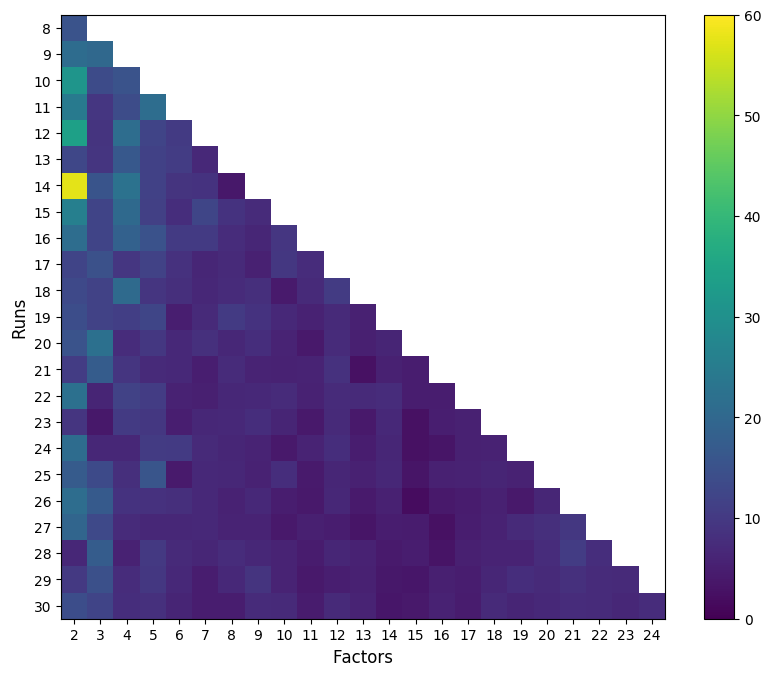}}
		\subfloat[AugUD  (Standard Deviation)]{
			\label{augud_ir_stdall} %% label for second subfigure
			\includegraphics[width=0.48\textwidth]{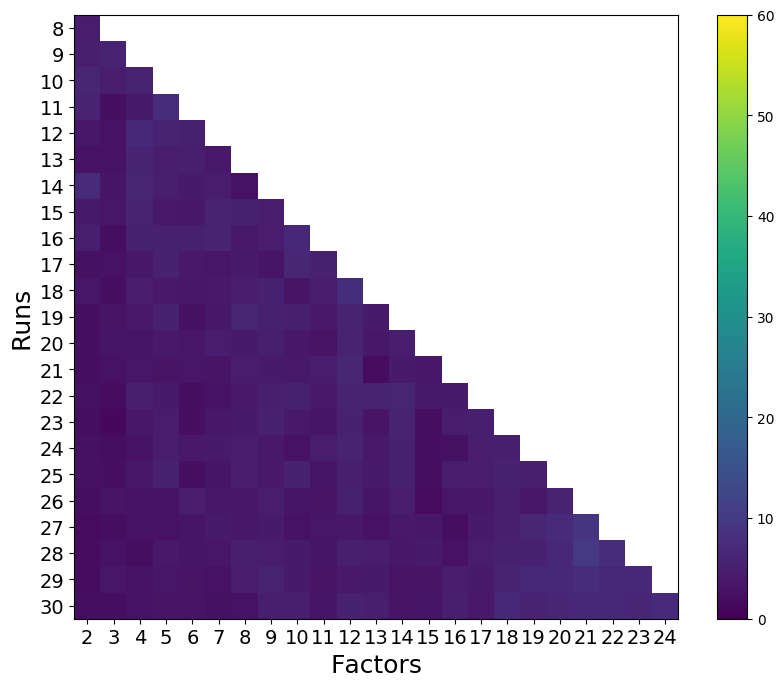}}
		\caption{Average improvement ratios (\%) of nested UD and AugUD against random augmentation. For each sub-figure, the x-axis represents the number of factors and the y-axis represents the number of runs.}
		\label{AUD_OPT_ALL} %% label for entire figure
	\end{figure}

	\begin{figure}[htp!]
		\centering 
		\subfloat[Grid]{\label{cliff_Grid}	
			\includegraphics[width=0.32\linewidth]{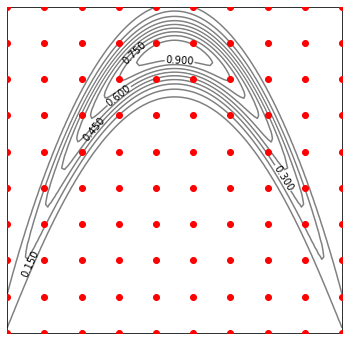}}
		\subfloat[Rand]{\label{cliff_Rand}	
			\includegraphics[width=0.32\linewidth]{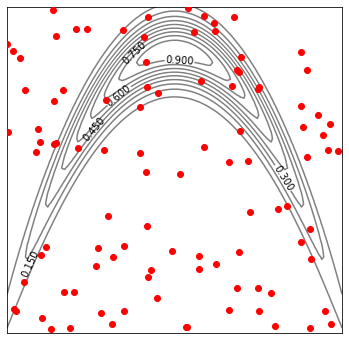}}
		\subfloat[LHS]{\label{cliff_LHS}	
			\includegraphics[width=0.32\linewidth]{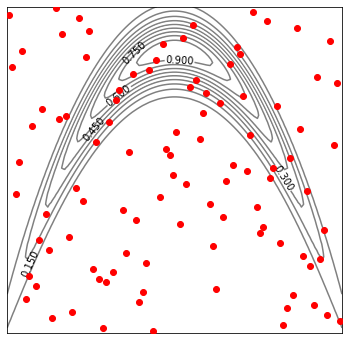}}\\
		\subfloat[Sobol]{\label{cliff_Sobol}	
			\includegraphics[width=0.32\linewidth]{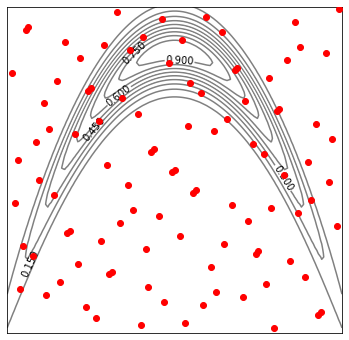}} 
		\subfloat[UD]{\label{cliff_UD}	
			\includegraphics[width=0.32\linewidth]{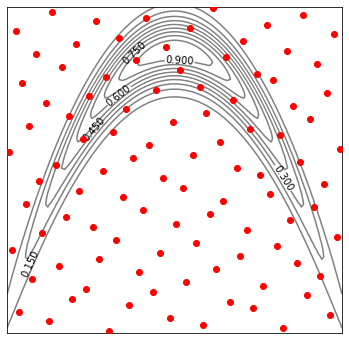}}
		\subfloat[SeqRand]{\label{cliff_SeqRand}	
			\includegraphics[width=0.32\linewidth]{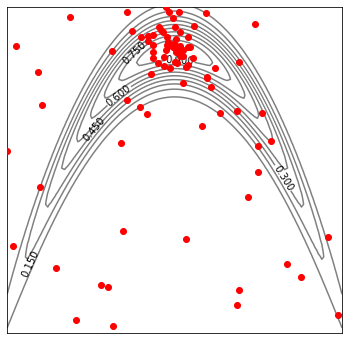}}\\
		\subfloat[TPE]{\label{cliff_TPE}
			\includegraphics[width=0.32\linewidth]{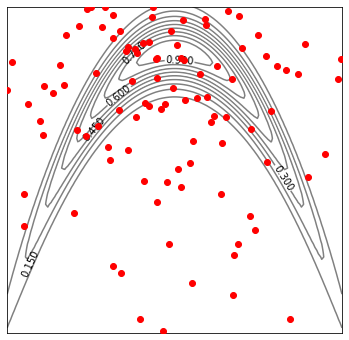}}
		\subfloat[SMAC]{\label{cliff_SMAC}	
			\includegraphics[width=0.32\linewidth]{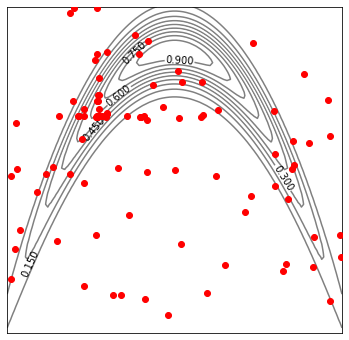}} 
		\subfloat[GP-EI]{\label{cliff_Spearmint}	
			\includegraphics[width=0.32\linewidth]{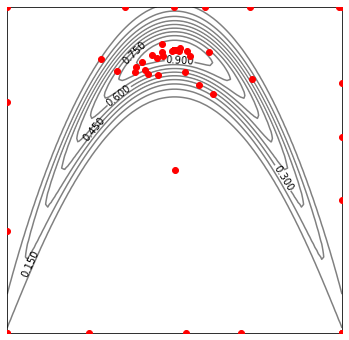}}
		\caption{The evaluated design points by each benchmark method against the ground truth contour plot of the cliff function. Each red point represents an evaluated point, and the actual optimal point of the cliff function is located in the upper center area.}
		\label{cliff_Demo}
	\end{figure}
	
	\newpage
	\begin{figure}[htp!]
		\centering 
		\subfloat[Grid]{\label{octopus_Grid}	
			\includegraphics[width=0.32\linewidth]{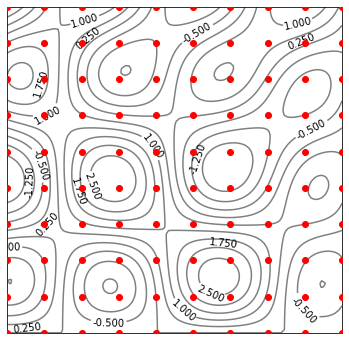}}
		\subfloat[Rand]{\label{octopus_Rand}	
			\includegraphics[width=0.32\linewidth]{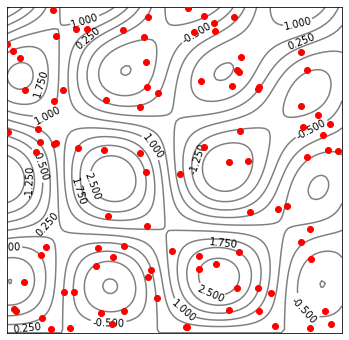}}
		\subfloat[LHS]{\label{octopus_LHS}	
			\includegraphics[width=0.32\linewidth]{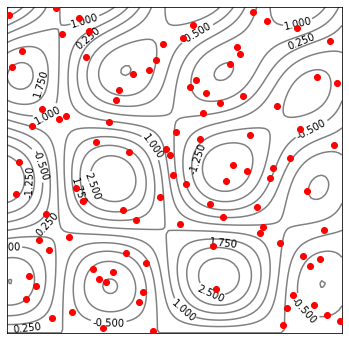}} \\
		\subfloat[Sobol]{\label{octopus_Sobol}	
			\includegraphics[width=0.32\linewidth]{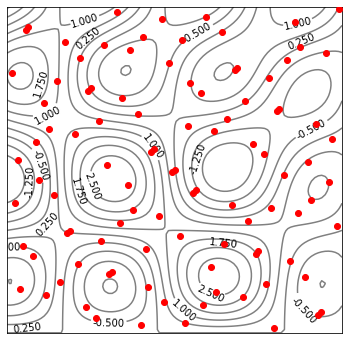}} 
		\subfloat[UD]{\label{octopus_UD}	
			\includegraphics[width=0.32\linewidth]{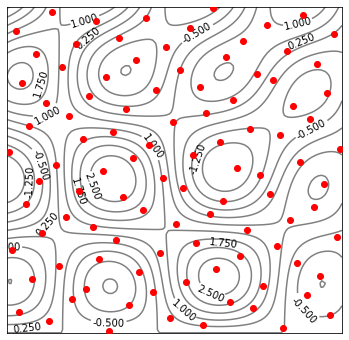}}
		\subfloat[SeqRand]{\label{octopus_SeqRand}	
			\includegraphics[width=0.33\linewidth]{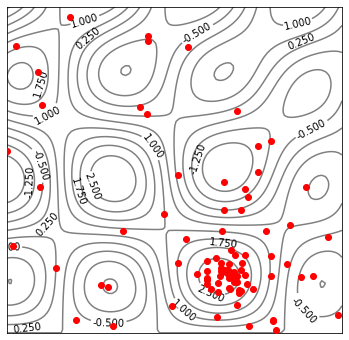}}\\
		\subfloat[TPE]{\label{octopus_TPE}
			\includegraphics[width=0.32\linewidth]{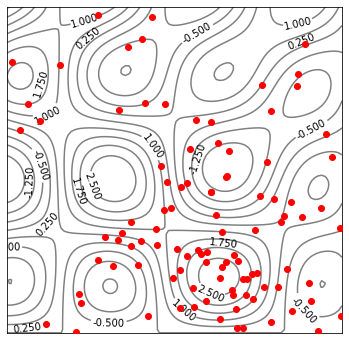}}
		\subfloat[SMAC]{\label{octopus_SMAC}	
			\includegraphics[width=0.32\linewidth]{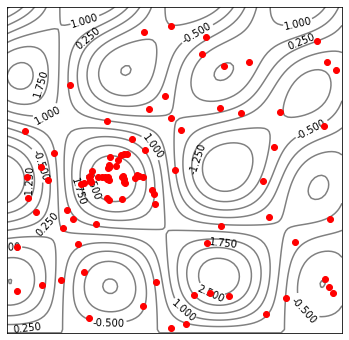}} 
		\subfloat[GP-EI]{\label{octopus_Spearmint}	
			\includegraphics[width=0.32\linewidth]{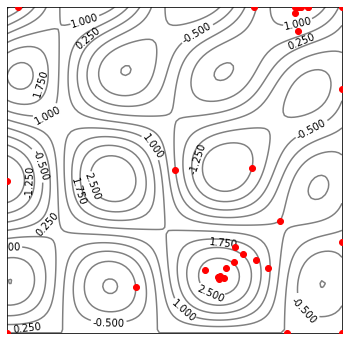}}
		\caption{The evaluated design points by each benchmark method against the ground truth contour plot of the octopus function. Each red point represents an evaluated point, and the actual optimal point of the octopus function is located in the center-left area.}
		\label{octopus_Demo}
	\end{figure}
	
	\begin{sidewaystable}
		\begin{table}[H]
			\tiny
			\renewcommand\tabcolsep{2.5pt}
			\renewcommand\arraystretch{1.5}
			\begin{center}
				\caption{Average optimal values found over 32 synthetic global optimization tasks (100 runs).} \label{SimuRes}
\begin{tabular}{c|cccccccccc|c}
					\hline
					Name&Grid&Rand&LHS&Sobol&UD&SeqRand&TPE&SMAC&GP-EI&SeqUD&Scale\\ \hline
					bukin6& $\underline{1.845}$$\pm$0.000 & $\underline{1.607}$$\pm$0.851 & $\underline{1.366}$$\pm$0.865 & $\underline{1.564}$$\pm$0.000 & $\underline{3.286}$$\pm$0.000 & $\underline{0.688}$$\pm$0.343 & $\underline{1.156}$$\pm$0.554 & $\underline{0.564}$$\pm$0.448 & $\mathbf{0.352}$$\pm$0.183 & $\underline{0.455}$$\pm$0.126 & $\times  10$ \\
					crossit& $\underline{-2.050}$$\pm$0.000 & $\underline{-2.021}$$\pm$0.038 & $\underline{-2.024}$$\pm$0.035 & $\underline{-2.060}$$\pm$0.000 & $\underline{-2.058}$$\pm$0.000 & $\underline{-2.048}$$\pm$0.044 & $\underline{-2.051}$$\pm$0.014 & $\underline{-2.045}$$\pm$0.025 & $\underline{-1.922}$$\pm$0.084 & $\mathbf{-2.063}$$\pm$0.000 & $\times  1$ \\
					egg& $\mathbf{-9.206}$$\pm$0.000 & $\underline{-7.438}$$\pm$0.985 & $\underline{-7.636}$$\pm$0.902 & $\underline{-6.736}$$\pm$0.000 & $\underline{-6.601}$$\pm$0.000 & $\underline{-7.470}$$\pm$1.369 & $\underline{-7.835}$$\pm$1.077 & $\underline{-7.234}$$\pm$1.236 & $\underline{-5.825}$$\pm$1.616 & $\underline{-7.355}$$\pm$0.286 & $\times  100$ \\
					holder& $\underline{-1.736}$$\pm$0.000 & $\underline{-1.589}$$\pm$0.248 & $\underline{-1.610}$$\pm$0.289 & $\underline{-1.213}$$\pm$0.000 & $\underline{-1.777}$$\pm$0.000 & $\underline{-1.740}$$\pm$0.349 & $\underline{-1.782}$$\pm$0.124 & $\underline{-1.819}$$\pm$0.228 & $\mathbf{-1.915}$$\pm$0.022 & $\underline{-0.950}$$\pm$0.000 & $\times  10$ \\
					langer& $\underline{-3.507}$$\pm$0.000 & $\underline{-3.287}$$\pm$0.638 & $\underline{-3.258}$$\pm$0.628 & $\underline{-2.781}$$\pm$0.000 & $\mathbf{-3.668}$$\pm$0.000 & $\underline{-3.358}$$\pm$0.841 & $\underline{-3.493}$$\pm$0.641 & $\underline{-3.380}$$\pm$0.753 & $\underline{-3.401}$$\pm$0.797 & $\underline{-1.478}$$\pm$0.000 & $\times  1$ \\
\hline 
				\end{tabular}
			\end{center}
		\end{table}
	\end{sidewaystable}
	
	\begin{sidewaystable}
		\begin{table}[H]
			\tiny
			\renewcommand\tabcolsep{2.5pt}
			\renewcommand\arraystretch{1.5}
			\begin{center}
				\caption{Average optimal values found over 32 synthetic global optimization tasks (1000 runs).} \label{SimuAblationRes}
				\begin{tabular}{c|ccccccccc|c}
					\hline
					Name&Grid&Rand&LHS&Sobol&UD&SeqRand&TPE&SMAC&SeqUD&Scale\\\hline
					bukin6& $\underline{2.811}$$\pm$0.000 & $\underline{4.737}$$\pm$2.387 & $\underline{4.658}$$\pm$2.510 & $\underline{0.978}$$\pm$0.000 & $\underline{6.392}$$\pm$0.000 & $\underline{2.397}$$\pm$1.164 & $\underline{2.996}$$\pm$1.475 & $\mathbf{0.864}$$\pm$0.465 & 0.931$\pm$0.433 & $\times  1$ \\
					crossit& $\underline{-2.048}$$\pm$0.000 & $\underline{-2.059}$$\pm$0.004 & $\underline{-2.058}$$\pm$0.005 & $\underline{-2.060}$$\pm$0.000 & $\underline{-2.050}$$\pm$0.000 & $\underline{-2.062}$$\pm$0.000 & $\underline{-2.062}$$\pm$0.000 & $\underline{-2.062}$$\pm$0.000 & $\mathbf{-2.063}$$\pm$0.000 & $\times  1$ \\
					egg& $\underline{-8.485}$$\pm$0.000 & $\underline{-8.903}$$\pm$0.345 & $\underline{-8.913}$$\pm$0.364 & $\underline{-8.957}$$\pm$0.000 & $\underline{-8.799}$$\pm$0.000 & $\underline{-9.102}$$\pm$0.527 & $\mathbf{-9.224}$$\pm$0.513 & -9.126$\pm$0.350 & $\underline{-7.166}$$\pm$0.007 & $\times  100$ \\
					holder& $\underline{-1.830}$$\pm$0.000 & $\underline{-1.892}$$\pm$0.025 & $\underline{-1.889}$$\pm$0.024 & $\underline{-1.891}$$\pm$0.000 & $\underline{-1.903}$$\pm$0.000 & $\underline{-1.918}$$\pm$0.002 & $\underline{-1.918}$$\pm$0.004 & $\underline{-1.921}$$\pm$0.000 & $\mathbf{-1.921}$$\pm$0.000 & $\times  10$ \\
					langer& $\underline{-4.022}$$\pm$0.000 & $\underline{-3.942}$$\pm$0.135 & $\underline{-3.933}$$\pm$0.137 & $\underline{-4.034}$$\pm$0.000 & $\underline{-3.887}$$\pm$0.000 & $\underline{-4.107}$$\pm$0.029 & $\underline{-3.940}$$\pm$0.517 & $\underline{-4.055}$$\pm$0.099 & $\mathbf{-4.122}$$\pm$0.030 & $\times  1$ \\
					levy& $\underline{0.122}$$\pm$0.000 & $\underline{4.988}$$\pm$4.569 & $\underline{5.004}$$\pm$4.784 & $\underline{4.350}$$\pm$0.000 & $\underline{2.727}$$\pm$0.000 & $\underline{0.068}$$\pm$0.061 & $\underline{0.224}$$\pm$0.258 & $\underline{1.045}$$\pm$2.268 & $\mathbf{0.001}$$\pm$0.001 & $\times  0.0  1$ \\
\hline
				\end{tabular}
			\end{center}
		\end{table}	
	\end{sidewaystable}
	
	\begin{sidewaystable}
		\begin{table}[H]
			\tiny
			\renewcommand\tabcolsep{2.5pt}
			\renewcommand\arraystretch{1.25}
			\begin{center}
				\caption{Average RMSE over different SVM-Reg tasks (5-fold CV).} \label{res_reg_svm_cv}
				\begin{tabular}{c|cccccccccc|c}
					\hline
					Data Set &             Grid              &             Rand              &              LHS              &             Sobol             &              UD               &     SeqRand     &            TPE             &             SMAC              &           GP-EI            &           SeqUD            &    scale                \\ \hline
					R1& $\underline{5.311}$$\pm$0.137 & $\underline{5.320}$$\pm$0.124 & $\underline{5.318}$$\pm$0.150 & $\underline{5.318}$$\pm$0.128 & 5.305$\pm$0.128 & 5.307$\pm$0.153 & 5.302$\pm$0.129 & 5.306$\pm$0.143 & $\mathbf{5.293}$$\pm$0.132 & 5.302$\pm$0.135 & $\times  0.  1$ \\
					R2& $\underline{7.380}$$\pm$0.184 & $\underline{7.372}$$\pm$0.179 & $\underline{7.362}$$\pm$0.184 & $\underline{7.356}$$\pm$0.176 & $\underline{7.369}$$\pm$0.181 & $\underline{7.351}$$\pm$0.182 & $\underline{7.360}$$\pm$0.177 & $\underline{7.361}$$\pm$0.186 & $\underline{7.353}$$\pm$0.175 & $\mathbf{7.334}$$\pm$0.181 & $\times  0.  1$ \\
					R3& $\mathbf{2.296}$$\pm$0.099 & 2.299$\pm$0.098 & $\underline{2.299}$$\pm$0.100 & 2.299$\pm$0.098 & 2.298$\pm$0.099 & 2.296$\pm$0.101 & 2.297$\pm$0.097 & 2.298$\pm$0.097 & 2.298$\pm$0.099 & 2.297$\pm$0.097 & $\times  10000$ \\
					R4& $\underline{8.102}$$\pm$0.165 & $\underline{8.094}$$\pm$0.181 & $\underline{8.114}$$\pm$0.170 & $\underline{8.120}$$\pm$0.166 & $\underline{8.098}$$\pm$0.174 & $\underline{8.071}$$\pm$0.182 & $\underline{8.081}$$\pm$0.173 & $\underline{8.091}$$\pm$0.173 & $\mathbf{8.055}$$\pm$0.166 & $\underline{8.070}$$\pm$0.172 & $\times  100$ \\
					R5& 1.564$\pm$0.020 & $\underline{1.565}$$\pm$0.020 & $\underline{1.565}$$\pm$0.021 & $\underline{1.565}$$\pm$0.021 & 1.564$\pm$0.020 & 1.564$\pm$0.020 & $\underline{1.564}$$\pm$0.020 & $\mathbf{1.563}$$\pm$0.020 & 1.564$\pm$0.020 & 1.564$\pm$0.021 & $\times  1$ \\
\hline
				\end{tabular}
				
				\caption{Average RMSE over different SVM-Reg tasks (test set).} \label{res_reg_svm_test}
				\begin{tabular}{c|cccccccccc|c}
					\hline
					Data Set &             Grid              &             Rand              &              LHS              &             Sobol             &              UD               &          SeqRand           &            TPE             &             SMAC              &           GP-EI            &           SeqUD            &  scale                  \\ \hline
					R1& 5.176$\pm$0.166 & 5.142$\pm$0.131 & 5.148$\pm$0.187 & 5.145$\pm$0.156 & 5.138$\pm$0.159 & 5.157$\pm$0.178 & 5.157$\pm$0.147 & $\mathbf{5.127}$$\pm$0.173 & 5.134$\pm$0.143 & 5.140$\pm$0.136 & $\times  0.  1$ \\
					R2& 7.466$\pm$0.252 & 7.476$\pm$0.232 & 7.453$\pm$0.237 & 7.475$\pm$0.238 & $\mathbf{7.441}$$\pm$0.243 & 7.452$\pm$0.235 & 7.473$\pm$0.228 & 7.453$\pm$0.233 & $\underline{7.469}$$\pm$0.234 & $\underline{7.470}$$\pm$0.234 & $\times  0.  1$ \\
					R3& 2.439$\pm$0.114 & 2.442$\pm$0.113 & 2.441$\pm$0.118 & $\mathbf{2.433}$$\pm$0.111 & 2.451$\pm$0.119 & 2.440$\pm$0.117 & 2.440$\pm$0.105 & 2.436$\pm$0.112 & 2.436$\pm$0.116 & 2.437$\pm$0.110 & $\times  10000$ \\
					R4& 8.102$\pm$0.325 & 8.108$\pm$0.317 & 8.141$\pm$0.301 & $\underline{8.124}$$\pm$0.297 & 8.114$\pm$0.294 & 8.085$\pm$0.306 & 8.085$\pm$0.305 & 8.120$\pm$0.240 & $\mathbf{8.075}$$\pm$0.301 & 8.104$\pm$0.320 & $\times  100$ \\
					R5& 1.577$\pm$0.020 & 1.575$\pm$0.019 & 1.575$\pm$0.019 & 1.578$\pm$0.017 & 1.579$\pm$0.014 & $\mathbf{1.575}$$\pm$0.018 & 1.576$\pm$0.017 & 1.580$\pm$0.016 & 1.579$\pm$0.015 & 1.575$\pm$0.017 & $\times  1$ \\
					R6& $\underline{9.764}$$\pm$0.336 & 9.325$\pm$0.430 & 9.303$\pm$0.355 & $\underline{9.278}$$\pm$0.330 & 9.279$\pm$0.344 & $\mathbf{9.261}$$\pm$0.336 & $\underline{9.271}$$\pm$0.338 & $\underline{9.310}$$\pm$0.351 & 9.272$\pm$0.353 & 9.268$\pm$0.348 & $\times  0.  1$ \\
\hline
				\end{tabular}
			\end{center}
		\end{table}	
	\end{sidewaystable}
	
	\begin{sidewaystable}
		\begin{table}[H]
			\tiny
			\renewcommand\tabcolsep{2.5pt}
			\renewcommand\arraystretch{1.25}
			\begin{center}
				\caption{Average RMSE over different XGB-Reg tasks (5-fold CV).} \label{res_reg_xgb_cv}
				\begin{tabular}{c|ccccccccc|c}
					\hline
					Data Set &             Rand              &              LHS              &             Sobol             &              UD               &            SeqRand            &              TPE              &             SMAC              &             GP-EI             &             SeqUD             &         scale          \\ \hline
					R1& $\underline{5.127}$$\pm$0.141 & $\underline{5.141}$$\pm$0.101 & $\underline{5.086}$$\pm$0.113 & 5.040$\pm$0.110 & $\underline{5.066}$$\pm$0.131 & 5.020$\pm$0.128 & $\mathbf{4.992}$$\pm$0.140 & 5.090$\pm$0.196 & $\underline{5.049}$$\pm$0.149 & $\times  0.  1$ \\
					R2& $\underline{7.235}$$\pm$0.201 & $\underline{7.243}$$\pm$0.232 & $\underline{7.262}$$\pm$0.197 & $\underline{7.255}$$\pm$0.213 & 7.171$\pm$0.188 & 7.206$\pm$0.210 & 7.199$\pm$0.219 & 7.216$\pm$0.227 & $\mathbf{7.167}$$\pm$0.198 & $\times  0.  1$ \\
					R3& $\underline{2.304}$$\pm$0.097 & $\underline{2.304}$$\pm$0.098 & 2.302$\pm$0.098 & $\underline{2.304}$$\pm$0.098 & 2.303$\pm$0.097 & 2.301$\pm$0.099 & 2.302$\pm$0.098 & $\underline{2.307}$$\pm$0.099 & $\mathbf{2.300}$$\pm$0.099 & $\times  10000$ \\
					R4& $\underline{7.126}$$\pm$0.280 & $\underline{7.153}$$\pm$0.272 & $\underline{7.121}$$\pm$0.269 & $\underline{7.156}$$\pm$0.283 & $\underline{6.996}$$\pm$0.272 & 6.951$\pm$0.271 & 6.961$\pm$0.283 & 7.030$\pm$0.399 & $\mathbf{6.940}$$\pm$0.240 & $\times  100$ \\
					R5& $\underline{1.570}$$\pm$0.019 & 1.573$\pm$0.021 & $\underline{1.569}$$\pm$0.019 & 1.569$\pm$0.019 & $\mathbf{1.569}$$\pm$0.019 & 1.570$\pm$0.019 & 1.569$\pm$0.019 & 1.573$\pm$0.019 & 1.569$\pm$0.019 & $\times  1$ \\
					R6& $\underline{2.592}$$\pm$0.157 & $\underline{2.637}$$\pm$0.175 & $\underline{2.580}$$\pm$0.123 & $\underline{2.592}$$\pm$0.148 & 2.464$\pm$0.136 & 2.488$\pm$0.100 & 2.522$\pm$0.137 & $\mathbf{2.442}$$\pm$0.120 & 2.475$\pm$0.099 & $\times  0.  1$ \\
\hline
				\end{tabular}
				
				\caption{Average RMSE over different XGB-Reg tasks (test set).} \label{res_reg_xgb_test}
				\begin{tabular}{c|ccccccccc|c}
					\hline
					Data Set &             Rand              &              LHS              &             Sobol             &              UD               &            SeqRand            &              TPE              &             SMAC              &             GP-EI             &             SeqUD             &          scale          \\ \hline
					R1& $\underline{4.950}$$\pm$0.209 & $\underline{4.981}$$\pm$0.183 & 4.880$\pm$0.206 & 4.891$\pm$0.177 & 4.880$\pm$0.164 & 4.885$\pm$0.237 & 4.904$\pm$0.243 & $\underline{5.067}$$\pm$0.290 & $\mathbf{4.829}$$\pm$0.184 & $\times  0.  1$ \\
					R2& 7.261$\pm$0.245 & 7.310$\pm$0.272 & $\underline{7.315}$$\pm$0.179 & $\underline{7.412}$$\pm$0.222 & 7.297$\pm$0.276 & 7.319$\pm$0.241 & $\mathbf{7.260}$$\pm$0.217 & 7.353$\pm$0.266 & 7.283$\pm$0.253 & $\times  0.  1$ \\
					R3& $\mathbf{2.423}$$\pm$0.105 & 2.429$\pm$0.101 & 2.424$\pm$0.107 & 2.433$\pm$0.094 & 2.426$\pm$0.106 & 2.427$\pm$0.107 & 2.424$\pm$0.106 & 2.427$\pm$0.097 & 2.424$\pm$0.105 & $\times  10000$ \\
					R4& $\underline{7.335}$$\pm$0.337 & 7.227$\pm$0.309 & $\underline{7.369}$$\pm$0.194 & 7.261$\pm$0.210 & 7.180$\pm$0.279 & 7.160$\pm$0.222 & 7.178$\pm$0.261 & 7.312$\pm$0.487 & $\mathbf{7.146}$$\pm$0.274 & $\times  100$ \\
					R5& 1.578$\pm$0.017 & 1.580$\pm$0.016 & 1.579$\pm$0.016 & $\mathbf{1.576}$$\pm$0.018 & 1.581$\pm$0.017 & $\underline{1.582}$$\pm$0.014 & 1.578$\pm$0.016 & 1.580$\pm$0.017 & 1.576$\pm$0.018 & $\times  1$ \\
					R6& $\underline{2.511}$$\pm$0.249 & $\underline{2.522}$$\pm$0.210 & 2.439$\pm$0.179 & 2.454$\pm$0.225 & $\mathbf{2.411}$$\pm$0.235 & $\underline{2.528}$$\pm$0.235 & $\underline{2.504}$$\pm$0.230 & $\underline{2.533}$$\pm$0.197 & 2.436$\pm$0.193 & $\times  0.  1$ \\
\hline
				\end{tabular}
			\end{center}
		\end{table}	
	\end{sidewaystable}
	
	\begin{sidewaystable}
		\begin{table}[H]
			\tiny
			\renewcommand\tabcolsep{2.5pt}
			\renewcommand\arraystretch{1.25}
			\begin{center}
				\caption{Average RMSE over different Pipe-Reg tasks (5-fold CV).} \label{res_reg_pipe_cv}
				\begin{tabular}{c|ccccccccc|c}
					\hline
					Data Set &             Rand              &              LHS              &             Sobol             &              UD               &            SeqRand            &              TPE              &             SMAC              &             GP-EI             &             SeqUD             &          scale          \\ \hline
					R1& 5.160$\pm$0.111 & 5.162$\pm$0.130 & $\underline{5.172}$$\pm$0.115 & $\underline{5.171}$$\pm$0.124 & $\underline{5.232}$$\pm$0.171 & $\mathbf{5.094}$$\pm$0.138 & $\underline{5.196}$$\pm$0.149 & $\underline{5.277}$$\pm$0.147 & 5.119$\pm$0.087 & $\times  0.  1$ \\
					R2& 7.288$\pm$0.138 & $\underline{7.354}$$\pm$0.217 & $\underline{7.312}$$\pm$0.180 & 7.262$\pm$0.152 & 7.252$\pm$0.132 & 7.266$\pm$0.278 & $\mathbf{7.216}$$\pm$0.222 & $\underline{7.405}$$\pm$0.241 & 7.243$\pm$0.228 & $\times  0.  1$ \\
					R3& $\underline{2.298}$$\pm$0.096 & $\mathbf{2.296}$$\pm$0.096 & 2.298$\pm$0.099 & 2.297$\pm$0.099 & 2.296$\pm$0.099 & 2.296$\pm$0.098 & 2.297$\pm$0.098 & $\underline{2.310}$$\pm$0.104 & 2.296$\pm$0.101 & $\times  10000$ \\
					R4& $\underline{7.500}$$\pm$0.413 & $\underline{7.509}$$\pm$0.381 & $\underline{7.324}$$\pm$0.255 & $\underline{7.465}$$\pm$0.434 & $\underline{7.546}$$\pm$0.515 & $\underline{7.463}$$\pm$0.491 & $\underline{7.537}$$\pm$0.271 & $\underline{8.346}$$\pm$0.767 & $\mathbf{6.905}$$\pm$0.265 & $\times  100$ \\
					R5& 1.565$\pm$0.020 & 1.566$\pm$0.020 & 1.565$\pm$0.020 & 1.565$\pm$0.020 & 1.565$\pm$0.020 & 1.566$\pm$0.020 & 1.565$\pm$0.020 & 1.568$\pm$0.020 & $\mathbf{1.564}$$\pm$0.020 & $\times  1$ \\
					R6& 2.623$\pm$0.208 & $\underline{2.595}$$\pm$0.218 & $\underline{2.673}$$\pm$0.132 & $\underline{2.638}$$\pm$0.180 & 2.536$\pm$0.199 & $\mathbf{2.473}$$\pm$0.161 & 2.567$\pm$0.105 & $\underline{5.760}$$\pm$2.329 & 2.484$\pm$0.143 & $\times  0.  1$ \\
\hline
				\end{tabular}
				
				\caption{Average RMSE over different Pipe-Reg tasks (test set).} \label{res_reg_pipe_test}
				\begin{tabular}{c|ccccccccc|c}
					\hline
					Data Set &             Rand              &              LHS              &             Sobol             &              UD               &            SeqRand            &              TPE              &             SMAC              &             GP-EI             &             SeqUD             &       scale             \\ \hline
					R1& 5.019$\pm$0.246 & 4.979$\pm$0.253 & $\mathbf{4.973}$$\pm$0.203 & 5.060$\pm$0.162 & $\underline{5.184}$$\pm$0.214 & $\underline{5.073}$$\pm$0.282 & 5.025$\pm$0.228 & $\underline{5.159}$$\pm$0.187 & $\underline{5.066}$$\pm$0.245 & $\times  0.  1$ \\
					R2& 7.467$\pm$0.256 & $\underline{7.458}$$\pm$0.218 & $\underline{7.500}$$\pm$0.201 & 7.381$\pm$0.269 & $\underline{7.459}$$\pm$0.293 & $\underline{7.486}$$\pm$0.157 & $\mathbf{7.347}$$\pm$0.223 & $\underline{7.592}$$\pm$0.243 & 7.422$\pm$0.218 & $\times  0.  1$ \\
					R3& 2.442$\pm$0.104 & $\underline{2.445}$$\pm$0.103 & $\mathbf{2.431}$$\pm$0.111 & 2.434$\pm$0.111 & 2.440$\pm$0.106 & $\underline{2.444}$$\pm$0.106 & 2.444$\pm$0.111 & 2.441$\pm$0.119 & 2.438$\pm$0.116 & $\times  10000$ \\
					R4& $\underline{7.590}$$\pm$0.434 & $\underline{7.685}$$\pm$0.407 & $\underline{7.408}$$\pm$0.300 & $\underline{7.718}$$\pm$0.416 & $\underline{7.776}$$\pm$0.542 & $\underline{7.536}$$\pm$0.509 & $\underline{7.737}$$\pm$0.490 & $\underline{8.620}$$\pm$0.729 & $\mathbf{7.142}$$\pm$0.209 & $\times  100$ \\
					R5& 1.580$\pm$0.016 & $\mathbf{1.576}$$\pm$0.016 & 1.579$\pm$0.017 & 1.582$\pm$0.015 & 1.578$\pm$0.015 & 1.577$\pm$0.017 & 1.579$\pm$0.015 & 1.581$\pm$0.018 & 1.576$\pm$0.018 & $\times  1$ \\
					R6& 2.561$\pm$0.255 & $\underline{2.607}$$\pm$0.094 & $\underline{2.564}$$\pm$0.288 & $\underline{2.587}$$\pm$0.220 & $\underline{2.582}$$\pm$0.248 & 2.497$\pm$0.222 & $\underline{2.565}$$\pm$0.255 & $\underline{5.622}$$\pm$2.440 & $\mathbf{2.449}$$\pm$0.181 & $\times  0.  1$ \\
\hline
				\end{tabular}
			\end{center}
		\end{table}	
	\end{sidewaystable}
	
	\begin{sidewaystable}
		\begin{table}[H]
			\tiny
			\renewcommand\tabcolsep{2.5pt}
			\renewcommand\arraystretch{1.25}
			\begin{center}
				\caption{Average Accuracy (\%) over different SVM-Cls tasks (5-fold CV).} \label{res_cls_svm_cv}
				\begin{tabular}{c|cccccccccc}
					\hline
					Data Set &             Grid             &             Rand             &             LHS              &            Sobol             &              UD              &           SeqRand            &             TPE              &             SMAC             &            GP-EI             &            SeqUD             \\ \hline
					C1& $\underline{98.17}$$\pm$0.74 & $\underline{98.24}$$\pm$0.76 & 98.27$\pm$0.68 & $\underline{98.24}$$\pm$0.72 & $\underline{98.31}$$\pm$0.67 & $\underline{98.31}$$\pm$0.72 & $\underline{98.31}$$\pm$0.72 & $\underline{98.24}$$\pm$0.74 & $\mathbf{98.41}$$\pm$0.68 & $\underline{98.31}$$\pm$0.77 \\
					C2& 72.48$\pm$0.88 & 72.24$\pm$0.77 & 72.10$\pm$0.58 & 72.24$\pm$0.82 & 72.31$\pm$0.83 & 72.45$\pm$1.10 & 72.31$\pm$1.15 & $\mathbf{72.48}$$\pm$1.15 & 72.20$\pm$0.88 & 72.20$\pm$1.18 \\
					C3& $\underline{86.41}$$\pm$1.09 & $\underline{86.35}$$\pm$1.05 & $\underline{86.49}$$\pm$0.99 & $\underline{86.43}$$\pm$0.88 & $\underline{86.17}$$\pm$1.13 & 86.49$\pm$0.98 & $\underline{86.61}$$\pm$0.94 & $\underline{86.55}$$\pm$0.98 & $\underline{86.43}$$\pm$1.04 & $\mathbf{86.96}$$\pm$1.06 \\
					C4& $\underline{97.17}$$\pm$0.66 & $\underline{97.17}$$\pm$0.71 & 97.25$\pm$0.64 & 97.22$\pm$0.70 & 97.28$\pm$0.68 & 97.28$\pm$0.65 & $\underline{97.17}$$\pm$0.70 & $\underline{97.14}$$\pm$0.63 & 97.22$\pm$0.70 & $\mathbf{97.34}$$\pm$0.62 \\
					C5& $\underline{77.76}$$\pm$1.40 & 78.04$\pm$1.30 & $\underline{77.97}$$\pm$1.33 & 78.17$\pm$1.46 & 78.07$\pm$1.41 & 78.20$\pm$1.45 & 78.28$\pm$1.30 & 77.92$\pm$1.37 & 78.20$\pm$1.28 & $\mathbf{78.33}$$\pm$1.66 \\
					C6& $\underline{99.02}$$\pm$0.16 & 99.16$\pm$0.19 & 99.16$\pm$0.19 & 99.19$\pm$0.17 & 99.17$\pm$0.19 & $\mathbf{99.21}$$\pm$0.16 & $\mathbf{99.21}$$\pm$0.16 & $\underline{99.14}$$\pm$0.15 & 99.19$\pm$0.15 & 99.19$\pm$0.17 \\
\hline
				\end{tabular}
				
				\caption{Average Accuracy (\%) over different SVM-Cls tasks (test set).} \label{res_cls_svm_test}
				\begin{tabular}{c|cccccccccc}
					\hline
					Data Set&Grid&Rand&LHS&Sobol&UD&SeqRand&TPE&SMAC&GP-EI&SeqUD\\ \hline
					C1& $\underline{96.74}$$\pm$0.57 & 96.81$\pm$0.64 & $\underline{96.70}$$\pm$0.55 & 96.95$\pm$0.63 & 96.88$\pm$0.46 & 96.67$\pm$1.16 & 96.84$\pm$0.61 & $\mathbf{97.02}$$\pm$0.55 & $\underline{96.77}$$\pm$0.60 & $\mathbf{97.02}$$\pm$0.50 \\
					C2& 69.25$\pm$2.70 & 70.21$\pm$1.97 & 69.69$\pm$2.40 & 69.90$\pm$2.27 & 69.45$\pm$2.21 & 70.10$\pm$2.19 & 69.59$\pm$2.15 & $\underline{69.35}$$\pm$2.51 & 69.59$\pm$2.40 & $\mathbf{70.38}$$\pm$2.10 \\
					C3& $\underline{85.91}$$\pm$1.24 & $\underline{85.83}$$\pm$1.39 & $\underline{86.26}$$\pm$1.52 & $\mathbf{87.13}$$\pm$1.42 & $\underline{85.77}$$\pm$1.09 & $\underline{85.83}$$\pm$1.28 & $\underline{85.88}$$\pm$1.09 & $\underline{85.71}$$\pm$1.06 & $\underline{85.59}$$\pm$1.05 & $\underline{86.20}$$\pm$1.16 \\
					C4& 96.37$\pm$0.60 & $\underline{96.37}$$\pm$0.77 & 96.49$\pm$0.91 & 96.26$\pm$0.66 & 96.57$\pm$0.76 & 96.46$\pm$0.84 & $\underline{96.37}$$\pm$0.77 & $\mathbf{96.60}$$\pm$0.74 & 96.46$\pm$0.88 & $\underline{96.31}$$\pm$0.81 \\
					C5& $\mathbf{76.59}$$\pm$1.64 & 76.17$\pm$1.48 & $\underline{75.89}$$\pm$1.83 & 76.17$\pm$1.46 & 76.25$\pm$1.69 & $\underline{76.22}$$\pm$1.76 & 76.35$\pm$1.74 & 76.35$\pm$1.50 & 76.25$\pm$1.52 & 76.35$\pm$1.45 \\
					C6& 99.31$\pm$0.31 & 99.33$\pm$0.33 & $\mathbf{99.46}$$\pm$0.23 & $\mathbf{99.46}$$\pm$0.21 & $\mathbf{99.46}$$\pm$0.23 & $\mathbf{99.46}$$\pm$0.21 & 99.42$\pm$0.20 & $\mathbf{99.46}$$\pm$0.23 & 99.44$\pm$0.30 & $\mathbf{99.46}$$\pm$0.28 \\
					C7& 74.84$\pm$1.29 & $\mathbf{75.18}$$\pm$1.06 & $\underline{74.66}$$\pm$0.93 & 74.94$\pm$1.17 & 74.78$\pm$1.13 & 74.54$\pm$1.57 & 74.92$\pm$1.15 & 74.80$\pm$1.27 & 74.56$\pm$1.95 & $\mathbf{75.18}$$\pm$1.16 \\
\hline
				\end{tabular}
			\end{center}
		\end{table}	
	\end{sidewaystable}
	
	\begin{sidewaystable}
		\begin{table}[H]
			\tiny
			\renewcommand\tabcolsep{2.5pt}
			\renewcommand\arraystretch{1.25}
			\begin{center}
				\caption{Average Accuracy (\%) over different XGB-Cls tasks (5-fold CV).} \label{res_cls_xgb_cv}
				\begin{tabular}{c|ccccccccc}
					\hline
					Data Set &             Rand             &             LHS              &            Sobol             &              UD              &           SeqRand            &             TPE              &             SMAC             &            GP-EI             &            SeqUD             \\ \hline
					C1& $\underline{97.22}$$\pm$0.60 & 97.32$\pm$1.10 & $\underline{97.57}$$\pm$0.62 & $\mathbf{97.89}$$\pm$0.67 & 97.68$\pm$0.59 & $\underline{97.22}$$\pm$0.68 & 97.71$\pm$0.60 & $\underline{96.86}$$\pm$0.86 & 97.78$\pm$0.74 \\
					C2& 73.41$\pm$1.12 & 73.55$\pm$1.19 & 73.58$\pm$1.06 & $\underline{73.44}$$\pm$1.33 & 73.34$\pm$1.51 & 73.75$\pm$1.02 & $\mathbf{73.95}$$\pm$1.65 & 73.44$\pm$1.55 & 73.68$\pm$1.08 \\
					C3& $\underline{86.96}$$\pm$0.79 & $\underline{87.33}$$\pm$1.00 & 87.45$\pm$0.88 & $\underline{87.28}$$\pm$0.96 & 87.54$\pm$0.84 & $\mathbf{87.68}$$\pm$0.83 & 87.54$\pm$0.93 & 87.42$\pm$1.07 & 87.54$\pm$1.08 \\
					C4& $\underline{97.14}$$\pm$0.52 & $\underline{97.22}$$\pm$0.48 & $\mathbf{97.36}$$\pm$0.49 & 97.31$\pm$0.51 & $\underline{97.19}$$\pm$0.51 & 97.25$\pm$0.63 & 97.22$\pm$0.54 & 97.16$\pm$0.52 & 97.31$\pm$0.46 \\
					C5& 77.50$\pm$2.00 & 77.29$\pm$1.92 & 77.55$\pm$1.56 & $\underline{77.42}$$\pm$1.81 & 77.53$\pm$1.96 & 77.50$\pm$1.68 & 77.42$\pm$1.68 & 77.63$\pm$1.72 & $\mathbf{77.87}$$\pm$1.61 \\
					C6& $\underline{98.56}$$\pm$0.24 & $\underline{98.41}$$\pm$0.35 & $\underline{98.48}$$\pm$0.31 & $\underline{98.48}$$\pm$0.23 & 98.60$\pm$0.34 & 98.68$\pm$0.27 & 98.64$\pm$0.33 & $\underline{98.60}$$\pm$0.25 & $\mathbf{98.73}$$\pm$0.36 \\
					C7& $\underline{76.42}$$\pm$1.69 & $\underline{76.34}$$\pm$1.59 & $\underline{76.22}$$\pm$1.44 & $\underline{76.56}$$\pm$1.42 & 76.70$\pm$1.49 & 76.80$\pm$1.52 & 76.68$\pm$1.60 & 76.86$\pm$1.72 & $\mathbf{76.88}$$\pm$1.38 \\
\hline
				\end{tabular}
				
				\caption{Average Accuracy (\%) over different XGB-Cls tasks (test set).} \label{res_cls_xgb_test}
				\begin{tabular}{c|ccccccccc}
					\hline
					Data Set &             Rand             &             LHS              &            Sobol             &              UD              &           SeqRand            &             TPE              &             SMAC             &            GP-EI             &            SeqUD             \\ \hline
					C1& $\underline{96.11}$$\pm$1.08 & 96.39$\pm$0.80 & $\mathbf{96.70}$$\pm$0.89 & 96.67$\pm$0.83 & $\underline{96.35}$$\pm$0.74 & 96.11$\pm$1.02 & 96.32$\pm$0.77 & $\underline{95.33}$$\pm$1.22 & 96.28$\pm$0.90 \\
					C2& 69.55$\pm$2.40 & 69.01$\pm$2.58 & 69.52$\pm$2.22 & $\mathbf{69.97}$$\pm$1.37 & 69.69$\pm$2.36 & 69.76$\pm$1.43 & $\underline{69.14}$$\pm$2.05 & 69.83$\pm$1.87 & 69.28$\pm$2.14 \\
					C3& 86.72$\pm$1.38 & $\mathbf{86.75}$$\pm$1.06 & 86.55$\pm$1.35 & 86.43$\pm$1.73 & $\underline{86.12}$$\pm$1.21 & 86.64$\pm$1.49 & $\underline{85.97}$$\pm$1.47 & 86.41$\pm$1.72 & 86.52$\pm$1.49 \\
					C4& $\underline{96.20}$$\pm$0.84 & 96.57$\pm$0.73 & 96.51$\pm$0.73 & 96.46$\pm$0.90 & $\underline{95.97}$$\pm$0.77 & 96.26$\pm$0.58 & 96.31$\pm$0.55 & $\underline{95.97}$$\pm$0.73 & $\mathbf{96.63}$$\pm$0.54 \\
					C5& $\mathbf{76.48}$$\pm$1.38 & $\underline{75.91}$$\pm$1.30 & 76.15$\pm$1.62 & $\underline{75.49}$$\pm$1.69 & 75.86$\pm$1.42 & $\underline{75.68}$$\pm$1.70 & $\underline{75.68}$$\pm$1.39 & $\underline{75.70}$$\pm$1.80 & $\underline{75.62}$$\pm$1.29 \\
					C6& 98.73$\pm$0.51 & 98.52$\pm$0.63 & 98.79$\pm$0.35 & 98.60$\pm$0.40 & 98.75$\pm$0.37 & 98.83$\pm$0.67 & $\mathbf{98.91}$$\pm$0.54 & $\underline{98.50}$$\pm$0.65 & 98.81$\pm$0.41 \\
					C7& 75.36$\pm$1.62 & 75.08$\pm$1.32 & 75.12$\pm$1.51 & 74.88$\pm$1.31 & 74.80$\pm$1.22 & $\mathbf{75.62}$$\pm$1.93 & 74.84$\pm$1.53 & 74.74$\pm$0.97 & $\underline{74.48}$$\pm$1.14 \\
\hline
				\end{tabular}
			\end{center}
		\end{table}	
	\end{sidewaystable}
	
	\begin{sidewaystable}
		\begin{table}[H]
			\tiny
			\renewcommand\tabcolsep{2.5pt}
			\renewcommand\arraystretch{1.25}
			\begin{center}
				\caption{Average Accuracy (\%) over different Pipe-Cls tasks (5-fold CV).} \label{res_cls_pipe_cv}
				\begin{tabular}{c|ccccccccc}
					\hline
					Data Set &             Rand             &             LHS              &            Sobol             &              UD              &          SeqRand          &             TPE              &             SMAC             &            GP-EI             &            SeqUD             \\ \hline
					C1& 98.06$\pm$0.74 & 98.13$\pm$0.67 & $\underline{97.99}$$\pm$0.76 & 98.20$\pm$0.87 & 98.13$\pm$0.72 & 98.17$\pm$0.61 & 98.20$\pm$0.78 & $\mathbf{98.24}$$\pm$0.69 & 98.24$\pm$0.69 \\
					C2& $\underline{73.41}$$\pm$0.93 & $\underline{73.40}$$\pm$0.81 & 74.09$\pm$1.64 & $\underline{73.24}$$\pm$0.79 & 73.78$\pm$0.95 & 74.03$\pm$0.97 & 74.33$\pm$1.05 & $\underline{73.61}$$\pm$1.64 & $\mathbf{74.44}$$\pm$1.87 \\
					C3& $\underline{86.93}$$\pm$0.97 & $\underline{86.99}$$\pm$0.83 & $\underline{87.28}$$\pm$1.02 & $\underline{86.99}$$\pm$0.97 & 87.28$\pm$1.07 & $\mathbf{87.57}$$\pm$0.83 & $\underline{87.04}$$\pm$1.11 & 87.16$\pm$1.14 & 87.48$\pm$0.91 \\
					C4& 97.48$\pm$0.58 & 97.42$\pm$0.56 & 97.45$\pm$0.60 & 97.42$\pm$0.56 & $\mathbf{97.54}$$\pm$0.57 & 97.45$\pm$0.48 & 97.51$\pm$0.42 & 97.42$\pm$0.46 & 97.48$\pm$0.51 \\
					C5& $\underline{77.99}$$\pm$1.40 & 78.10$\pm$1.43 & 78.10$\pm$1.36 & 78.33$\pm$1.50 & $\mathbf{78.54}$$\pm$1.46 & 78.36$\pm$1.62 & 78.25$\pm$1.26 & 78.18$\pm$1.49 & 78.46$\pm$1.77 \\
					C6& $\underline{99.15}$$\pm$0.47 & $\underline{99.00}$$\pm$0.53 & $\underline{98.56}$$\pm$0.33 & $\underline{98.98}$$\pm$0.29 & 99.48$\pm$0.45 & $\mathbf{99.54}$$\pm$0.36 & 99.37$\pm$0.44 & $\underline{98.83}$$\pm$0.44 & $\underline{99.10}$$\pm$0.25 \\
					C7& $\underline{76.62}$$\pm$1.37 & $\underline{76.72}$$\pm$1.08 & $\underline{76.90}$$\pm$1.16 & $\underline{76.82}$$\pm$1.19 & $\mathbf{77.34}$$\pm$1.18 & 77.24$\pm$1.31 & 77.02$\pm$1.12 & 77.16$\pm$1.28 & 77.26$\pm$1.23 \\
\hline
				\end{tabular}
				
				\caption{Average Accuracy (\%) over different Pipe-Cls tasks (test set).} \label{res_cls_pipe_test}
				\begin{tabular}{c|ccccccccc}
					\hline
					Data Set &             Rand             &             LHS              &            Sobol             &              UD              &          SeqRand          &             TPE              &             SMAC             &            GP-EI             &            SeqUD             \\ \hline
					C1& 96.60$\pm$0.54 & 96.77$\pm$0.68 & $\underline{96.18}$$\pm$1.65 & $\mathbf{97.02}$$\pm$0.55 & 96.63$\pm$0.98 & 96.67$\pm$0.65 & 96.74$\pm$0.92 & 96.67$\pm$1.28 & 96.95$\pm$0.63 \\
					C2& $\underline{68.80}$$\pm$1.57 & $\underline{69.62}$$\pm$1.62 & 69.38$\pm$1.62 & $\underline{69.18}$$\pm$1.96 & 70.07$\pm$2.13 & 69.73$\pm$2.19 & 69.83$\pm$1.76 & $\mathbf{70.14}$$\pm$1.67 & 68.66$\pm$2.70 \\
					C3& $\underline{85.77}$$\pm$1.58 & $\underline{86.20}$$\pm$1.31 & $\underline{86.35}$$\pm$1.77 & $\underline{86.03}$$\pm$1.75 & 86.20$\pm$0.99 & 85.91$\pm$1.79 & $\underline{85.83}$$\pm$1.31 & $\mathbf{86.38}$$\pm$1.69 & 85.83$\pm$1.50 \\
					C4& 96.20$\pm$0.97 & 96.40$\pm$0.63 & 96.40$\pm$0.87 & $\mathbf{96.83}$$\pm$0.66 & 96.26$\pm$0.80 & 96.31$\pm$0.73 & 96.40$\pm$0.71 & 96.60$\pm$0.63 & 96.54$\pm$0.84 \\
					C5& $\underline{76.38}$$\pm$1.20 & 75.68$\pm$2.12 & 76.38$\pm$1.58 & $\mathbf{76.59}$$\pm$1.55 & 75.70$\pm$2.00 & 75.99$\pm$2.14 & 76.17$\pm$1.26 & 75.89$\pm$1.42 & 76.04$\pm$1.24 \\
					C6& $\underline{99.37}$$\pm$0.35 & $\underline{99.67}$$\pm$0.38 & $\underline{99.02}$$\pm$0.32 & $\underline{99.33}$$\pm$0.43 & $\mathbf{99.75}$$\pm$0.35 & 99.62$\pm$0.33 & 99.62$\pm$0.35 & $\underline{99.23}$$\pm$0.54 & $\underline{99.35}$$\pm$0.30 \\
					C7& $\underline{75.08}$$\pm$0.96 & $\mathbf{75.20}$$\pm$1.07 & $\underline{74.34}$$\pm$1.98 & $\underline{73.76}$$\pm$1.42 & 74.68$\pm$1.02 & 74.84$\pm$1.36 & 73.94$\pm$1.77 & 74.44$\pm$1.54 & 74.72$\pm$1.60 \\
\hline
				\end{tabular}
			\end{center}
		\end{table}
	\end{sidewaystable}

	\bibliography{sequdrefs}
	
\end{document}